\documentclass{article}

\usepackage[preprint, nonatbib]{neurips_2024}

\usepackage[utf8]{inputenc} 
\usepackage[T1]{fontenc}    
\usepackage{url}            
\usepackage{booktabs}       
\usepackage{amsfonts}       
\usepackage{nicefrac}       
\usepackage{microtype}      
\usepackage{xcolor}         
\usepackage{makecell} 
\usepackage{adjustbox}
\usepackage{pifont}
\usepackage{bbding}
\usepackage{tabulary,multirow,xspace}
\usepackage{fixmath,mathtools,nicefrac,mmstyle}
\usepackage{subcaption}
\usepackage{caption}
\usepackage{float}
\usepackage{wrapfig} 
\usepackage[misc]{ifsym} 
\usepackage{colortbl}
\usepackage{xcolor}
\definecolor{mygray1}{gray}{.95}
\definecolor{mygray2}{gray}{.9}
\definecolor{mygray3}{gray}{.95}
\usepackage{pifont}

\newlength\savewidth
\newcolumntype{x}[1]{>{\centering\arraybackslash}p{#1pt}}

\newcommand{\app}{\raise.17ex\hbox{$\scriptstyle\sim$}}

\definecolor{linkcolor}{RGB}{255,0,0}
\definecolor{urlcolor}{RGB}{255,105,180}
\definecolor{citecolor}{RGB}{66,168,235}

\usepackage[pagebackref=true,breaklinks=true,letterpaper=true,colorlinks,bookmarks=false,citecolor=blue,linkcolor=blue]{hyperref}

\usepackage[symbol]{footmisc}  
\renewcommand{\thefootnote}{\fnsymbol{footnote}}

\usepackage{tikz}
\usepackage{xcolor}
\usepackage{graphicx}
\usepackage{booktabs}

\newcommand \footnoteONLYtext[1]
{
	\let \mybackup \thefootnote
	\let \thefootnote \relax
	\footnotetext{#1}
	\let \thefootnote \mybackup
	\let \mybackup \imareallyundefinedcommand
}

\title{An Empirical Study of GPT-4o Image Generation Capabilities}

\author{Sixiang Chen$^{1*}$, Jinbin Bai$^{2*}$, Zhuoran Zhao$^{1*}$, Tian Ye$^{1*}$, Qingyu Shi$^{3}$, Donghao Zhou$^{4}$, Wenhao Chai$^{5}$,  \\
\textbf{Xin Lin$^{6}$, Jianzong Wu$^{3}$, Chao Tang$^{3}$, Shilin Xu$^{3}$, Tao Zhang$^{6}$, Haobo Yuan$^{6}$, Yikang Zhou$^{6}$,} \\
\textbf{Wei Chow$^{2}$, Linfeng Li$^{2}$, Xiangtai Li$^{3^\dagger}$, Lei Zhu$^{1,7^\dagger}$, Lu Qi$^{6^\dagger}$}
\textbf{} \\
  {$^{1}$The Hong Kong University of Science and Technology (GZ)}
  {$^{2}$National University of Singapore}\\
  {$^{3}$Peking University}
  {$^{4}$The Chinese University of Hong Kong}
  {$^{5}$University of Washington}
  {$^{6}$Wuhan University} \\
  {$^{7}$The Hong Kong University of Science and Technology}
}

\begin{document}

\maketitle
\footnotetext[1]{Equal contributions.  \Letter:~ \texttt{schen691@connect.hkust-gz.edu.cn} $^\dagger$Corresponding authors.}

\begin{abstract}
The landscape of image generation has rapidly evolved, from early GAN-based approaches to diffusion models and, most recently, to unified generative architectures that seek to bridge understanding and generation tasks.
Recent advances, especially the GPT-4o, have demonstrated the feasibility of high-fidelity multimodal generation, their architectural design remains mysterious and unpublished.
This prompts the question of whether image and text generation have already been successfully integrated into a unified framework for those methods.
In this work, we conduct an empirical study of GPT-4o's image generation capabilities, benchmarking it against leading open-source and commercial models. 
Our evaluation covers four main categories, including text-to-image, image-to-image, image-to-3D, and image-to-X generation, with more than 20 tasks.
Our analysis highlights the strengths and limitations of GPT-4o under various settings, and situates it within the broader evolution of generative modeling.
Through this investigation, we identify promising directions for future unified generative models, emphasizing the role of architectural design and data scaling.
For a high-definition version of the PDF, please refer to the link on GitHub: \href{https://github.com/Ephemeral182/Empirical-Study-of-GPT-4o-Image-Gen}{https://github.com/Ephemeral182/Empirical-Study-of-GPT-4o-Image-Gen}.

\end{abstract}

\section{Introduction}
Over the past decade, image generation has undergone a remarkable evolution—from the early successes of GANs~\cite{goodfellow2020generative} to the dominance of diffusion models~\cite{stablediffusion,sdxl,esser2024scaling}, which have significantly advanced image fidelity and diversity~\cite{heusel2017gans,barratt2018note}.
In parallel, Large Language Models (LLMs) have achieved exceptional performance across diverse natural language tasks by scaling autoregressive next-token prediction, demonstrating the power of unified modeling principles. 
These advances naturally raise a compelling question: can such principles be extended to image generation?

However, fundamental differences between autoregressive and diffusion-based paradigms present non-trivial challenges. 
Autoregressive models excel in sequential text generation, while diffusion models have become the \textit{de facto} standard for high-quality image synthesis. 
Bridging these modalities within a unified framework remains an open challenge. 
Several works~\cite{sun2023generative, wang2024illume, tong2024metamorph,seedx,dong2023dreamllm,chen2025multimodal} attempt to bridge this gap via multimodal connectors or instruction tuning, with LLMs serving as planning modules that produce intermediate representations for image generation.
While effective to some extent, these paradigms often exhibit limited interaction between text and image modalities, and struggle with content consistency—particularly in image-to-image generation and complex instruction-based synthesis.

To address these limitations, recent research explores unified generation models that integrate understanding and generation within a single architecture, following three main technical paradigms.
The first line of work represents both language and vision as discrete token sequences~\cite{liu2024world, team2024chameleon, wu2024vila, wang2024emu3, janus-pro, lumina-mgpt, wu2025harmonizing}, leveraging VQGAN~\cite{esser2021taming} or similar compressors to tokenize images for compatibility with autoregressive models.
A second direction integrates large language models directly into the diffusion process~\cite{zhou2024transfusion, zhao2024monoformer, xiao2024omnigen, ma2024janusflow}, employing them as denoising backbones for image generation and as unified sequence models for text. While promising, these approaches typically rely on intermediate compression modules such as VAEs or VQVAEs, which may limit visual fidelity or increase architectural complexity.
A third and increasingly prominent paradigm investigates discrete diffusion frameworks that natively support both image and text generation within a unified modeling space~\cite{SEDD, concrete, mdlm}. Building on this insight, recent works~\cite{li2024dual, swerdlow2025unified} propose fully end-to-end diffusion architectures based on shared Transformer backbones, demonstrating competitive performance and seamless modality integration comparable to similarly sized LLMs.

Despite these promising directions, such systems still lag behind the sophistication and generalization capabilities of proprietary models like Flux~\cite{flux} and Midjourney~\cite{Midjourney}, which may lack reasoning capabilities.

The recent release of GPT-4o~\cite{openai2025gpt4o} marks a significant milestone in multimodal generative modeling. 
As a native multimodal architecture, GPT-4o demonstrates strong capabilities in generating high-fidelity, photorealistic images while seamlessly unifying vision and language generation—reportedly in an autoregressive fashion.
However, its closed-source nature—particularly the lack of disclosure about its architecture, training regimen, and inference mechanisms—poses substantial challenges for scientific scrutiny. This motivates a careful empirical assessment of its capabilities relative to open-source state-of-the-art models.

Although the visual performance of GPT-4o and Gemini is widely recognized, much of their success likely stems from unprecedented scale in training data, model parameters, and compute resources. 
Prior studies, including diffusion models and connected-based models, suggest that scaling is a key enabler of generative quality—potentially more so than architectural novelty alone. 
These trends point to a promising trajectory for unified generative models: with sufficient scale, they may rival or even surpass today’s best proprietary systems.

In this study, we conduct a comprehensive evaluation of GPT-4o’s image generation performance, benchmarking its outputs against leading systems including Gemini 2.0 Flash Experimental~\cite{gemini-2-0-flash} and other state-of-the-art models. 
Building upon our comparative evaluation across text-to-image, image-to-image, image-to-3D, and image-to-X generation tasks, GPT-4o demonstrates several distinctive strengths:
\begin{itemize}
    \item \textbf{Exceptional Text Rendering Capability.} GPT-4o demonstrates exceptional capability in rendering textual elements within images, maintaining correct spelling, alignment, and formatting even in document-style generation tasks. This level of text fluency is rarely seen in prior models and is crucial for practical applications such as chart generation, document layout synthesis, and instruction-rich visual storytelling.
    \item \textbf{Compositional Generalization and Prompt Following.} GPT-4o displays impressive compositional abilities, accurately assembling complex scene elements, styles, or attributes described in prompts. This high prompt following enables it to handle fine-grained multi-attribute conditions in generation tasks with minimal loss of semantic detail.
    \item \textbf{Spatial Reasoning and Multi-View Consistency.} In generation tasks involving spatial manipulation, such as 3D view synthesis, camera control, and depth-conditioned rendering, GPT-4o maintains geometric consistency and viewpoint realism. This indicates an inherent capacity for spatial reasoning and structural awareness, even without explicit 3D modeling modules.
    \item \textbf{Comprehensive Image Transformation Capability.} GPT-4o shows strong generalization across a wide spectrum of image-to-image tasks, ranging from low-level image restoration to high-level perceptual understanding. Without task-specific tuning, it almost handles diverse transformations such as denoising, deblurring, relighting, segmentation, and depth estimation. This suggests the model has learned robust visual priors and spatial semantics, enabling it to perform correction and abstract structural prediction under a unified framework.
\end{itemize}

However, limitations remain in inconsistent generation, hallucination, and data bias in underrepresented cultural elements and non-Latin scripts, highlighting current trade-offs in model design and training data coverage.

While we do not analyze the internal architecture or implementation details of GPT-4o in this paper\footnote{There is currently no definitive evidence regarding the specific implementation details or architectural design of GPT-4o’s image generation capabilities. To ensure the credibility and accuracy of our analysis, we will refrain from making speculative claims in current version.}, we believe it plays an important role toward unified multimodal generation. We also emphasize that model architecture is only one part of this progress—training data, model scale, and optimization strategies are equally important. We hope future work will provide more empirical evidence to better understand such proprietary systems and their position within this evolving research landscape.

\section{Evaluation}

As GPT-4o’s image generation capability has only recently been released and no API is available, we conduct only qualitative comparisons between GPT-4o, Gemini 2.0 Flash~\cite{gemini-2-0-flash}, and other state-of-the-art models in their respective domains. 

\newcommand{\roundedboxpink}[1]{
  \tikz[baseline=(char.base)]{
    \node[anchor=south west, rounded corners, text height=1.5ex, text depth=.25ex, fill=pink, draw=none, text=black, font=\bfseries] (char) {#1};
  }
}
\newcommand{\roundedboxgreen}[1]{
  \tikz[baseline=(char.base)]{
    \node[anchor=south west, rounded corners, text height=1.5ex, text depth=.25ex, fill=green!30, draw=none, text=black, font=\bfseries] (char) {#1};
  }
}
\newcommand{\roundedboxblue}[1]{
  \tikz[baseline=(char.base)]{
    \node[anchor=south west, rounded corners, text height=1.5ex, text depth=.25ex, fill=blue!30, draw=none, text=black, font=\bfseries] (char) {#1};
  }
}
\newcommand{\roundedboxyellow}[1]{
  \tikz[baseline=(char.base)]{
    \node[anchor=south west, rounded corners, text height=1.5ex, text depth=.25ex, fill=yellow!50, draw=none, text=black, font=\bfseries] (char) {#1};
  }
}
\newcommand{\roundedboxred}[1]{
  \tikz[baseline=(char.base)]{
    \node[anchor=south west, rounded corners, text height=1.5ex, text depth=.25ex, fill=red!30, draw=none, text=black, font=\bfseries] (char) {#1};
  }
}
\newcommand{\roundedboxpurple}[1]{
  \tikz[baseline=(char.base)]{
    \node[anchor=south west, rounded corners, text height=1.5ex, text depth=.25ex, fill=purple!50, draw=none, text=black, font=\bfseries] (char) {#1};
  }
}
\newcommand{\roundedboxbrown}[1]{
  \tikz[baseline=(char.base)]{
    \node[anchor=south west, rounded corners, text height=1.5ex, text depth=.25ex, fill=brown!30, draw=none, text=black, font=\bfseries] (char) {#1};
  }
}
\newcommand{\roundedboxorange}[1]{
  \tikz[baseline=(char.base)]{
    \node[anchor=south west, rounded corners, text height=1.5ex, text depth=.25ex, fill=orange!30, draw=none, text=black, font=\bfseries] (char) {#1};
  }
}
\newcommand{\roundedboxgray}[1]{
  \tikz[baseline=(char.base)]{
    \node[anchor=south west, rounded corners, text height=1.5ex, text depth=.25ex, fill=gray!50, draw=none, text=black, font=\bfseries] (char) {#1};
  }
}

\definecolor{customcolorred}{RGB}{225,159,156} 
\definecolor{customcolorgreen}{RGB}{5,204,151} 

\newcommand{\boxedred}[1]{
  \tikz[baseline=(char.base)]{
    \node[anchor=south west, rectangle, text height=1.5ex, text depth=.25ex, fill=customcolorred, draw=none, text=black, font=\bfseries] (char) {#1};
  }
}
\newcommand{\boxedgreen}[1]{
  \tikz[baseline=(char.base)]{
    \node[anchor=south west, rectangle, text height=1.5ex, text depth=.25ex, fill=customcolorgreen, draw=none, text=black, font=\bfseries] (char) {#1};
  }
}

\begin{table*}[htp]
\centering
\renewcommand{\arraystretch}{1.4}
\caption{GPT-4o vs. Baselines: Qualitative error analysis across image generation tasks.}
\label{tab:error_case_all}
\resizebox{1.\textwidth}{!}{%
\tiny
\begin{tabular}{llllccc}
\toprule
Case Figure & Meta-task & Sub-task & GPT-4o & Gemini-2.0-flash & Domain-SOTA \\
\midrule
\textcolor{red}{Figure~\ref{fig:t2i_1}} & \multirow{11}{*}{Text-to-Image} & \multirow{4}{*}{Complex Text Following} & \roundedboxgreen{Success} & \roundedboxorange{Failure to Follow Instructions} & \roundedboxorange{Failure to Follow Instructions} \\
\textcolor{red}{Figure~\ref{fig:t2i_2}}&  &  & \roundedboxgreen{Success} & \roundedboxorange{Failure to Follow Instructions} & \roundedboxorange{Failure to Follow Instructions}  \\
\textcolor{red}{Figure~\ref{fig:t2i_3}}&  &  & \roundedboxgreen{Success} & \roundedboxgreen{Success} & \roundedboxgreen{Success} \\
\textcolor{red}{Figure~\ref{fig:t2i_4}}&  &  & \roundedboxgreen{Success} & \roundedboxgreen{Success} & \roundedboxgreen{Success} \\
\cline{3-6}
\textcolor{red}{Figure~\ref{fig:text_short1}}&  & \multirow{3}{*}{Text Rendering} & \roundedboxgreen{Success} & \roundedboxgreen{Success} & \roundedboxgreen{Success} \\
\textcolor{red}{Figure~\ref{fig:text_long1}}&  &  & \roundedboxgreen{Success} & \roundedboxyellow{Low Visual Quality} & \roundedboxyellow{Low Visual Quality} \\
\textcolor{red}{Figure~\ref{fig:text_long2}}&  &  & \roundedboxgreen{Success} & \roundedboxyellow{Low Visual Quality} & \roundedboxyellow{Low Visual Quality} \\
\cline{3-6}
\textcolor{red}{Figure~\ref{fig:doc_1}}&  & \multirow{3}{*}{Document Generation} & \roundedboxgreen{Success} & \roundedboxyellow{Low Visual Quality} & \roundedboxyellow{Low Visual Quality}  \\
\textcolor{red}{Figure~\ref{fig:doc_2}}&  &  & \roundedboxgreen{Success} & \roundedboxyellow{Low Visual Quality} & \roundedboxyellow{Low Visual Quality}  \\
\textcolor{red}{Figure~\ref{fig:doc_3}}&  &  & \roundedboxgreen{Success} & \roundedboxyellow{Low Visual Quality} & \roundedboxyellow{Low Visual Quality} \\
\cline{3-6}
\textcolor{red}{Figure~\ref{fig:panorama_1}}&  & Panorama & \roundedboxblue{Lack of Knowledge} & \roundedboxgreen{Success} & \roundedboxgreen{Success} \\
\cline{2-6}

\textcolor{red}{Figure~\ref{fig:style_transfer_1}}& \multirow{32}{*}{Image-to-Image} & \multirow{2}{*}{Style Transfer} & \roundedboxgreen{Success} & \roundedboxblue{Lack of Knowledge} & \roundedboxblue{Lack of Knowledge} \\
\textcolor{red}{Figure~\ref{fig:style_transfer_2}}&  & & \roundedboxgreen{Success} & \roundedboxblue{Lack of Knowledge} & \roundedboxblue{Lack of Knowledge} \\
\cline{3-6}
\textcolor{red}{Figure~\ref{fig:edit_1}}&  & \multirow{6}{*}{Image Editing} & \roundedboxyellow{Low Visual Quality} & \roundedboxgreen{Success} & \roundedboxorange{Failure to Follow Instructions} \\
\textcolor{red}{Figure~\ref{fig:edit_2}}&  &  & \roundedboxorange{Failure to Follow Instructions} & 
\roundedboxorange{Failure to Follow Instructions} & \roundedboxorange{Failure to Follow Instructions} \\
\textcolor{red}{Figure~\ref{fig:edit_3}}&  &  & \roundedboxgreen{Success} & \roundedboxorange{Failure to Follow Instructions} & \roundedboxorange{Failure to Follow Instructions} \\
\textcolor{red}{Figure~\ref{fig:edit_4}}&  &  & \roundedboxgreen{Success} & \roundedboxorange{Failure to Follow Instructions} & \roundedboxorange{Failure to Follow Instructions} \\
\textcolor{red}{Figure~\ref{fig:edit_5}}&  &  & \roundedboxgreen{Success} & \roundedboxorange{Failure to Follow Instructions} & \roundedboxorange{Failure to Follow Instructions} \\
\textcolor{red}{Figure~\ref{fig:edit_6}}&  &  & \roundedboxgreen{Success} & \roundedboxred{Inconsistent Generation} & \roundedboxorange{Failure to Follow Instructions} \\
\cline{3-6}
\textcolor{red}{Figure~\ref{fig:custom_1}}&  & Single-Concept Customization & \roundedboxgreen{Success} & \roundedboxorange{Failure to Follow Instructions} & \roundedboxgreen{Success} \\
\textcolor{red}{Figure~\ref{fig:custom_2}}&  & Multi-Concept Customization & \roundedboxred{Inconsistent Generation} & \roundedboxred{Inconsistent Generation} & \roundedboxgreen{Success} \\
\cline{3-6}
\textcolor{red}{Figure~\ref{fig:story_generation_1}}&  & \multirow{2}{*}{Story Image Generation} & \roundedboxgreen{Success} & \roundedboxorange{Failure to Follow Instructions} &\roundedboxgreen{Success} \\
\textcolor{red}{Figure~\ref{fig:story_generation_2}}&  &  & \roundedboxgreen{Success} & \roundedboxred{Inconsistent Generation} & \roundedboxgreen{Success}\\
\cline{3-6}
\textcolor{red}{Figure~\ref{fig:low_level_1}}&  & Low-Level Vision-Denoising

& \roundedboxyellow{Low Visual Quality} & \roundedboxyellow{Low Visual Quality} & \roundedboxgreen{Success} \\
\textcolor{red}{Figure~\ref{fig:low_level_2}}&  &Low-Level Vision-Deraining  & \roundedboxgreen{Success} & \roundedboxred{Inconsistent Generation} & \roundedboxgreen{Success} \\
\textcolor{red}{Figure~\ref{fig:low_level_3}}&  &Low-Level Vision-Dehazing  & \roundedboxgreen{Success} & \roundedboxyellow{Low Visual Quality} & \roundedboxgreen{Success} \\
\textcolor{red}{Figure~\ref{fig:low_level_4}}&  &Low-Level Vision-Low Light Enhancement  & \roundedboxyellow{Low Visual Quality} & \roundedboxyellow{Low Visual Quality} & \roundedboxgreen{Success}  \\
\textcolor{red}{Figure~\ref{fig:low_level_5}}&  &Low-Level Vision-Debluring  & \roundedboxgreen{Success} & \roundedboxyellow{Low Visual Quality} & \roundedboxgreen{Success}  \\
\textcolor{red}{Figure~\ref{fig:low_level_6}}&  &Low-Level Vision-Super Resolution  & \roundedboxgreen{Success} & \roundedboxyellow{Low Visual Quality} & \roundedboxgreen{Success}  \\
\textcolor{red}{Figure~\ref{fig:low_level_7}}&  &Low-Level Vision-Inpainting  & \roundedboxred{Inconsistent Generation} & \roundedboxred{Inconsistent Generation} & \roundedboxgreen{Success} \\
\textcolor{red}{Figure~\ref{fig:low_level_8}}&  &Low-Level Vision-Outpainting  & \roundedboxred{Inconsistent Generation} & \roundedboxgreen{Success} & \roundedboxgreen{Success} \\
\textcolor{red}{Figure~\ref{fig:low_level_9}}&  &Low-Level Vision-Colorization  & \roundedboxgreen{Success} & \roundedboxgreen{Success} & \roundedboxgreen{Success} \\
\textcolor{red}{Figure~\ref{fig:low_level_10}}&  & Low-Level Vision-Shadow Removal & \roundedboxgreen{Success} & \roundedboxorange{Failure to Follow Instructions} & \roundedboxgreen{Success} \\
\textcolor{red}{Figure~\ref{fig:low_level_11}}&  &Low-Level Vision-Reflection Removal  & \roundedboxred{Inconsistent Generation} & \roundedboxorange{Failure to Follow Instructions} & \roundedboxgreen{Success} \\
\textcolor{red}{Figure~\ref{fig:low_level_12}}&  &Low-Level Vision-Relighting   & \roundedboxgreen{Success} & \roundedboxorange{Failure to Follow Instructions} & \roundedboxgreen{Success}\\
\cline{3-6}
\textcolor{red}{Figure~\ref{fig:control_1}}&  & Spatial Control-Canny & \roundedboxred{Inconsistent Generation} & \roundedboxorange{Failure to Follow Instructions} & \roundedboxgreen{Success} \\
\textcolor{red}{Figure~\ref{fig:control_2}}&  & Spatial Control-Depth & \roundedboxgreen{Success} & \roundedboxorange{Failure to Follow Instructions} & \roundedboxgreen{Success} \\
\textcolor{red}{Figure~\ref{fig:control_3}}&  & Spatial Control-Sketch & \roundedboxred{Inconsistent Generation} & \roundedboxred{Inconsistent Generation}  & \roundedboxgreen{Success} \\
\textcolor{red}{Figure~\ref{fig:control_4}}&  &Spatial Control-Pose  & \roundedboxgreen{Success} & \roundedboxred{Inconsistent Generation} & \roundedboxgreen{Success} \\
\textcolor{red}{Figure~\ref{fig:control_5}}&  & Spatial Control-Mask & \roundedboxred{Inconsistent Generation} & \roundedboxorange{Failure to Follow Instructions} & \roundedboxred{Inconsistent Generation} \\
\cline{3-6}
\textcolor{red}{Figure~\ref{fig:custom_3}}&  & \multirow{2}{*}{Camera Control} & \roundedboxred{Inconsistent Generation} & \roundedboxorange{Failure to Follow Instructions} & \roundedboxgreen{Success} \\
\textcolor{red}{Figure~\ref{fig:custom_4}}&  &  & \roundedboxorange{Failure to Follow Instructions} & \roundedboxorange{Failure to Follow Instructions} & \roundedboxgreen{Success} \\
\cline{3-6}
\textcolor{red}{Figure~\ref{fig:custom_5}}&  & In-Context Visual Prompting & \roundedboxorange{Failure to Follow Instructions} & \roundedboxorange{Failure to Follow Instructions} & N/A \\
\cline{2-6}

\textcolor{red}{Figure~\ref{fig:3D_1}}& \multirow{3}{*}{Image-to-3D} & Image to 3D Modeling & \roundedboxgreen{Success} & \roundedboxorange{Failure to Follow Instructions}  & \roundedboxorange{Failure to Follow Instructions} \\
\textcolor{red}{Figure~\ref{fig:3D_2}}& & UV Map to 3D Rendering & \roundedboxgreen{Success} & \roundedboxred{Inconsistent Generation}  & \roundedboxorange{Failure to Follow Instructions} \\
\textcolor{red}{Figure~\ref{fig:3D_4}}& & Novel View Synthesis & \roundedboxgreen{Success} & \roundedboxgreen{Success} & \roundedboxorange{Failure to Follow Instructions} \\
\cline{2-6}

\textcolor{red}{Figure~\ref{fig:res_seg}}& \multirow{17}{*}{Image-to-X} & \multirow{3}{*}{Image Segmentation} & \roundedboxorange{Failure to Follow Instructions} & \roundedboxorange{Failure to Follow Instructions} & \roundedboxgreen{Success} \\
\textcolor{red}{Figure~\ref{fig:semseg}}&  &  & \roundedboxgreen{Success} & \roundedboxorange{Failure to Follow Instructions} & \roundedboxgreen{Success} \\
\textcolor{red}{Figure~\ref{fig:panseg}}&  &  & \roundedboxgreen{Success} & \roundedboxorange{Failure to Follow Instructions} & \roundedboxgreen{Success} \\
\cline{3-6}
\textcolor{red}{Figure~\ref{fig:edge_det}}&  & \multirow{2}{*}{Edge Detection} & \roundedboxgreen{Success} & \roundedboxgreen{Success} & \roundedboxgreen{Success} \\
\textcolor{red}{Figure~\ref{fig:matting}}&  &  & \roundedboxgreen{Success} & \roundedboxorange{Failure to Follow Instructions} & \roundedboxgreen{Success}  \\
\cline{3-6}
\textcolor{red}{Figure~\ref{fig:sailent_object_det}}&  & \multirow{4}{*}{Salient Object} & \roundedboxgreen{Success} & \roundedboxorange{Failure to Follow Instructions} & \roundedboxgreen{Success}  \\
\textcolor{red}{Figure~\ref{fig:mirror_det}}&  &  & \roundedboxgreen{Success} & \roundedboxorange{Failure to Follow Instructions} & \roundedboxgreen{Success} \\
\textcolor{red}{Figure~\ref{fig:shadow_det}}&  &  & \roundedboxgreen{Success} & \roundedboxgreen{Success} & \roundedboxgreen{Success}  \\
\textcolor{red}{Figure~\ref{fig:camouflage_object}}&  &  & \roundedboxgreen{Success} & \roundedboxgreen{Success} & \roundedboxgreen{Success}  \\
\cline{3-6}
\textcolor{red}{Figure~\ref{fig:depth}}&  & Depth Estimation & \roundedboxgreen{Success} & \roundedboxorange{Failure to Follow Instructions} & \roundedboxgreen{Success}  \\
\textcolor{red}{Figure~\ref{fig:normal}}&  & Normal Estimation & \roundedboxgreen{Success} & \roundedboxorange{Failure to Follow Instructions} & \roundedboxgreen{Success}  \\
\textcolor{red}{Figure~\ref{fig:layout_det}}&  & Layout Detection & \roundedboxred{Inconsistent Generation} & \roundedboxred{Inconsistent Generation} & \roundedboxgreen{Success} \\
\cline{3-6}
\textcolor{red}{Figure~\ref{fig:text_det}}&  & Text Detection & \roundedboxorange{Failure to Follow Instructions} & \roundedboxorange{Failure to Follow Instructions} & \roundedboxgreen{Success} \\
\cline{3-6}
\textcolor{red}{Figure~\ref{fig:obj_track_1}}&  & \multirow{4}{*}{Object Tracking} & \roundedboxred{Inconsistent Generation} & \roundedboxred{Inconsistent Generation} & \roundedboxgreen{Success} \\
\textcolor{red}{Figure~\ref{fig:obj_track_2}}&  &  & \roundedboxred{Inconsistent Generation} & \roundedboxred{Inconsistent Generation} & \roundedboxgreen{Success} \\
\textcolor{red}{Figure~\ref{fig:obj_track_3}}&    &  & \roundedboxred{Inconsistent Generation} & \roundedboxred{Inconsistent Generation} & \roundedboxgreen{Success} \\
\textcolor{red}{Figure~\ref{fig:obj_track_4}}&    &  & \roundedboxred{Inconsistent Generation} & \roundedboxred{Inconsistent Generation} & \roundedboxgreen{Success} \\
\bottomrule
\end{tabular}%
}
\end{table*}

To systematically compare these models' performance across diverse image generation tasks including text-to-image generation, image-to-image generation, text/image to 3D generation, and various image-to-X generation, we conduct a detailed case study focused on analyzing the performance of these models. This qualitative analysis provides insight into gpt 4o’s strengths and limitations in various tasks, as shown in Table~\ref{tab:error_case_all}.

\roundedboxyellow{Low Visual Quality}: The image synthesis model fails to generate fine-grained object details or produces blurry outputs. Typical cases include distorted human bodies or unrealistic hand shapes.

\roundedboxred{Inconsistent Generation}: The image synthesis model produces inconsistent output or image details with input image.

\roundedboxblue{Lack of Knowledge}: The image synthesis model lacks domain-specific knowledge, such as particular artistic styles, and thus generates visually plausible but incorrect results.


\roundedboxorange{Failure to Follow Instructions}: The image synthesis model misinterprets the input prompt and produces inconsistent results. For example, it may fail to capture specified numbers, colors, or object arrangements.

\clearpage

\subsection{Text-to-Image Tasks}
\subsubsection{Complex Text Following Capability}

Recent progress in text-to-image generation has shown impressive abilities in generating diverse and realistic images based on text prompts. However, composing multiple objects with various attributes and relationships accurately into one scene remains a significant challenge for current text-to-image generative models~\cite{saharia2022photorealistic, ramesh2022hierarchical, betker2023improving, podell2023sdxl,bai2024meissonic}. In this section, we assess models' ability for compositional text-to-image generation from four perspectives following~\cite{huang2025t2i}, which include attribute binding, numeracy, object relationship, and complex compositions. Attribute binding evaluates whether the model correctly assigns attributes, such as color, shape, and texture to the appropriate objects. Numeracy evaluates whether the number of generated objects matches the quantities specified in the prompt. Object relationships refer to both spatial (2D/3D) and non-spatial interactions among objects. Complex compositions evaluate the model's ability to handle multiple types of constraints simultaneously, especially given long or detailed prompts.

\begin{figure}[h]
    \centering
    \includegraphics[width=0.96\textwidth]{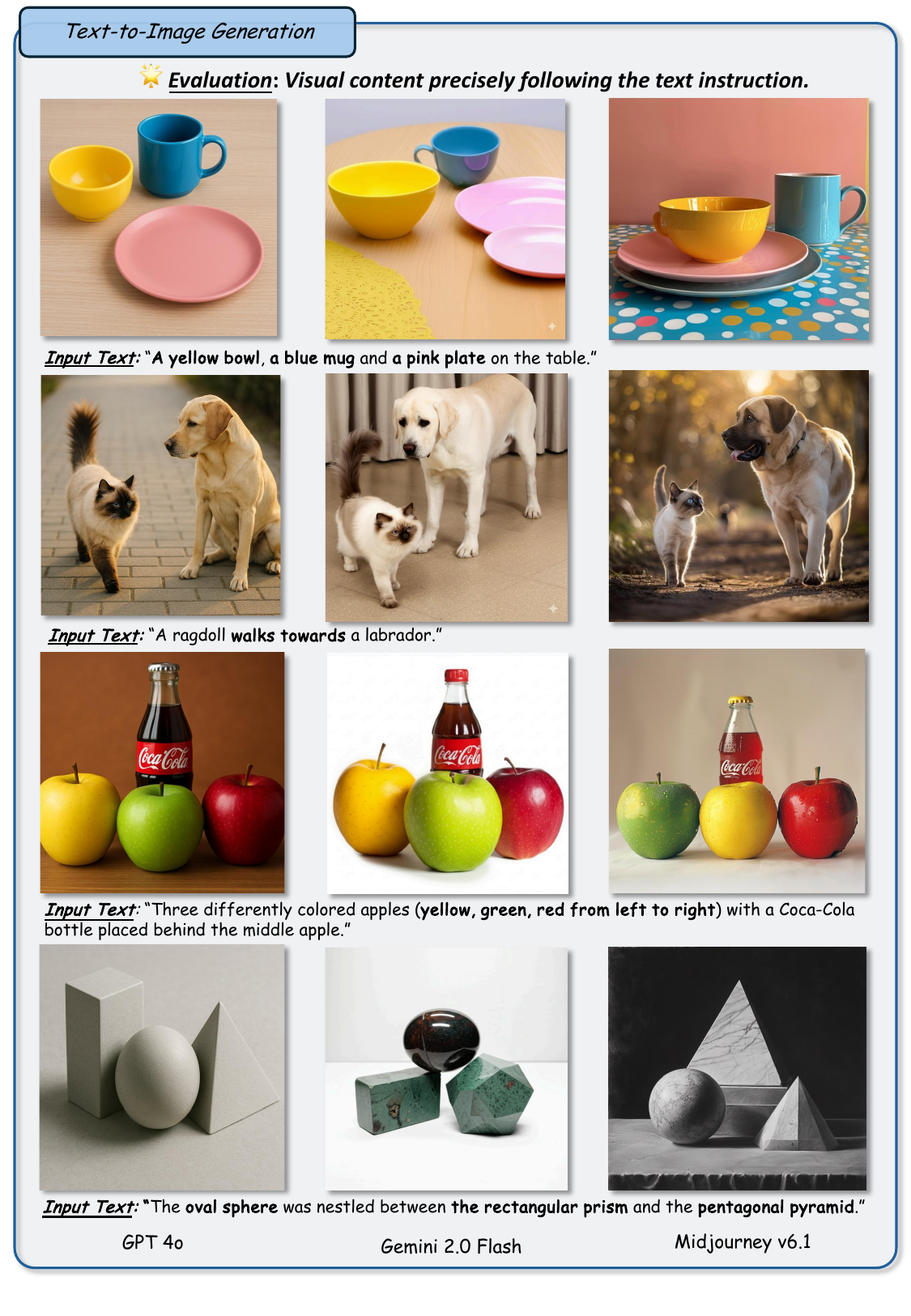}
    \caption{\textit{\textbf{Task:}} Compositional text-to-image generation. Evaluate the image-text alignment on attribute binding, numeracy, and object relationship. \textit{\textbf{Setup:}} Each row shows a text prompt and the generated outputs from GPT-4o, Gemini 2.0 Flash~\cite{gemini-2-0-flash}, and Midjourney v6.1~\cite{Midjourney}. \textit{\textbf{Observation:}} GPT-4o outperforms Gemini 2.0 Flash and Midjourney v6.1 across all aspects. However, GPT-4o struggles with uncommon objects with a special geometry.}
    \label{fig:t2i_1}
\end{figure}

\begin{figure}[h]
    \centering
    \includegraphics[width=0.96\textwidth]{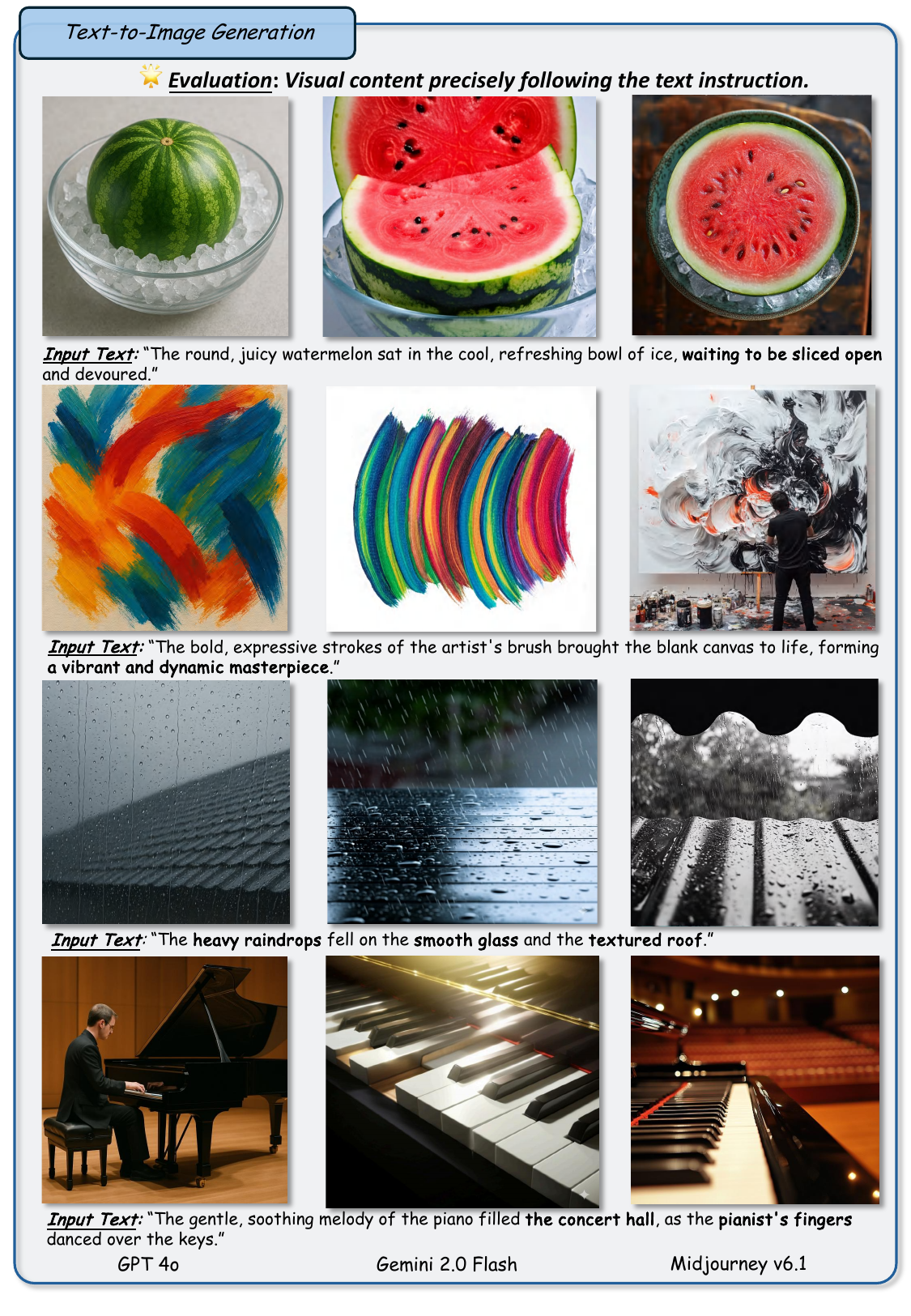}
    \caption{\textit{\textbf{Task:}} Compositional text-to-image generation. Evaluate the image-text alignment on attribute binding and complex compositions. \textit{\textbf{Setup:}} Each row shows a text prompt and the generated outputs from GPT-4o, Gemini 2.0 Flash~\cite{gemini-2-0-flash}, and Midjourney v6.1~\cite{Midjourney}. \textit{\textbf{Observation:}} GPT-4o outperforms the other two models in generating objects aligned with the text prompts accurately. But for more abstract and creative tasks, Midjourney v6.1 performs the best.}
    \label{fig:t2i_2}
\end{figure}

As shown in Figure~\ref{fig:t2i_1} row 1, GPT-4o outperforms both Gemini 2.0 Flash and Midjourney in numeracy tasks. While GPT-4o accurately represents a single plate, Gemini 2.0 and Midjourney represent two plates instead. In terms of understanding object relationships, GPT-4o is the only model that correctly infers the action ``walk towards'' from the ragdoll to the labrador. However, GPT-4o struggles with more complex terms like ``pentagonal pyramid'', failing to interpret it correctly (see Figure~\ref{fig:t2i_1} row 4). This suggests that GPT-4o may have difficulty accurately interpreting objects with unusual geometries. When it comes to abstract prompts, GPT-4o also appears to lack imagination (see Figure~\ref{fig:t2i_2} row 2), whereas Midjourney v6.1 demonstrates better creativity in this case, outperforming both GPT-4o and Gemini 2.0 Flash.

\begin{figure}[h]
    \centering
    \includegraphics[width=1\textwidth]{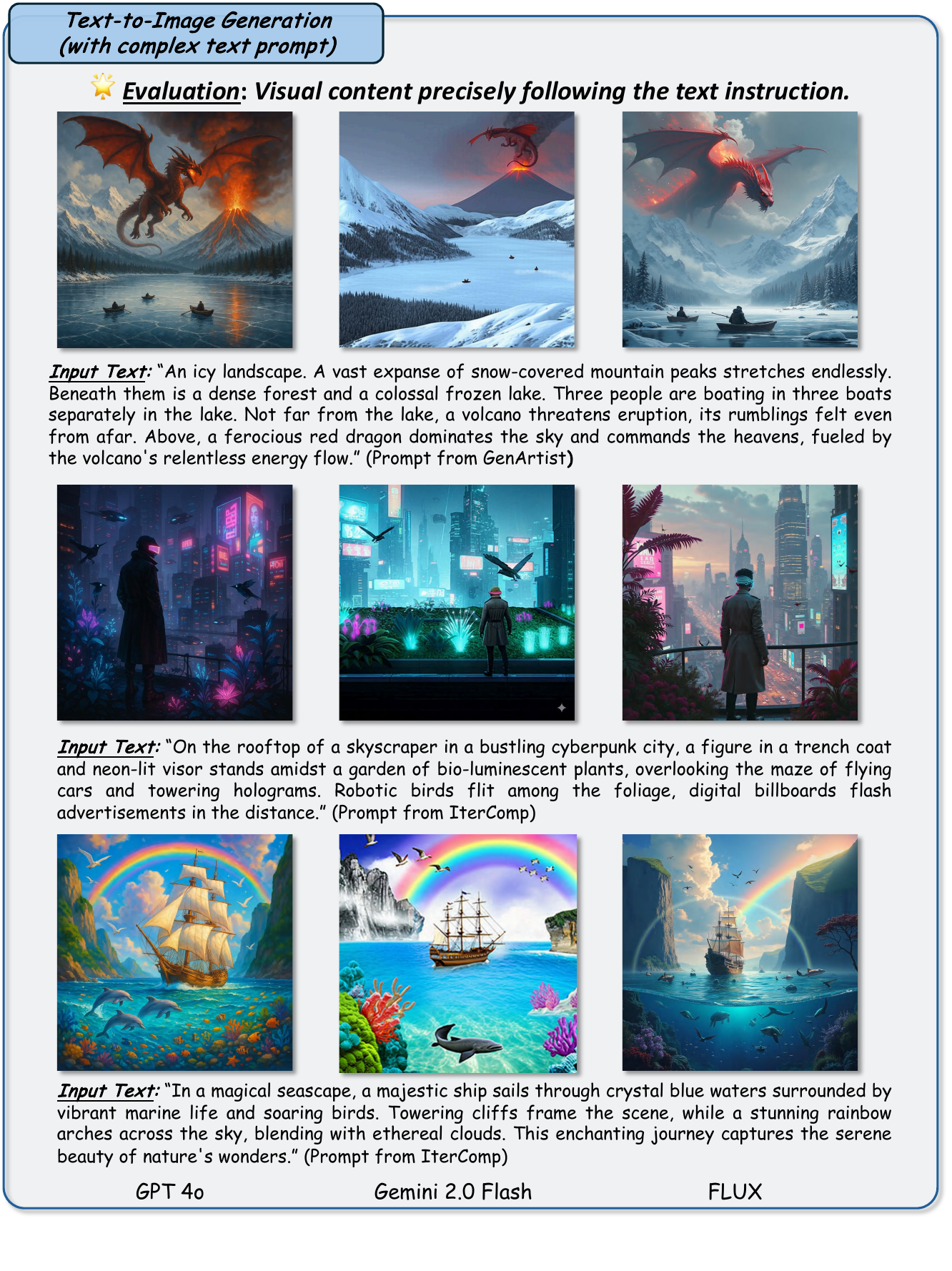}
    \caption{\textit{\textbf{Task:}} Compositional text-to-image generation. Evaluate the image-text alignment on complex compositions. \textit{\textbf{Setup:}} Each row shows a text prompt and the generated outputs from GPT-4o, Gemini 2.0 Flash~\cite{gemini-2-0-flash}, and FLUX.1-Pro~\cite{flux}. \textit{\textbf{Observation:}} GPT-4o and FLUX can generate more harmonious and natural scene than Gemini 2.0 Flash.}
    \label{fig:t2i_3}
\end{figure}

\begin{figure}[h]
    \centering
    \includegraphics[width=1\textwidth]{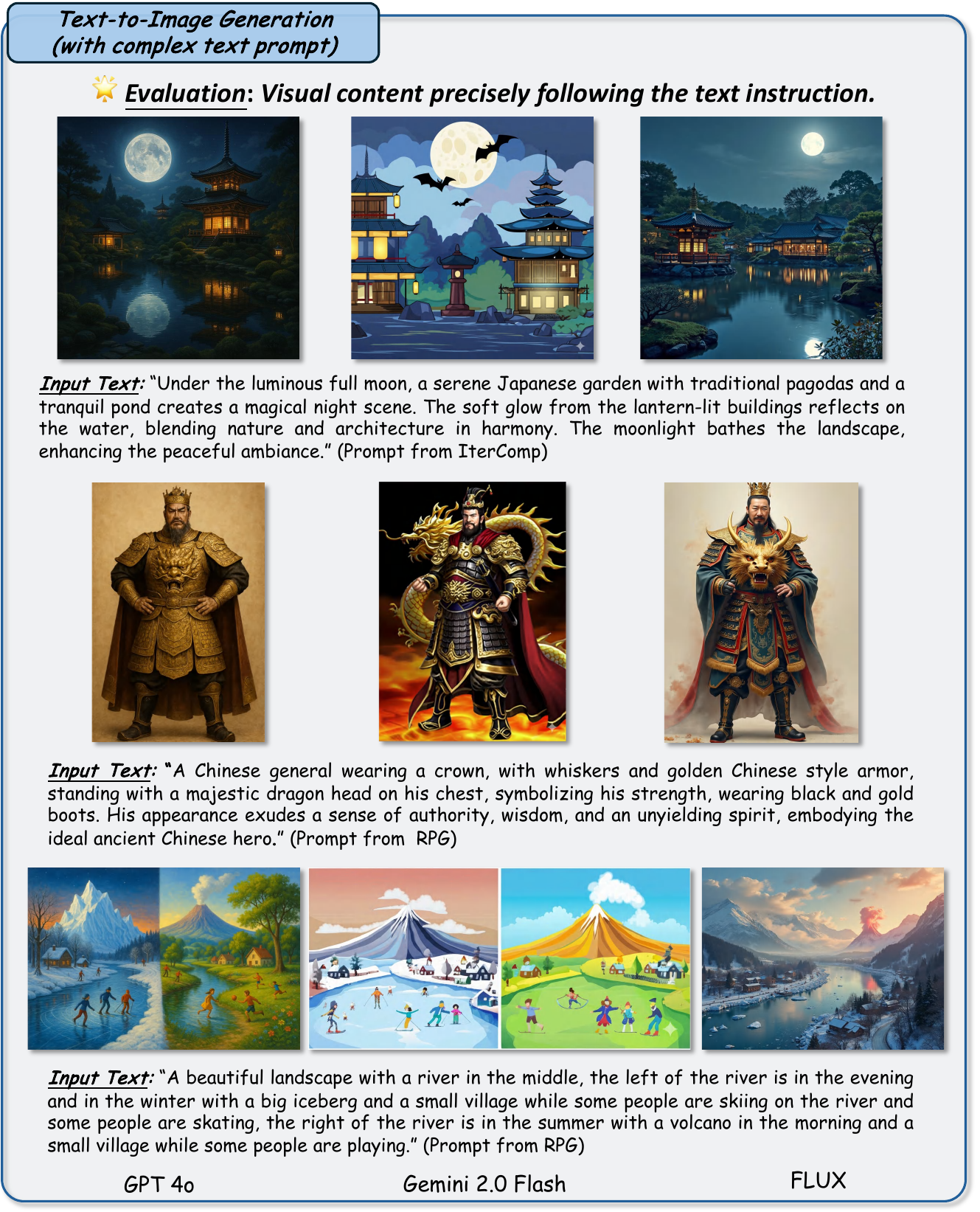}
    \caption{\textit{\textbf{Task:}} Compositional text-to-image generation. Evaluate the image-text alignment on complex compositions. \textit{\textbf{Setup:}} Each row shows a text prompt and the generated outputs from GPT-4o, Gemini 2.0 Flash~\cite{gemini-2-0-flash}, and FLUX.1-Pro~\cite{flux}. \textit{\textbf{Observation:}} GPT-4o struggles to generate culturally related elements and maintain boundary continuity (see rows 2 and 3), similar to Gemini 2.0 Flash and FLUX.}
    \label{fig:t2i_4}
\end{figure}

For complex text-to-image generation, we evaluate GPT-4o's performance with Gemini 2.0 Flash~\cite{gemini-2-0-flash} and FLUX.1-Pro~\cite{flux}, using the text prompts collected from~\cite{zhang2024itercomp, wang2024genartist, yang2024mastering}. As shown in Figure~\ref{fig:t2i_3}, both GPT-4o and FLUX excel at generating realistic and harmonious scenes align with the text prompts. However, we observe that GPT-4o shows limitations in generating culturally related elements. For example, the generated crown for the Chinese general is western-style rather than chinese-style (see Figure~\ref{fig:t2i_4} row 2). Additionally, in large scene generation, GPT-4o struggles to maintain boundary continuity, whereas FLUX produces a more natural composition (see Figure~\ref{fig:t2i_4} row 3).

Overall, we conclude that GPT-4o excels at text-to-image generation in terms of attribute binding, generative numeracy, object relationship, and complex compositions. However, it exhibits limitations in generating uncommon objects, culturally specific elements and in maintaining continuity when composing large scenes.

\clearpage


\subsubsection{Text Rendering}
Text rendering is a task that aims at generating texts (characters, sentences, or even paragraphs) on an image. The text content is usually guided by the input prompt. Previous models~\cite{sd3, ideogram-3-0} show good capability in generating short text (within 10 words, such as signs or short phrases), but their ability to generate long texts remains limited.

\begin{figure}[h]
    \centering
    \includegraphics[width=1\textwidth]{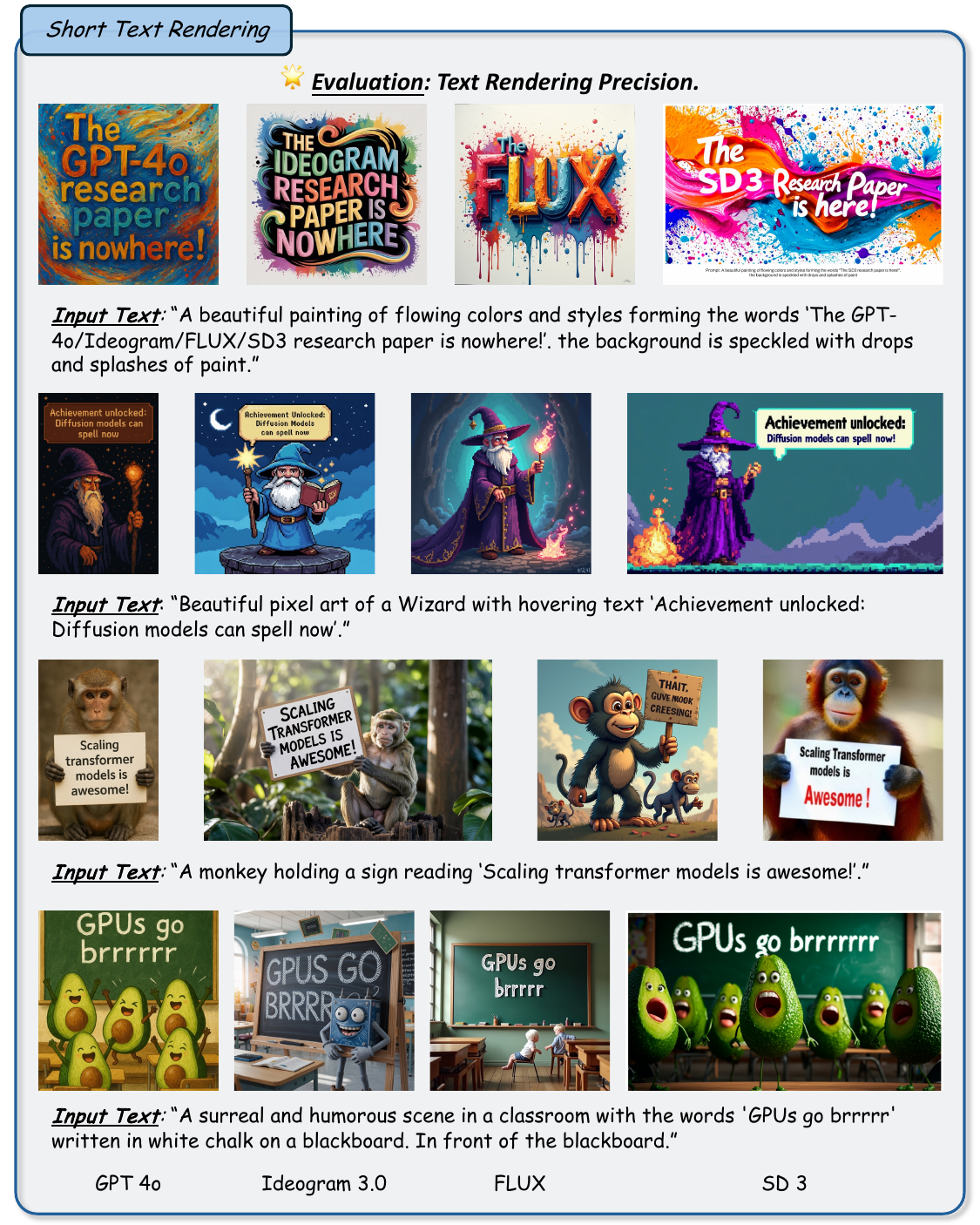}
    \caption{\textit{\textbf{Task:}} Short text rendering. Generate prompt-aligned, concise textual content (typically within 10 words) on an image.
\textit{\textbf{Setup:}} Each sample is produced based on a guiding text prompt. Comparisons are made with prior SOTA models~\cite{sd3, ideogram-3-0} and FLUX~\cite{flux}.
\textit{\textbf{Observations:}} GPT-4o achieves performance on par with existing SOTA baselines in rendering short texts, consistently following the prompt with minimal errors. All evaluated methods—except FLUX~\cite{flux}—deliver high-fidelity results in this setting.}
    \label{fig:text_short1}
\end{figure}

As shown in Figure~\ref{fig:text_short1}, GPT-4o demonstrates comparable abilities to existing state-of-the-art (SOTA) baselines when generating short texts. All the methods except FLUX~\cite{flux} perform well at rendering short text following the prompt. In this section, we primarily focus on long text rendering to examine whether GPT-4o can surpass these baselines for extended textual content.

\begin{figure}[h]
    \centering
    \includegraphics[width=0.92\textwidth]{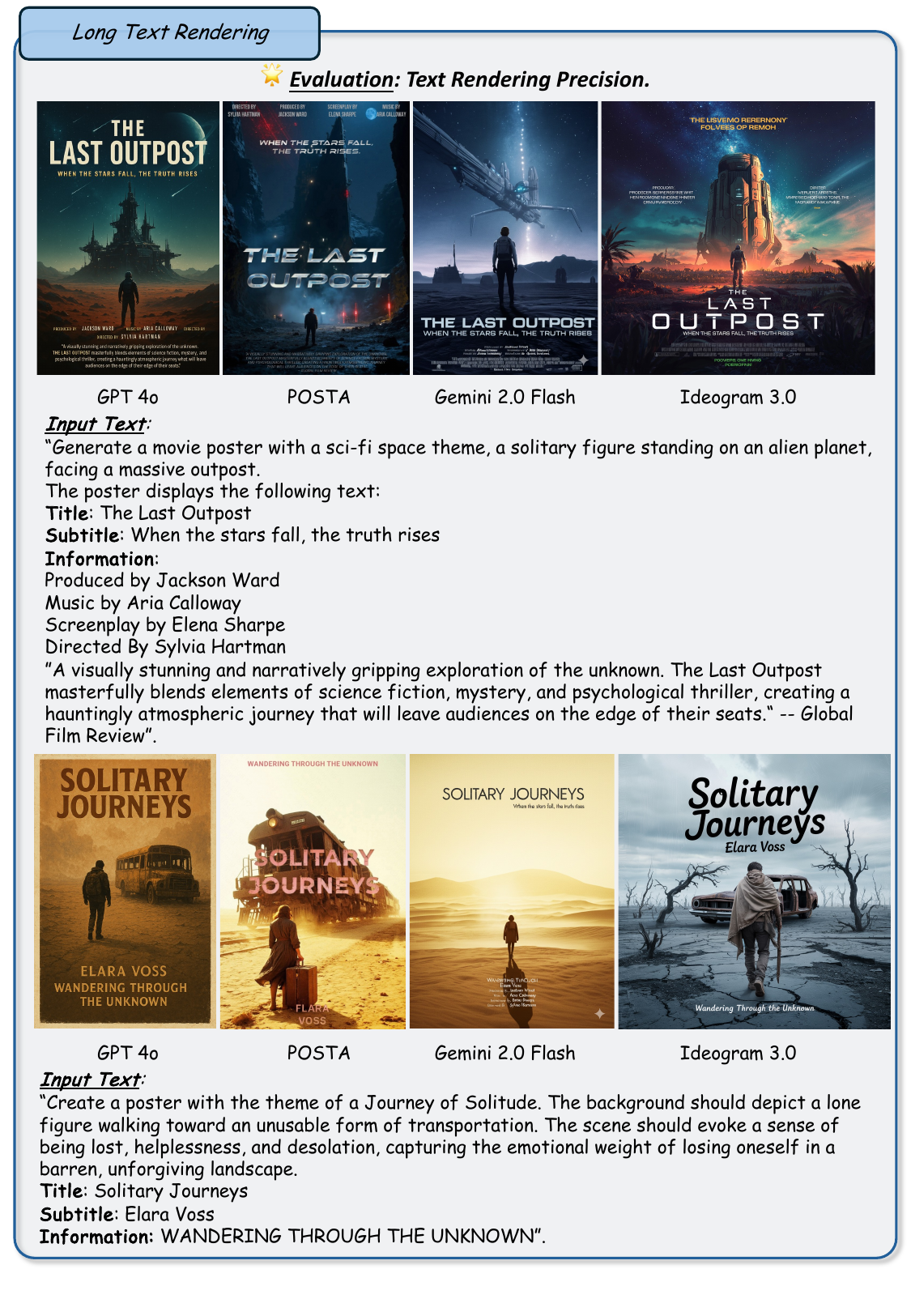}
    \caption{
\textit{\textbf{Task:}} Long text rendering. Generate extended, coherent, and prompt-consistent textual content on an image.
\textit{\textbf{Setup:}} Evaluations are conducted against advanced baselines including POSTA~\cite{posta}, Gemini 2.0 Flash~\cite{gemini-2-0-flash}, Ideogram 3.0~\cite{ideogram-3-0}, and Playground-v3~\cite{playground-v3}.
\textit{\textbf{Observations:}} GPT-4o excels in long text rendering by producing coherent, detailed textual information with very few character errors. In contrast, models like Gemini 2.0 Flash, Ideogram 3.0, and Playground-v3 often exhibit increased errors or generate vague text when faced with lengthy prompts, while POSTA's tailored multi-step pipeline sometimes yields competitive precision. Overall, GPT-4o outperforms most commercial models and rivals specialized research approaches in extended text generation.}
    \label{fig:text_long1}
\end{figure}

\begin{figure}[h]
    \centering
    \includegraphics[width=0.95\textwidth]{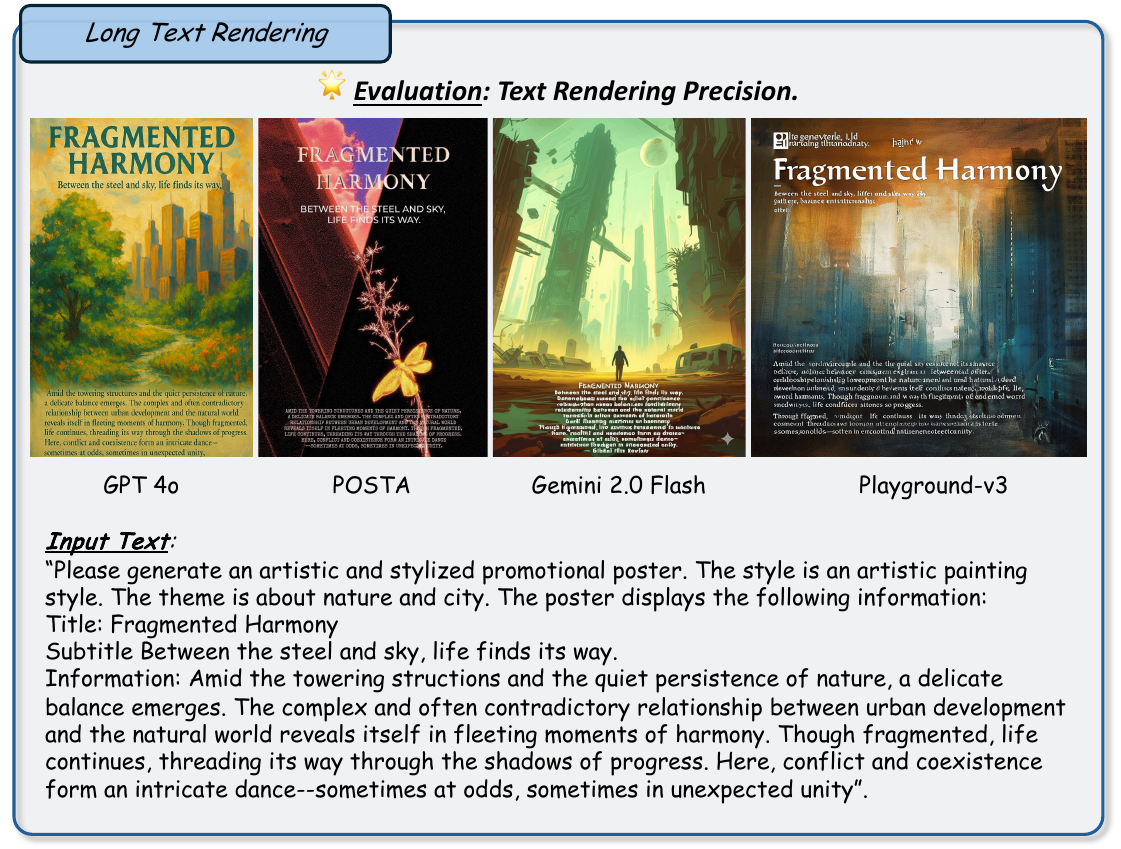}
    \caption{\textit{\textbf{Task:}} Long text rendering. The \textit{\textbf{Setup}} and \textit{\textbf{Observations}} are the same as Figure~\ref{fig:text_long1}.}
    \label{fig:text_long2}
\end{figure}

We choose POSTA~\cite{posta}, Gemini 2.0 Flash~\cite{gemini-2-0-flash}, Ideogram 3.0~\cite{ideogram-3-0}, and Playground-v3~\cite{playground-v3} as the baselines because of their established capabilities in rendering longer texts. The results are shown in Figure~\ref{fig:text_long1} and Figure~\ref{fig:text_long2}.

From these examples, we make the following key observations: 
\noindent
\begin{itemize} 
\item \textbf{GPT-4o’s strength in long text generation:} Compared with other baselines, GPT-4o demonstrates a superior ability to generate long, coherent text. In example 1 and example 3, GPT-4o produces detailed textual information with fewer than three characters generated incorrectly across more than 100 characters of text. 
\item \textbf{Baseline limitations:} When the input prompt becomes extremely long, models such as Gemini 2.0 Flash, Ideogram 3.0, and Playground-v3 often produce significantly more errors or produce vague text patches that are difficult to recognize. 
\item \textbf{POSTA’s performance:} As a model specifically designed for poster-style text generation, POSTA performs closely to, or in some instances slightly more precisely than, GPT-4o. We hypothesize this is due to its multi-step pipeline tailored for long text rendering. 
\end{itemize}

Overall, we conclude that GPT-4o \textbf{excels at long text rendering}, offering overwhelming performance compared to most existing commercial models, and delivering results on par with the latest specialized research models.

\clearpage

\subsubsection{Document Generation}

\begin{figure}[h]
    \centering
    \includegraphics[width=1\textwidth]{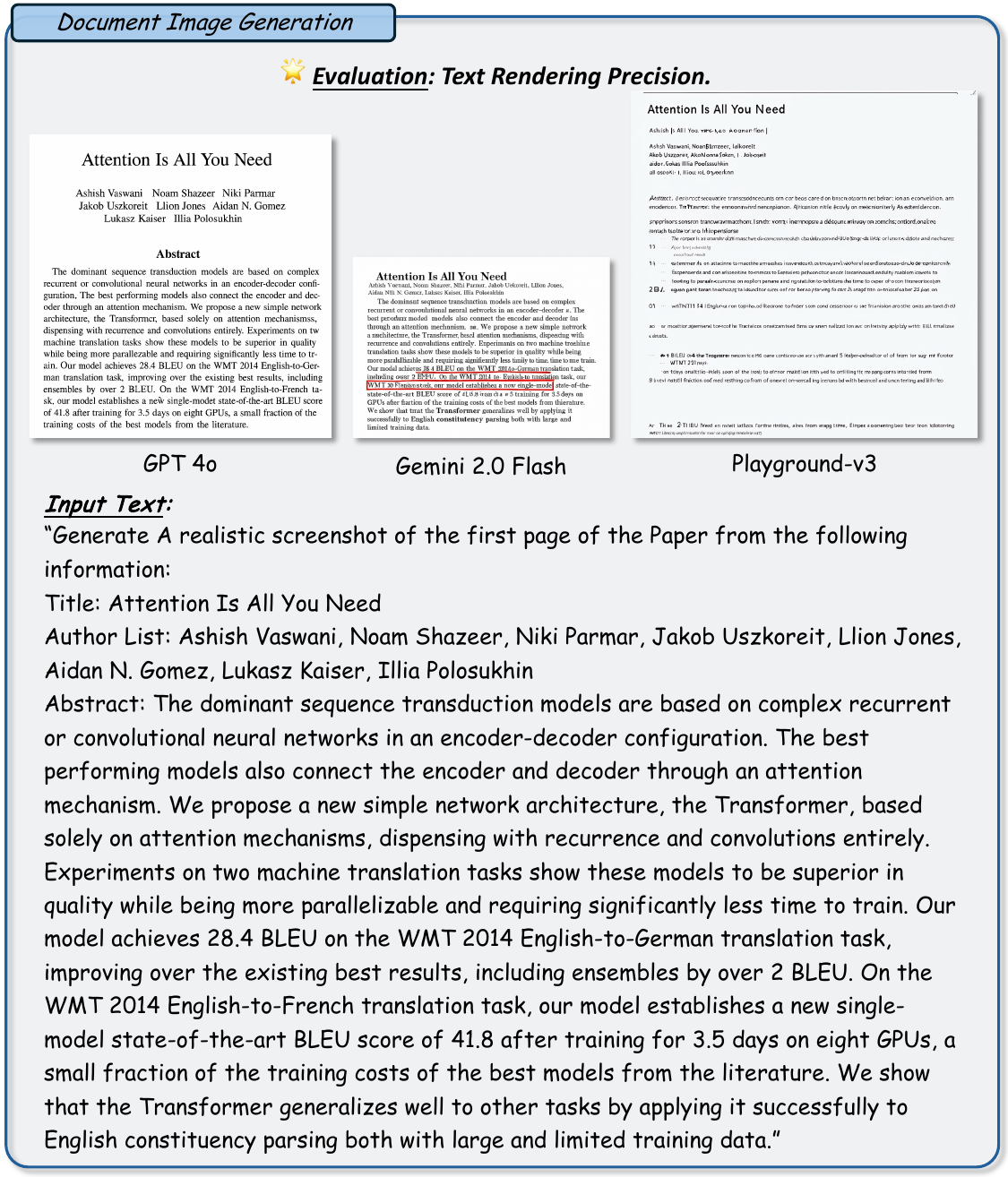}
    \caption{\textit{\textbf{Task:}} Document image generation. \textit{\textbf{Setup:}} Each row shows a text prompt and the generated outputs from GPT-4o, Gemini 2.0 Flash~\cite{gemini-2-0-flash}, and Playground-v3~\cite{playground-v3}. \textit{\textbf{Observation:}} GPT-4o can generate more consistent and accurate font and format than the other two models.}
    \label{fig:doc_1}
\end{figure}

\begin{figure}[h]
    \centering
    \includegraphics[width=1\textwidth]{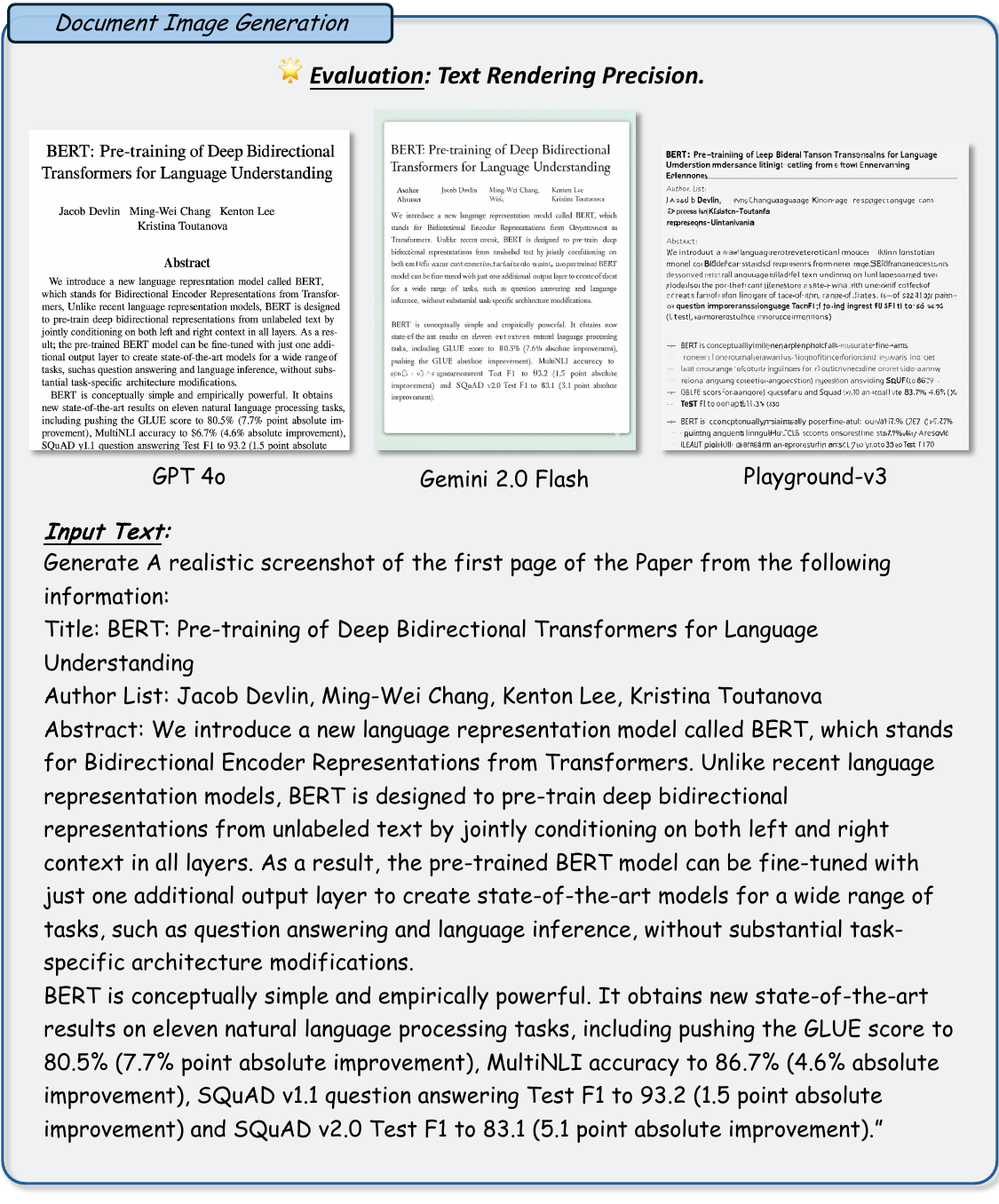}
    \caption{\textit{\textbf{Task:}} Document image generation. The \textit{\textbf{Setup}} and \textit{\textbf{Observations}} are the same as Fig.~\ref{fig:doc_1}.}
    \label{fig:doc_2}
\end{figure}

\begin{figure}[h]
    \centering
    \includegraphics[width=1\textwidth]{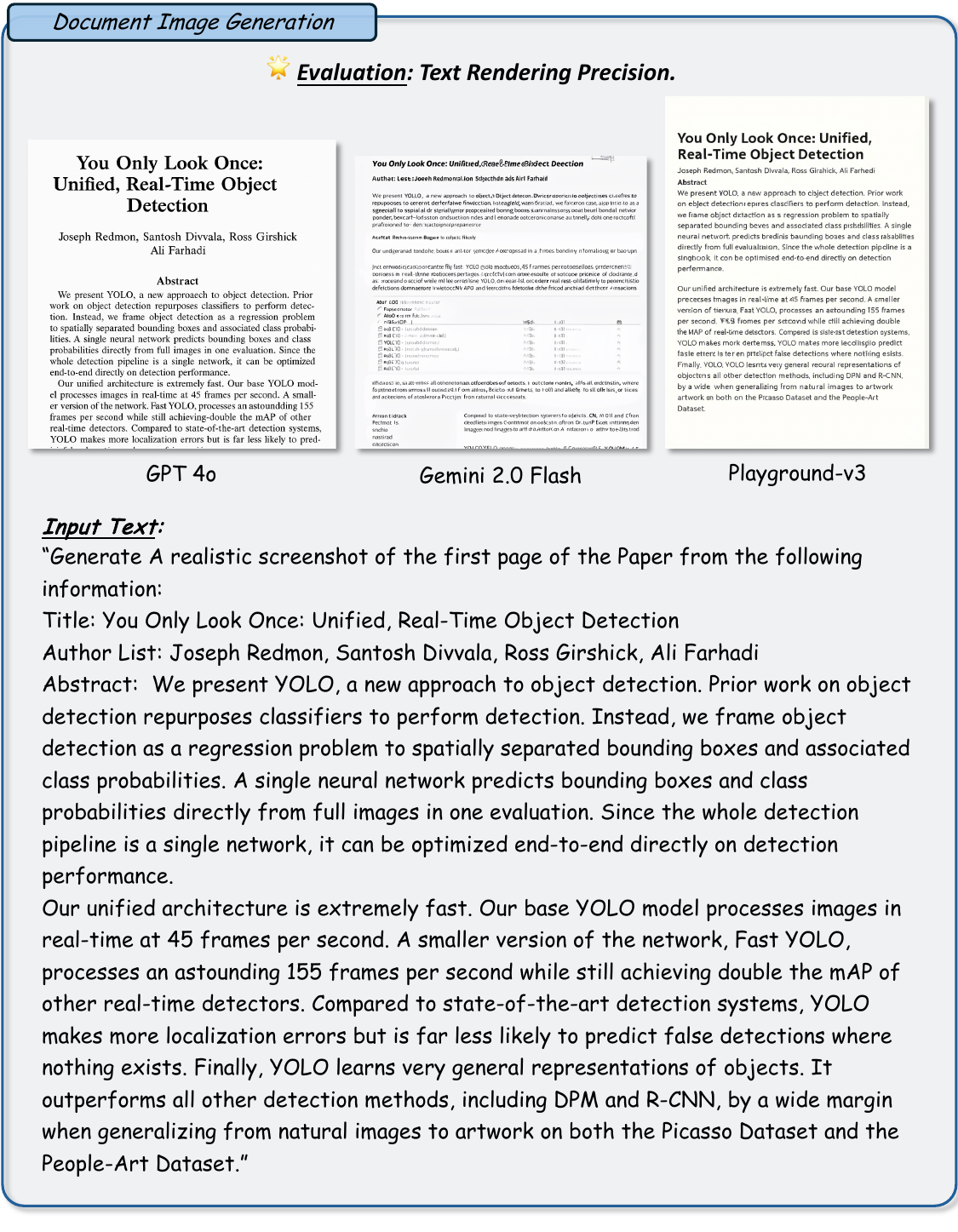}
    \caption{\textit{\textbf{Task:}} Document image generation. The \textit{\textbf{Setup}} and \textit{\textbf{Observations}} are the same as Fig.~\ref{fig:doc_1}.}
    \label{fig:doc_3}
\end{figure}

We also explore a novel task: document image generation with GPT-4o, comparing its performance with Gemini 2.0 Flash~\cite{gemini-2-0-flash} and Playground-v3~\cite{playground-v3}. As shown in Figure~\ref{fig:doc_1} -~\ref{fig:doc_3}, GPT-4o produces document images with cleaner layouts and more consistent content.

\clearpage

\subsubsection{Panorama Image Generation}

\begin{figure}[h]
    \centering
    \includegraphics[width=1\textwidth]{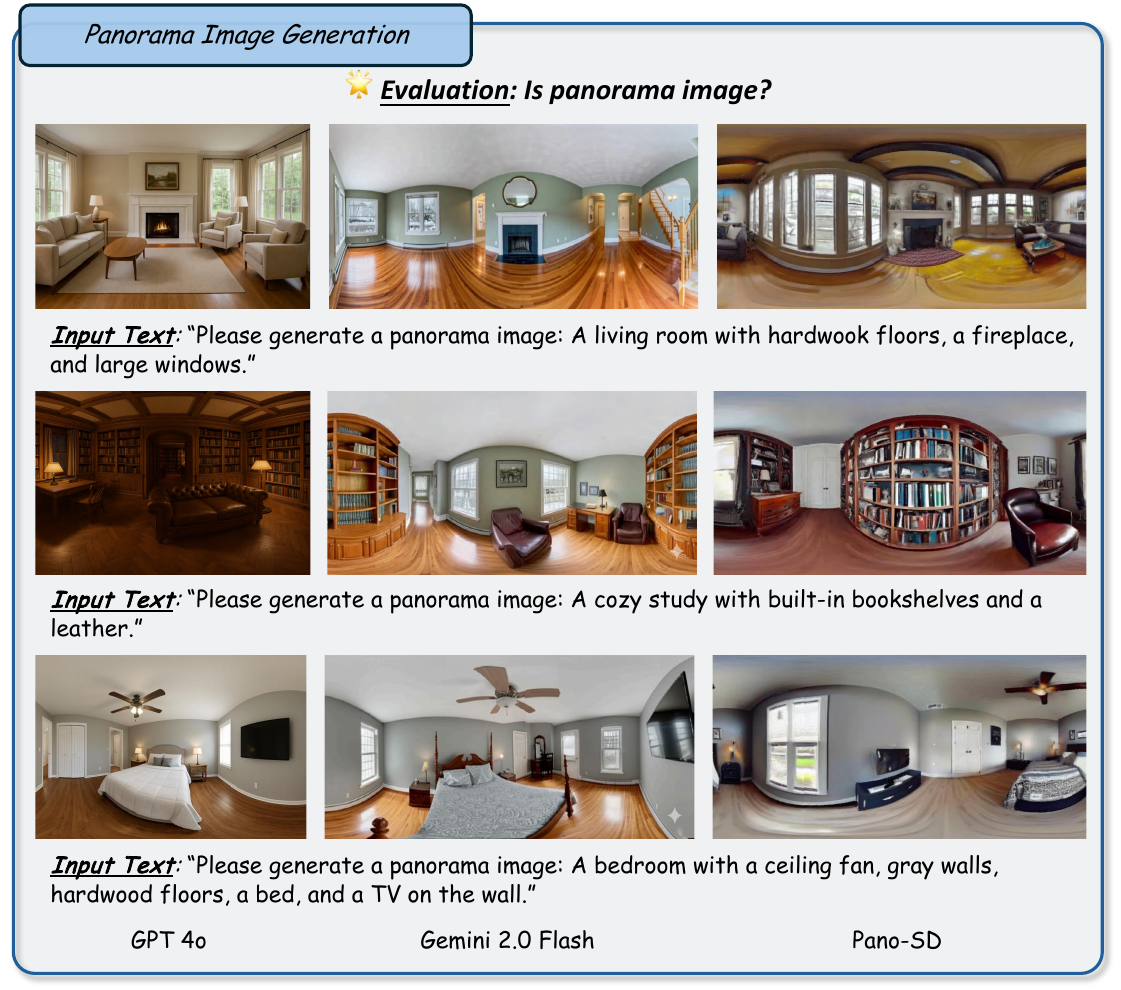}
    \caption{
\textit{\textbf{Task:}} Panorama image generation, aiming to create immersive 360-degree views of static scenes.
\textit{\textbf{Setup:}} We compare GPT-4o with established baselines such as Pano-SD~\cite{pano-sd} and Gemini 2.0 Flash~\cite{gemini-2-0-flash} to evaluate the generation of coherent panoramic images.
\textit{\textbf{Observations:}} While the baseline models reliably produce panoramas with seamlessly connected left and right sides, GPT-4o tends to only approximate a panoramic view and struggles to maintain continuity across image boundaries. This shortfall is likely due to limited panorama image representation in its training data and a tendency to generate images with a higher vertical aspect ratio rather than a wider one, rendering it inferior to the baselines in this task.}
    \label{fig:panorama_1}
\end{figure}

Panorama image generation aims at creating a 360-degree view of a static scene, enabling immersive and comprehensive visual experiences. In our experiments, we select Pano-SD~\cite{pano-sd} and Gemini 2.0 Flash~\cite{gemini-2-0-flash} as the baselines, with representative results illustrated in Figure~\ref{fig:panorama_1}. The comparisons reveal that while the baseline models can generate coherent panorama-like images with seamlessly connectable left and right sides, GPT-4o struggles to produce a true panorama. In most cases, GPT-4o generates images that approximate a panoramic view but still fall short in ensuring the necessary continuity across the image boundaries. We attribute this limitation to the insufficient representation of panorama images in its training data, as well as a predisposition towards generating images with a higher vertical aspect ratio rather than a wider one. Consequently, in the realm of panorama image generation, GPT-4o is inferior to the existing baseline models.

\clearpage

\subsection{Image-to-Image Tasks}

\subsubsection{Style Transfer}
Style transfer is a classic yet evolving task in computer vision, aiming to render an image in a specific artistic style while preserving the original content. It bridges the domains of vision and art, enabling applications such as digital artwork creation, film post-production, and virtual reality environment design. Early approach~\cite{gatys2016image} used convolutional neural networks to separate and recombine content and style representations from images. This seminal work enabled the artistic stylization of photographs by optimizing pixel values to match a desired style. To improve efficiency, Johnson et al.~\cite{johnson2016perceptual} proposed feed-forward networks for real-time style transfer using perceptual losses. Later methods such as AdaIN~\cite{huang2017arbitrary} and WCT~\cite{li2017universal} enabled arbitrary style transfer without retraining for each new style. Transformer-based models like StyTr$^2$~\cite{deng2022stytr2} have been introduced to enhance style transfer quality and better preserve structural details. More recently, with the rapid development of image synthesis techniques, especially diffusion models, style transfer has seen further advancements in both quality and controllability. However, transferring specific artistic styles still typically requires a non-trivial amount of training data.

\begin{figure}[h]
    \centering
    \includegraphics[width=1\textwidth]{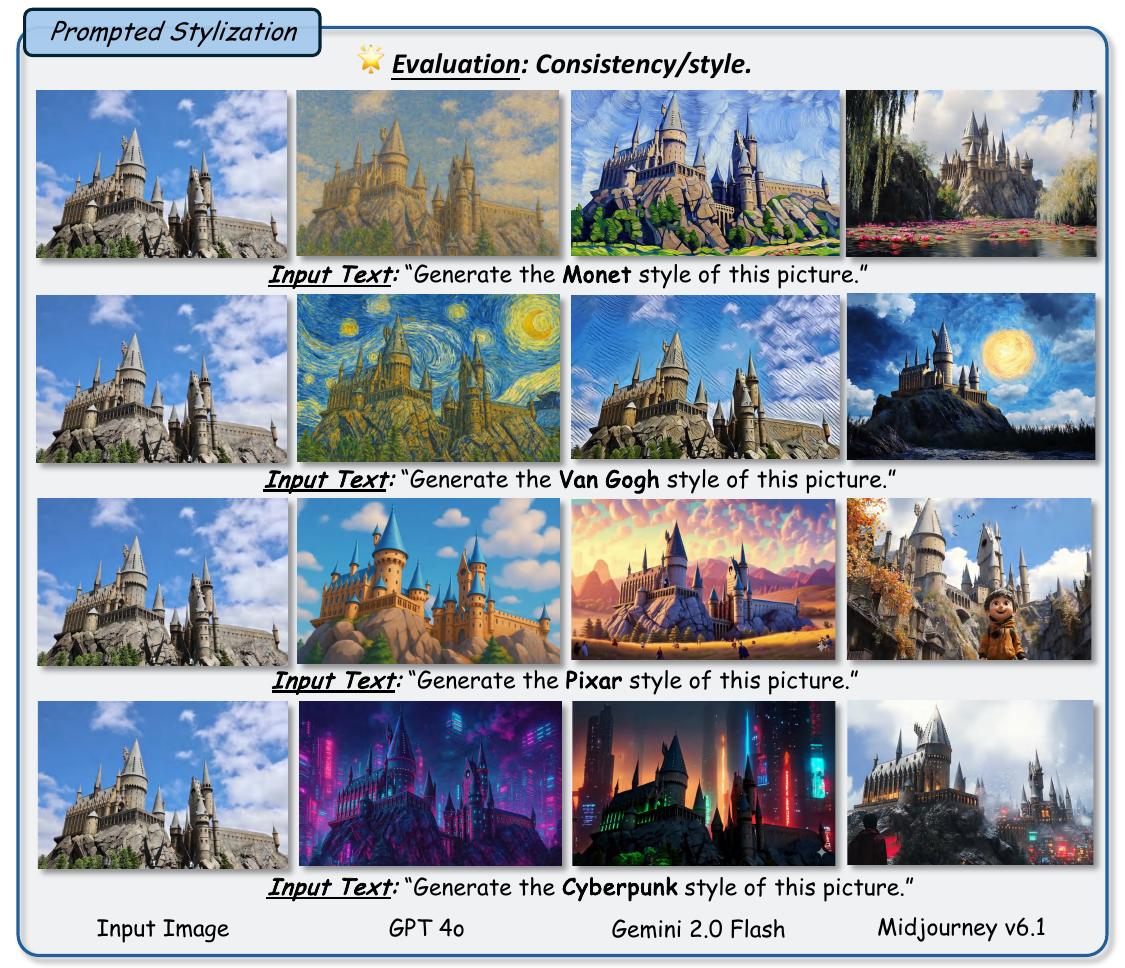}
    \caption{\textbf{Task:} Style transfer, aiming to render an image in a specific artistic style while preserving the original content.
\textit{\textbf{Setup:}} We compare GPT-4o with Gemini 2.0 Flash~\cite{gemini-2-0-flash} and Midjourney v6.1~\cite{Midjourney} on natural scene style transfer across multiple artistic domains.
\textit{\textbf{Observations:}} GPT-4o exhibits significantly better content preservation compared to Midjourney v6.1, maintaining fine-grained content details and structural consistency. In terms of style, it faithfully adheres to the textual description, effectively rendering vivid color palettes and soft contours that characterize the target style. This alignment notably surpasses both Gemini 2.0 Flash and Midjourney v6.1, highlighting GPT-4o’s strong capabilities in preserving content and faithfully rendering diverse styles. }
    \label{fig:style_transfer_1}
\end{figure}

\begin{figure}[h]
    \centering
    \includegraphics[width=0.88\textwidth]{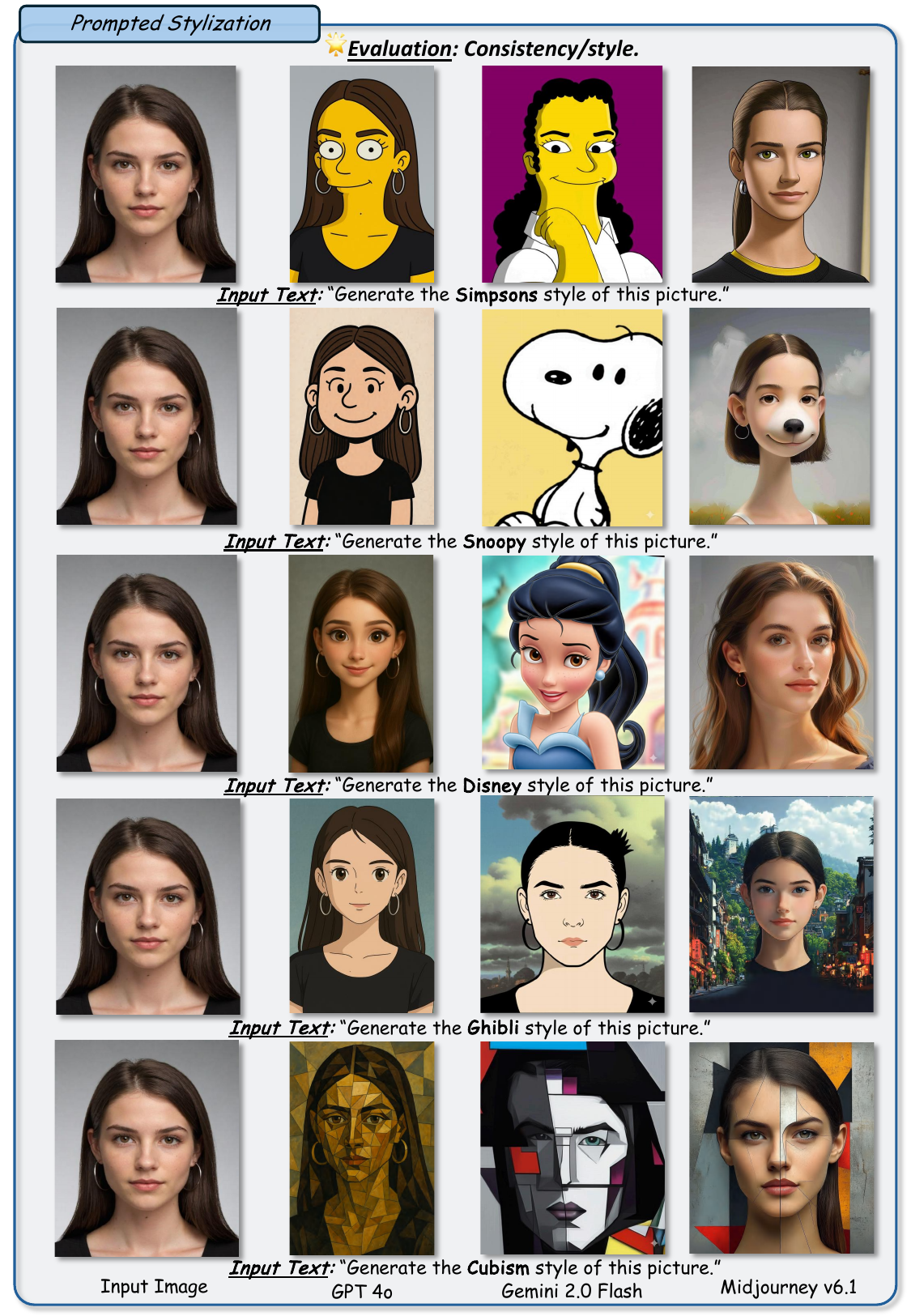}
    \caption{\textbf{Task:} Style transfer, aiming to render an image in a specific artistic style while preserving the original content.
\textit{\textbf{Setup:}} We compare GPT-4o with Gemini 2.0 Flash~\cite{gemini-2-0-flash} and Midjourney v6.1~\cite{Midjourney} on human face style transfer across multiple artistic domains.
\textit{\textbf{Observations:}} GPT-4o exhibits significantly better content preservation compared to Gemini 2.0 Flash and Midjourney v6.1, maintaining fine-grained content details and structural consistency. In terms of style, it faithfully adheres to the textual description, effectively rendering vivid color palettes and soft contours that characterize the target style. This alignment notably surpasses both Gemini 2.0 Flash and Midjourney v6.1 far away, highlighting GPT-4o’s strong capabilities in preserving content and faithfully rendering diverse styles. }
    \label{fig:style_transfer_2}
\end{figure}

To comprehensively evaluate the style transfer capability of GPT-4o, we conduct comparisons against several recent competitive models, including Gemini 2.0 Flash~\cite{gemini-2-0-flash} and Midjourney v6.1~\cite{Midjourney}. Specifically, Figure~\ref{fig:style_transfer_1} illustrates style transfer results for natural scenes, while Figure~\ref{fig:style_transfer_2} focuses on human facial images. Across a diverse range of styles, such as Monet, Van Gogh, Pixar, Cyberpunk, Snoopy, Disney, Ghibli, and Cubism, GPT-4o demonstrates consistently superior performance in both stylistic fidelity and content preservation.

Notably, in the case of Ghibli style transfer, GPT-4o exhibits remarkable fidelity to the original artistic aesthetics, closely resembling the target style with vivid color palettes and soft contours. In contrast, both Gemini and Midjourney often produce inconsistent visual styles and textures. Furthermore, GPT-4o excels at preserving fine-grained content details, such as facial structure, earrings, clothing, and hairstyles, which are often misrepresented or lost in the outputs of other models. These results suggest that GPT-4o not only captures high-level style semantics but also maintains strong spatial consistency and semantic alignment.

\clearpage

\subsubsection{Image Editing}
Image editing involves modifying the visual elements, composition, or data of an image to achieve a desired outcome. This process can range from minor refinements to significant alterations, while maintaining the integrity of the original image. Over time, image editing techniques have evolved from manual, labor-intensive methods to sophisticated AI-driven approaches. Prior works~\cite{brooks2022instructpix2pix, fu2024mgie, brack2023ledits, Zhang2023MagicBrush, bai2023integrating, feng2024item, bai2024humanedit, huang2024dual} have demonstrated the ability to perform various editing tasks based on textual instructions, such as adding, removing, or replacing objects; altering backgrounds, colors, or styles; and adjusting the number, size, or positions of objects. However, these models still exhibit limitations in certain scenarios, particularly in preserving non-edited regions, maintaining consistent image characteristics, and ensuring seamless blending between edited and non-edited areas.

\begin{figure}[h]
    \centering
    \includegraphics[width=0.9\textwidth]{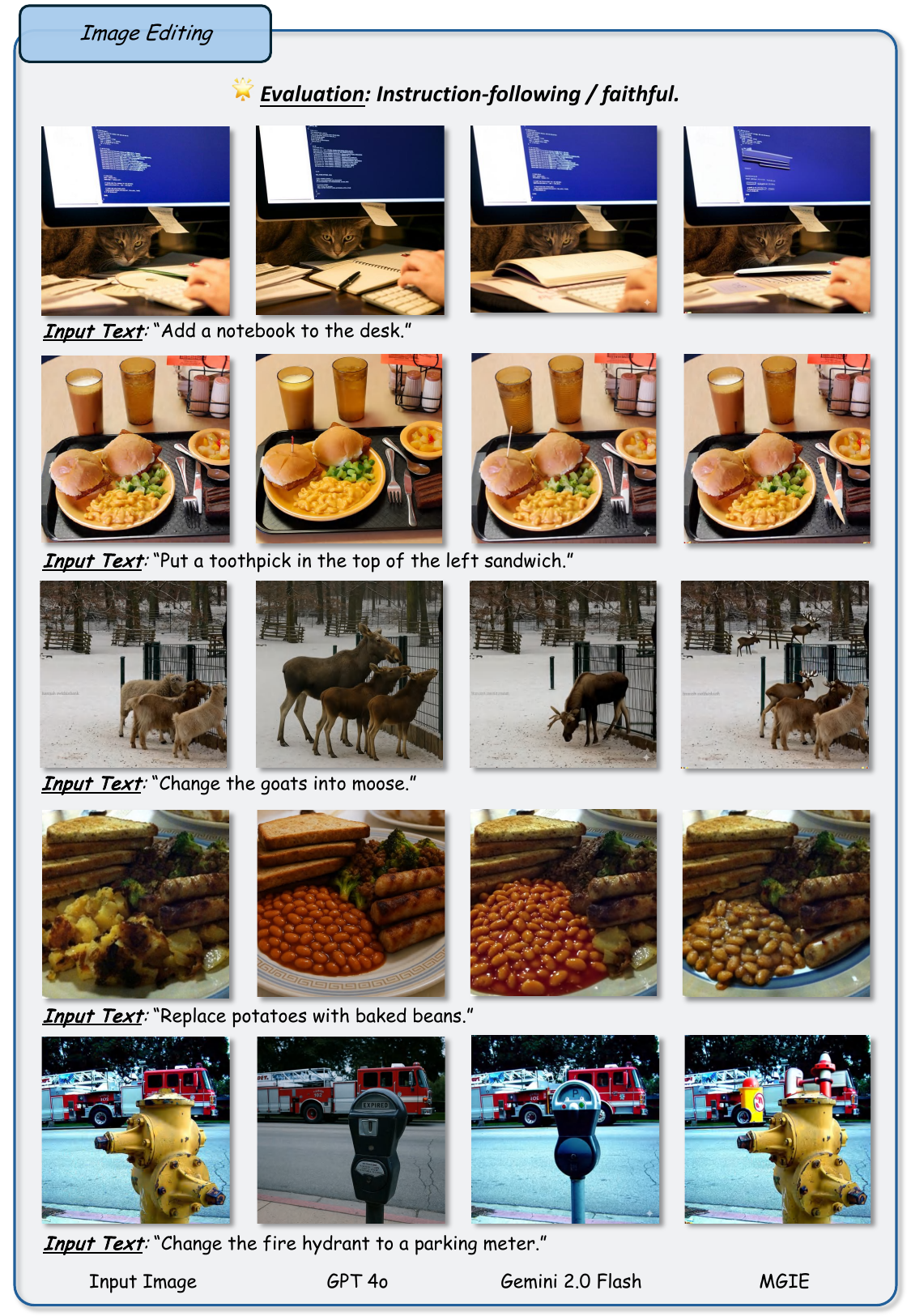}
    \caption{
\textit{\textbf{Task:}} Image editing for modifying visual elements and composition.
\textit{\textbf{Setup:}} GPT-4o vs. Gemini 2.0 Flash~\cite{gemini-2-0-flash}/MGIE~\cite{fu2024mgie}.
\textit{\textbf{Observations:}} GPT-4o achieves higher success rates than MGIE (examples 2/5) but occasionally alters unintended elements (bread in example 4) or lighting/shading structures (example 5). This likely stems from stronger generalization capacity and creative adaptation focus in training, though reduced fidelity suggests insufficient constraints on structural details during fine-tuning.
    }
    \label{fig:edit_1}
\end{figure}

\begin{figure}[h]
    \centering
    \includegraphics[width=0.9\textwidth]{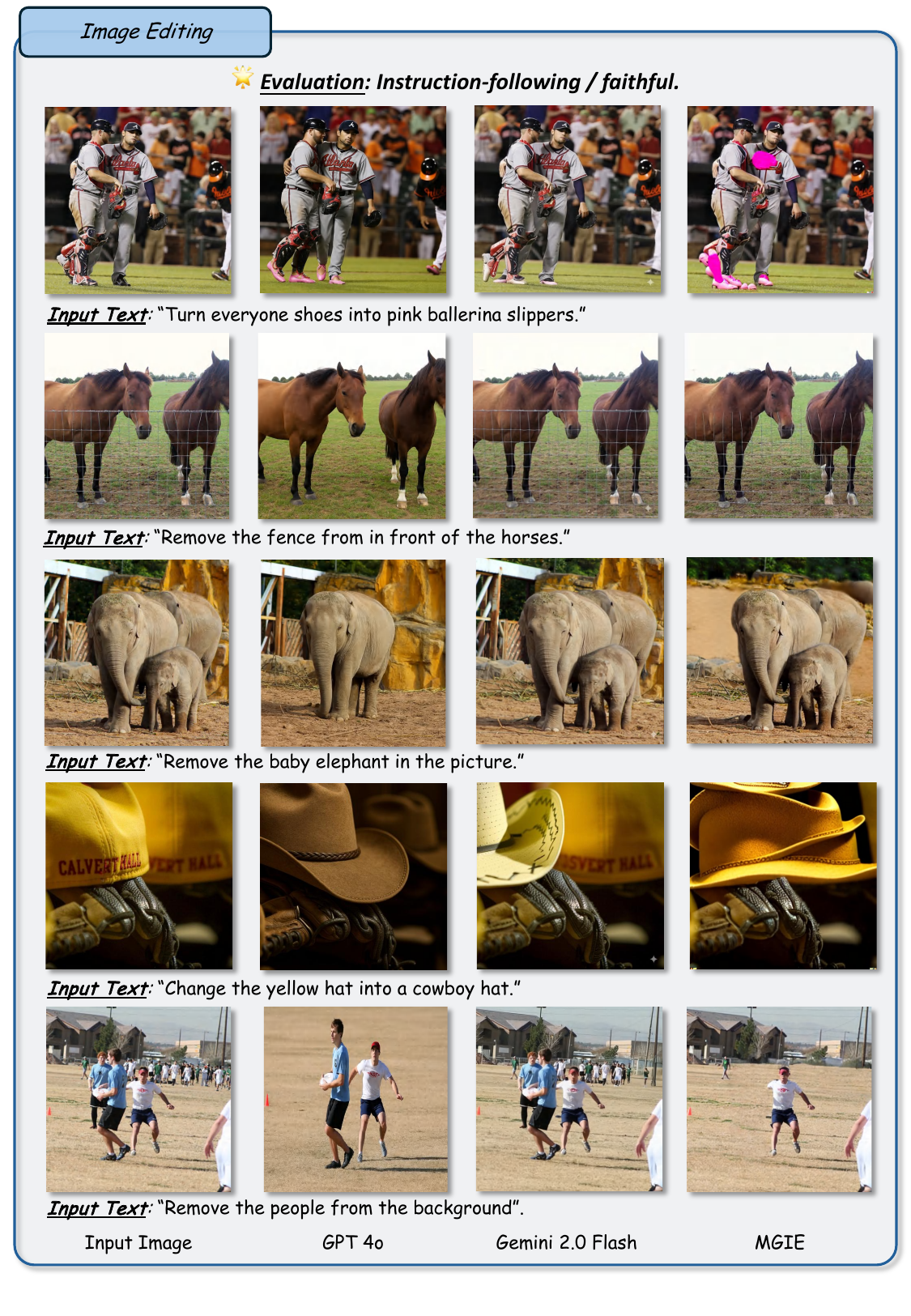}
    \caption{
\textit{\textbf{Task:}} Image editing for modifying visual elements and composition.
\textit{\textbf{Setup:}} GPT-4o vs. Gemini 2.0 Flash~\cite{gemini-2-0-flash}/MGIE~\cite{fu2024mgie}.
\textit{\textbf{Observations:}} From examples 1-3, GPT-4o shows higher success in fine detail edits and large-scale edits with occlusions. This likely stems from GPT-4o's stronger contextual understanding and ability to infer missing or obscured elements, enabling more precise localized edits and coherent large-scale modifications even with partial visibility. However, it sometimes erases non-target elements (e.g., the house in example 5) and significantly alters global lighting (example 4).
    }
    \label{fig:edit_2}
\end{figure}

\begin{figure}[h]
    \centering
    \includegraphics[width=0.9\textwidth]{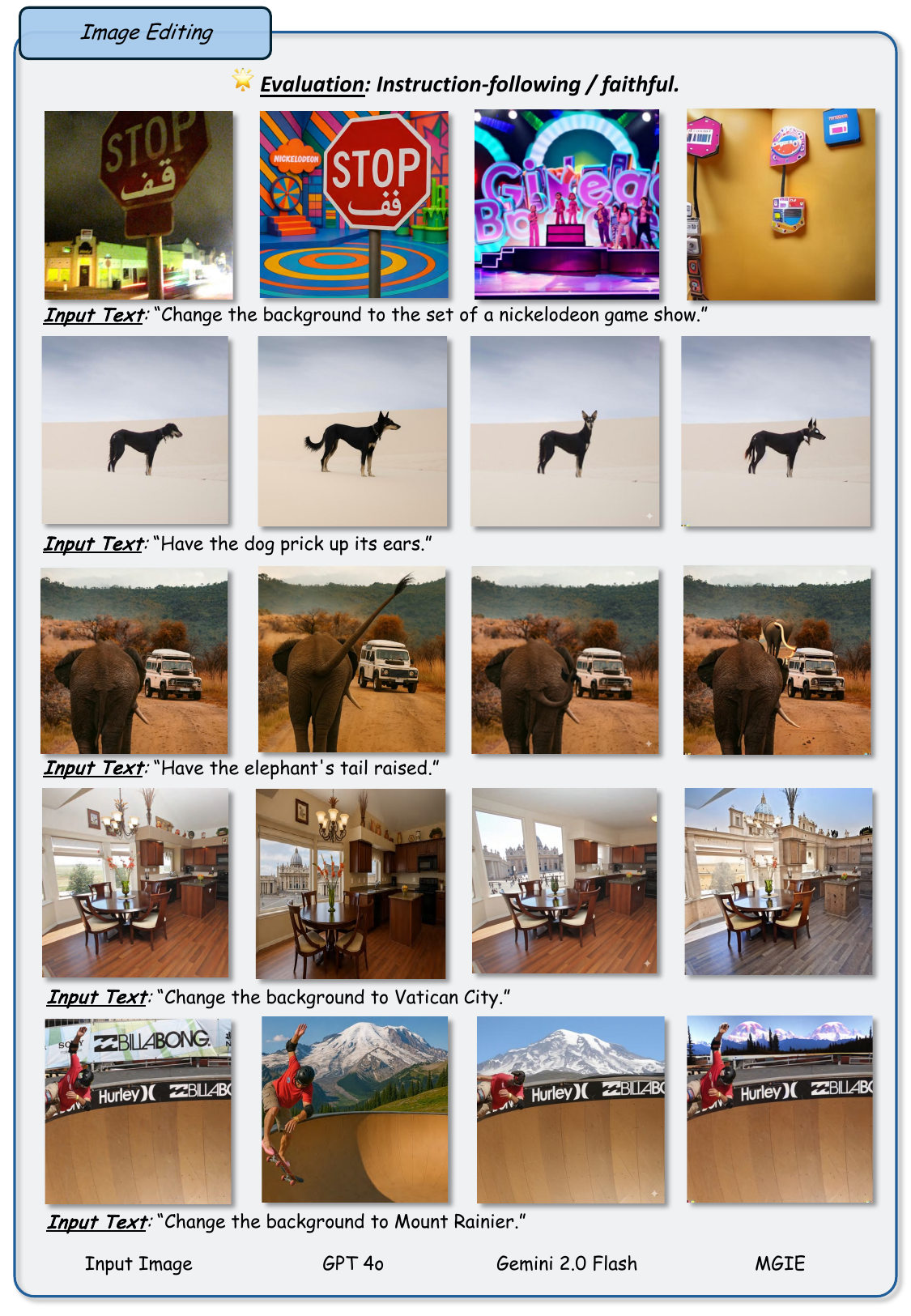}
    \caption{
\textit{\textbf{Task:}} Image editing for modifying visual elements and composition.
\textit{\textbf{Setup:}} GPT-4o vs. Gemini 2.0 Flash~\cite{gemini-2-0-flash}/MGIE~\cite{fu2024mgie}.
\textit{\textbf{Observations:}} From Example 1, GPT-4o demonstrates superior performance in style editing, effectively interpreting style instructions and preserving global image structure—a capability lacking in baseline models (MGIE, Gemini 2.0 Flash, and MagicBrush, as will be shown later). This likely stems from its stronger cross-modal comprehension and structural awareness during training.
    }
    \label{fig:edit_3}
\end{figure}

\begin{figure}[h]
    \centering
    \includegraphics[width=0.9\textwidth]{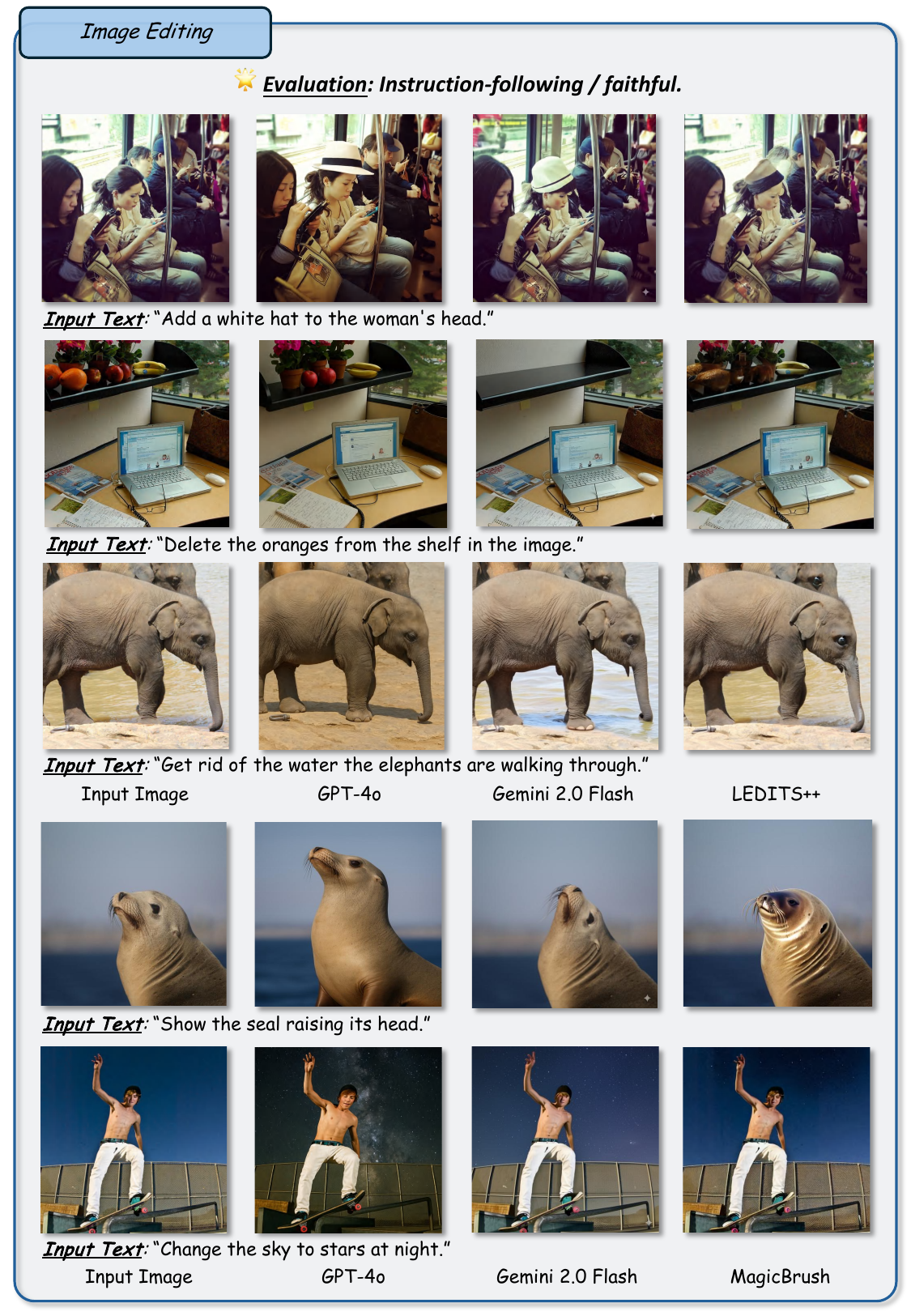}
    \caption{
\textit{\textbf{Task:}} Image editing for modifying visual elements and composition.
\textit{\textbf{Setup:}} GPT-4o vs. Gemini 2.0 Flash~\cite{gemini-2-0-flash}/LEDITS++~\cite{brack2023ledits}/MagicBrush~\cite{Zhang2023MagicBrush}.
\textit{\textbf{Observations:}} From Examples 2 and 3, GPT-4o demonstrates stronger comprehension of instructions involving `the oranges on the shelf' and `the water the elephants are walking through', translating this understanding into more accurate edits. This suggests better grounding of textual prompts in visual context during generation.
    }
    \label{fig:edit_4}
\end{figure}

\begin{figure}[h]
    \centering
    \includegraphics[width=0.9\textwidth]{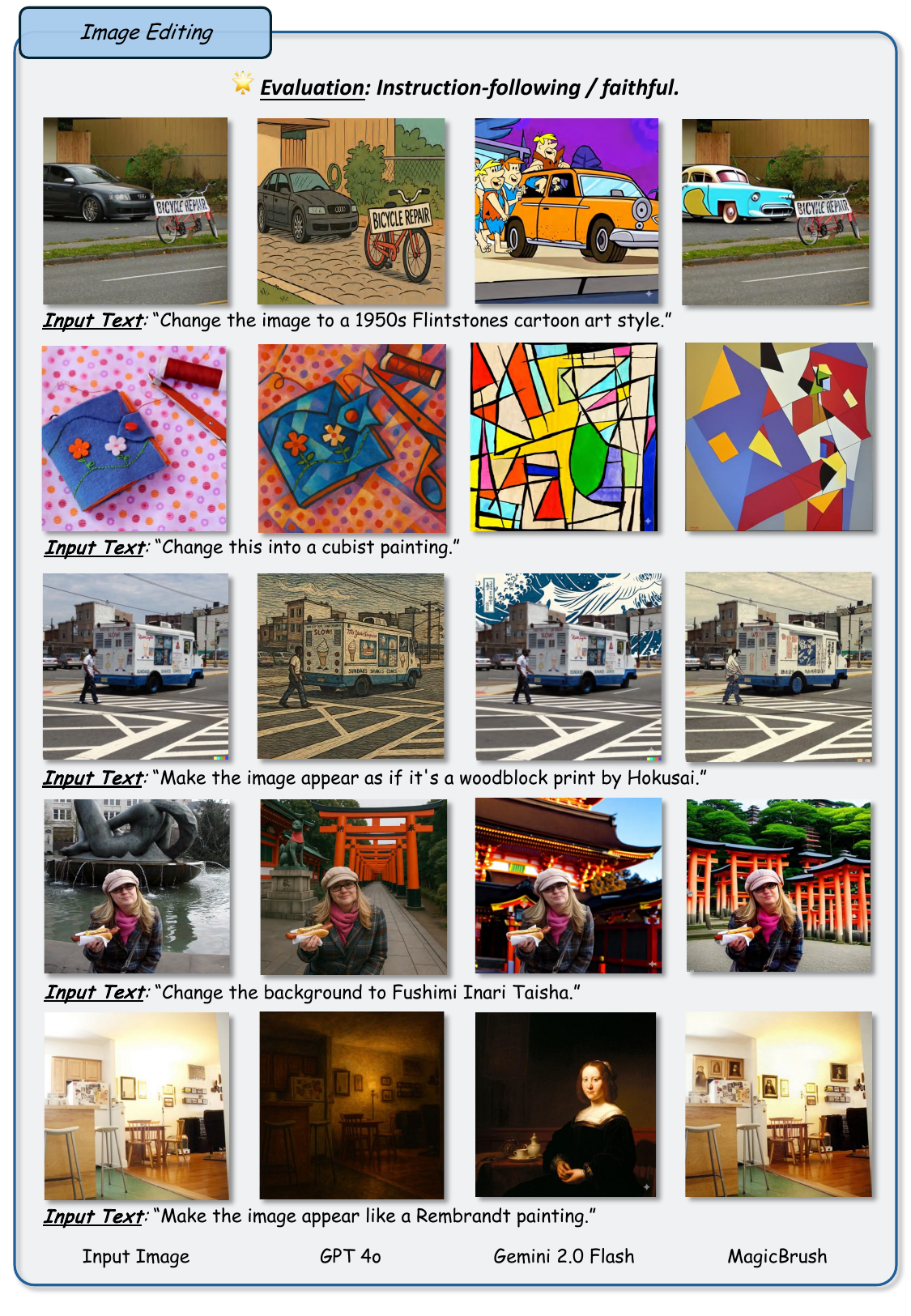}
    \caption{
\textit{\textbf{Task:}} Image editing for modifying visual elements and composition.
\textit{\textbf{Setup:}} GPT-4o vs. Gemini 2.0 Flash~\cite{gemini-2-0-flash}/MagicBrush~\cite{Zhang2023MagicBrush}.
\textit{\textbf{Observations:}} This set of examples further demonstrates GPT-4o's robust capabilities in style editing and background modification, consistent with the findings previously presented in Figure~\ref{fig:edit_3}.
    }
    \label{fig:edit_5}
\end{figure}

\begin{figure}[h]
    \centering
    \includegraphics[width=0.9\textwidth]{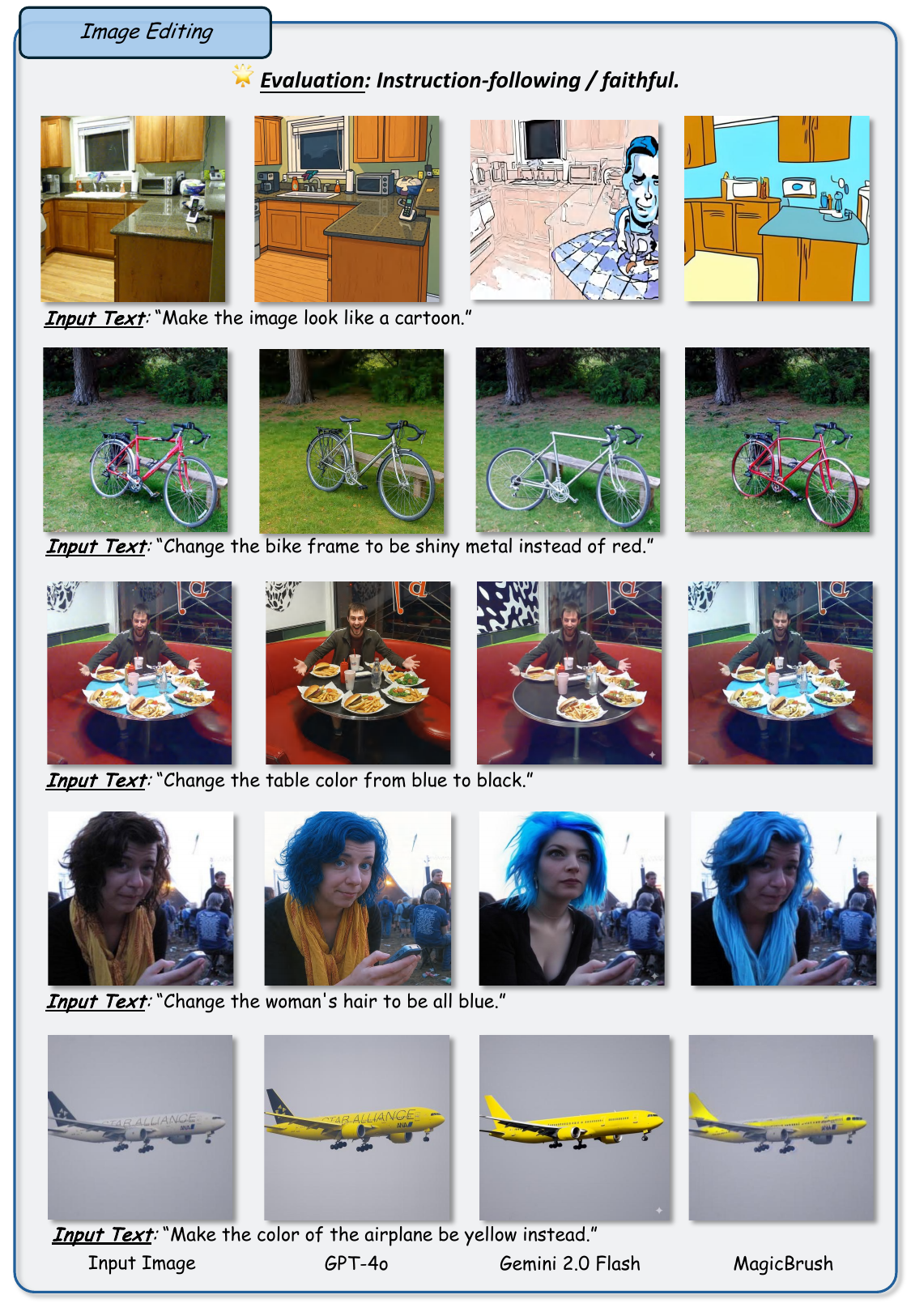}
    \caption{
\textit{\textbf{Task:}} Image editing for modifying visual elements and composition.
\textit{\textbf{Setup:}} GPT-4o vs. Gemini 2.0 Flash~\cite{gemini-2-0-flash}/MagicBrush~\cite{Zhang2023MagicBrush}.
\textit{\textbf{Observations:}} Example 4 highlights GPT-4o's superior image understanding—accurately distinguishing between hair and a scarf (where MagicBrush fails) to execute the edit. In Example 5, its precise retention of the plane's logo and text further demonstrates robust object-preservation capabilities.
    }
    \label{fig:edit_6}
\end{figure}

We compare GPT-4o with MGIE~\cite{fu2024mgie}, LEDITS++~\cite{brack2023ledits}, MagicBrush~\cite{Zhang2023MagicBrush}, and Gemini 2.0 Flash~\cite{gemini-2-0-flash}, which are representative of current SOTA methods. 
These experiments evaluate GPT-4o’s subject preservation and instruction-following capabilities to determine its effectiveness compared with existing methods.
Comparative results are shown in Figure~\ref{fig:edit_1} through Figure~\ref{fig:edit_6}.
We find that GPT-4o achieves performance comparable to, and in many cases surpassing, SOTA baselines in image editing tasks. 
From these examples, GPT-4o exhibits the fewest failure cases, demonstrating a strong generalization ability across a wide variety of editing tasks. It consistently outperforms baseline models across multiple editing scenarios. We highlight several key observations:

\begin{itemize}
\item \textbf{Strengths of GPT-4o in image editing:}
\begin{itemize}
\item \textbf{Fine-grained editing:} GPT-4o shows a superior ability to handle fine-grained editing tasks. For instance, in example 2 of Figure~\ref{fig:edit_1} and example 1 of Figure~\ref{fig:edit_2}, GPT-4o successfully modified small, detailed objects such as a toothpick and pink ballerina slippers, outperforming prior methods.
\item \textbf{Substantial image transformations:} GPT-4o excels at large-scale edits, such as background changes or object transformations, while maintaining visual coherence and realism. These complex edits require robust contextual and semantic understanding. Example 1 in Figure~\ref{fig:edit_3} illustrates GPT-4o’s effective handling of a major background alteration task.
\item \textbf{Subject preservation:} GPT-4o demonstrates strong subject-preserving capabilities, avoiding common artifacts such as facial distortions or component loss. In example 2 of Figure~\ref{fig:edit_1}, GPT-4o retains the content of a drink that Gemini 2.0 Flash erroneously altered. Similarly, in example 5 of Figure~\ref{fig:edit_6}, GPT-4o best preserves fuselage patterns and textual markings on an airplane.
\item \textbf{Instruction and original image adherence:} GPT-4o shows a notable ability to follow instructions and maintain the structure of the original image, particularly in style editing and tasks involving object quantity, size, or position. This likely stems from its advanced understanding of both the image content and the editing instructions. For example, Figure~\ref{fig:edit_5} demonstrates GPT-4o’s capability in style translation. Example 2 in Figure~\ref{fig:edit_4} shows its understanding of the term “orange” in both textual and visual contexts. A similar ability is illustrated in example 4 of Figure~\ref{fig:edit_6}.
\end{itemize}
\item \textbf{Limitations of GPT-4o in image editing:}
\begin{itemize}
\item GPT-4o underperforms in scenarios where strict preservation of the original image’s lighting, shading, and color tones is required. In such cases, the edited images may exhibit noticeable shifts in visual consistency. This is evident in examples 1 and 5 of Figure~\ref{fig:edit_1} and example 4 of Figure~\ref{fig:edit_2}.
\item In some cases, GPT-4o may fail to retain image details outside the intended edit region. For instance, example 4 in Figure~\ref{fig:edit_1} shows a degradation in image quality in non-targeted areas.
\end{itemize}
\end{itemize}

In summary, GPT-4o demonstrates substantial advancements in image editing, showing exceptional capabilities in detailed and large-scale edits, subject preservation, and adherence to instructions. While there are limitations in strictly maintaining original image characteristics such as lighting and tonal consistency, GPT-4o significantly reduces failure cases and outperforms existing baselines across a wide range of editing tasks, pushing the boundaries of current SOTA performance.

\clearpage

\subsubsection{Customization}
Customization, also known as subject-driven generation or personalization, aims to enable visual generative models to generate visual concepts from given reference images.
Initial methods \cite{gal2022image, ruiz2023dreambooth} have achieved this by optimizing text embeddings or model weights.
Subsequent approaches \cite{kumari2023multi, gu2024mix, jiang2024mc, zhang2024ssr, shi2024relationbooth, zhou2024magictailor} expanded on these approaches to handle multiple visual concepts.
Customization plays a crucial role in making visual generative models more flexible and applicable across diverse domains. By empowering models to adapt to user-provided inputs, it ensures outputs are tailored to specific visual concepts. 
This is particularly significant in industries such as artistic creation and advertising, where individualization and creativity are paramount. 

\begin{figure}[h]
    \centering
    \includegraphics[width=1\textwidth]{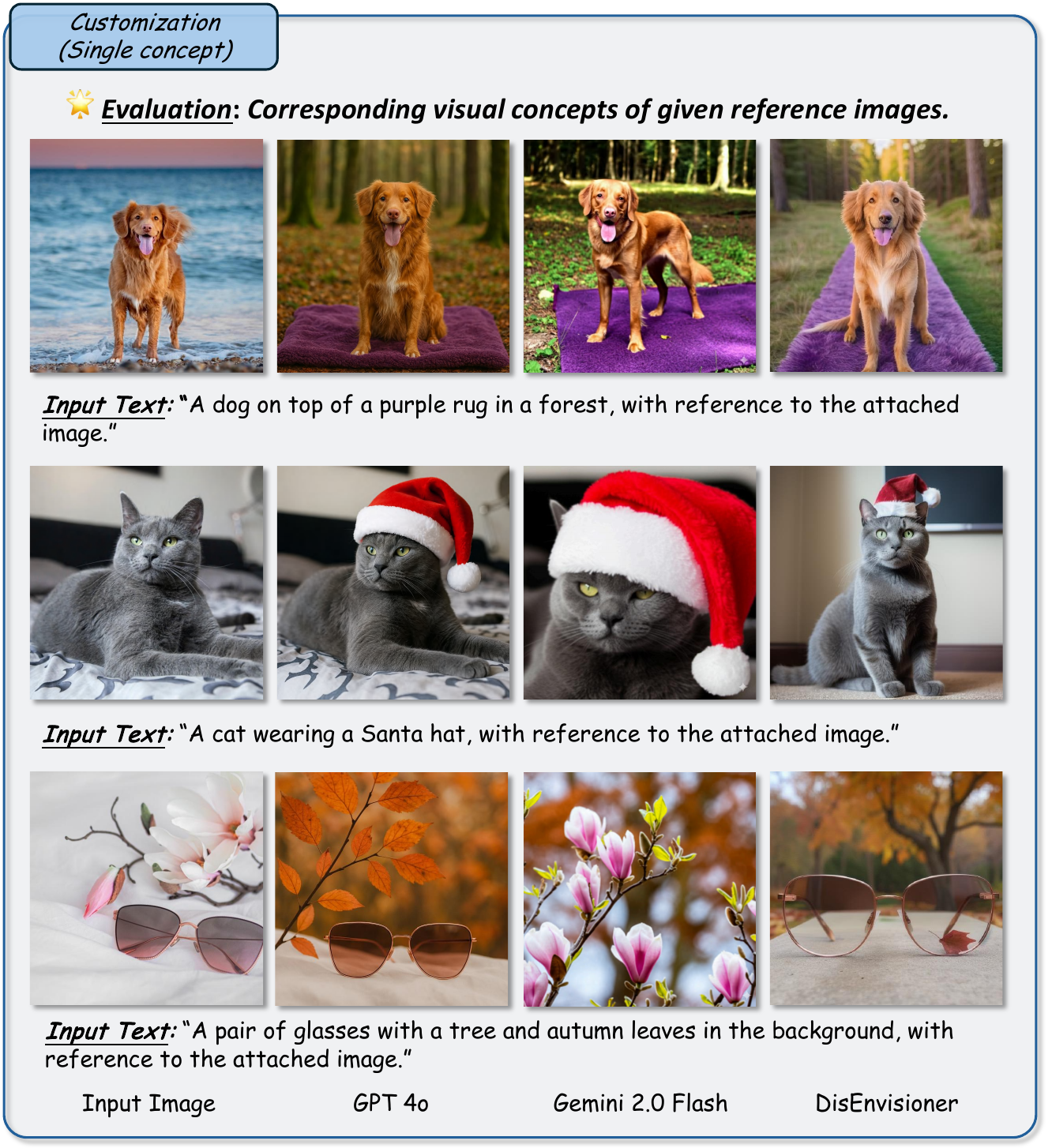}
    \caption{\textit{\textbf{Task:}} Single-concept customization. The goal is to generate images that faithfully reproduce a single visual concept from reference images while aligning with a given textual description.  
    \textit{\textbf{Setup:}} Reference images are collected from prior works \cite{disenvisioner}, and results are compared across GPT-4o, Gemini 2.0 Flash~\cite{gemini-2-0-flash}, and DisEnvisioner~\cite{disenvisioner}. Each row includes the input reference image, text prompt, and the corresponding outputs.  
    \textit{\textbf{Observations:}} GPT-4o demonstrates strong performance in faithfully reproducing the single visual concept with high fidelity while adhering closely to the given textual description. It consistently outperforms Gemini 2.0 Flash and achieves results comparable to the SOTA method DisEnvisioner. However, some generated images still exhibit minor ``copy-paste'' artifacts, indicating room for further improvement.} 
    \label{fig:custom_1}
\end{figure}

\begin{figure}[h]
    \centering
    \includegraphics[width=1\textwidth]{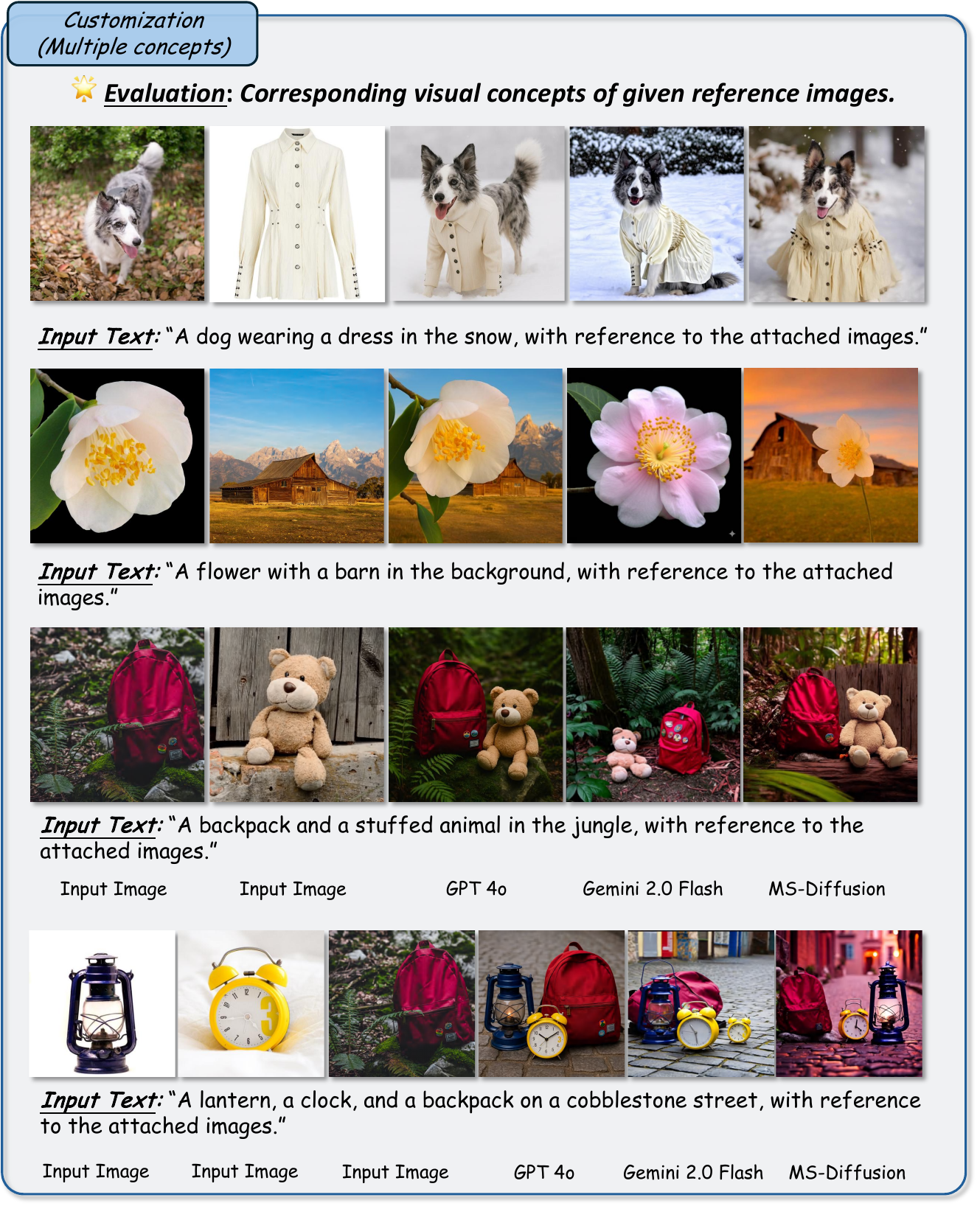}
    \caption{\textit{\textbf{Task:}} Multi-concept customization. The goal is to generate images that effectively combine multiple visual concepts from reference images while aligning with a given textual description.  
    \textit{\textbf{Setup:}} Reference images are collected from prior works \cite{msdiffusion}, and results are compared across GPT-4o, Gemini 2.0 Flash~\cite{gemini-2-0-flash}, and MS-Diffusion~\cite{msdiffusion}. Each row includes the input reference images, text prompt, and the corresponding outputs.  
    \textit{\textbf{Observations:}} GPT-4o achieves competitive results in combining multiple visual concepts, showing strong fidelity to individual concepts and alignment with text prompts. However, its performance declines with unique or complex combinations. Despite this, GPT-4o outperforms Gemini 2.0 Flash and achieves results on par with SOTA methods.} 
    \label{fig:custom_2}
\end{figure}

To evaluate the performance of GPT-4o in this challenging task, we collect reference images from previous relevant works \cite{disenvisioner, msdiffusion}, and conduct qualitative comparisons as shown in Figure~\ref{fig:custom_1} and Figure~\ref{fig:custom_2}.
For single-concept customization, we compare GPT-4o with Gemini 2.0 Flash and DisEnvisioner \cite{disenvisioner}. 
The results demonstrate that GPT-4o not only faithfully reproduces the visual concept from the reference image but also accurately adheres to the given textual description. 
In this task, GPT-4o significantly outperforms Gemini 2.0 Flash and achieves performance on par with the SOTA customization method.
However, the images generated by GPT-4o still exhibit some ``copy-paste'' artifacts, leaving room for further improvement in the future.
For multi-concept customization, we compare GPT-4o with Gemini 2.0 Flash and MS-Diffusion \cite{msdiffusion}. 
In this task, GPT-4o can still achieve competitive results for customizing multiple visual concepts in different contexts. 
Unfortunately, it struggles with certain unique combinations (e.g., making a dog wear a human dress), which could be attributed to the lack of relevant customization training data.

Overall, GPT-4o demonstrates impressive performance in both single-concept and multi-concept customization tasks, showcasing strong concept fidelity and great text alignment. Despite some limitations, GPT-4o achieves remarkable results on par with SOTA customization methods and outperforms Gemini 2.0 Flash.

\clearpage

\subsubsection{Story Image Generation}

\begin{figure}[h]
    \centering
    \includegraphics[width=1\textwidth]{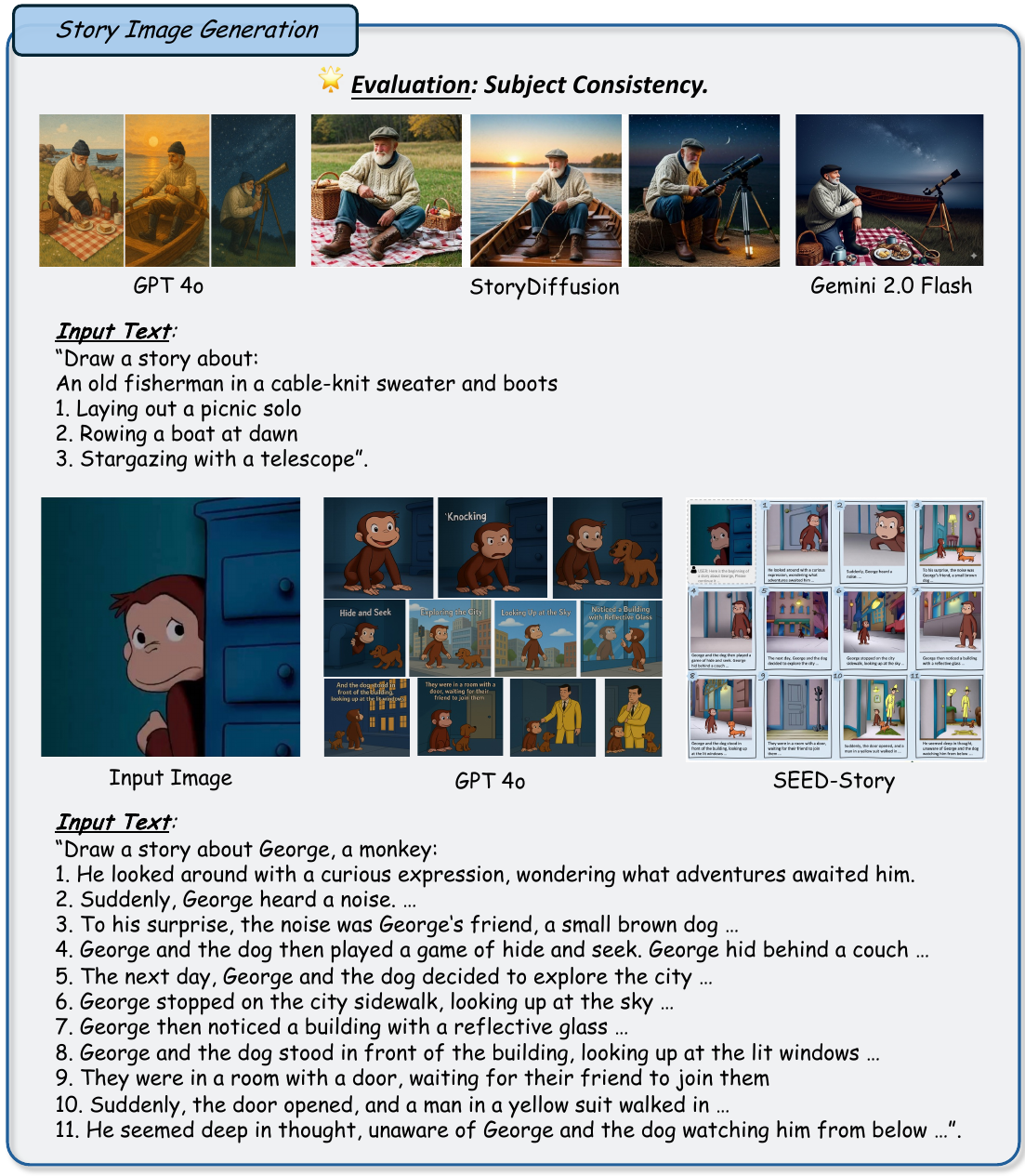}
    \caption{
\textit{\textbf{Task:}} Story image generation. The goal is to generate coherent story sequences based on narrative text, optionally conditioned on initial story frames or character images.
\textit{\textbf{Setup:}} Each example combines an input narrative (and, when available, reference character images) with a series of generated story panels. We compare outputs from GPT-4o against Gemini 2.0 Flash~\cite{gemini-2-0-flash}, StoryDiffusion~\cite{storydiffusion}, and SEED-Story~\cite{SEED-Story}.
\textit{\textbf{Observations:}} GPT-4o exhibits strong narrative coherence and panel continuity, matching or surpassing general baselines.}
    \label{fig:story_generation_1}
\end{figure}

\begin{figure}[h]
    \centering
    \includegraphics[width=1\textwidth]{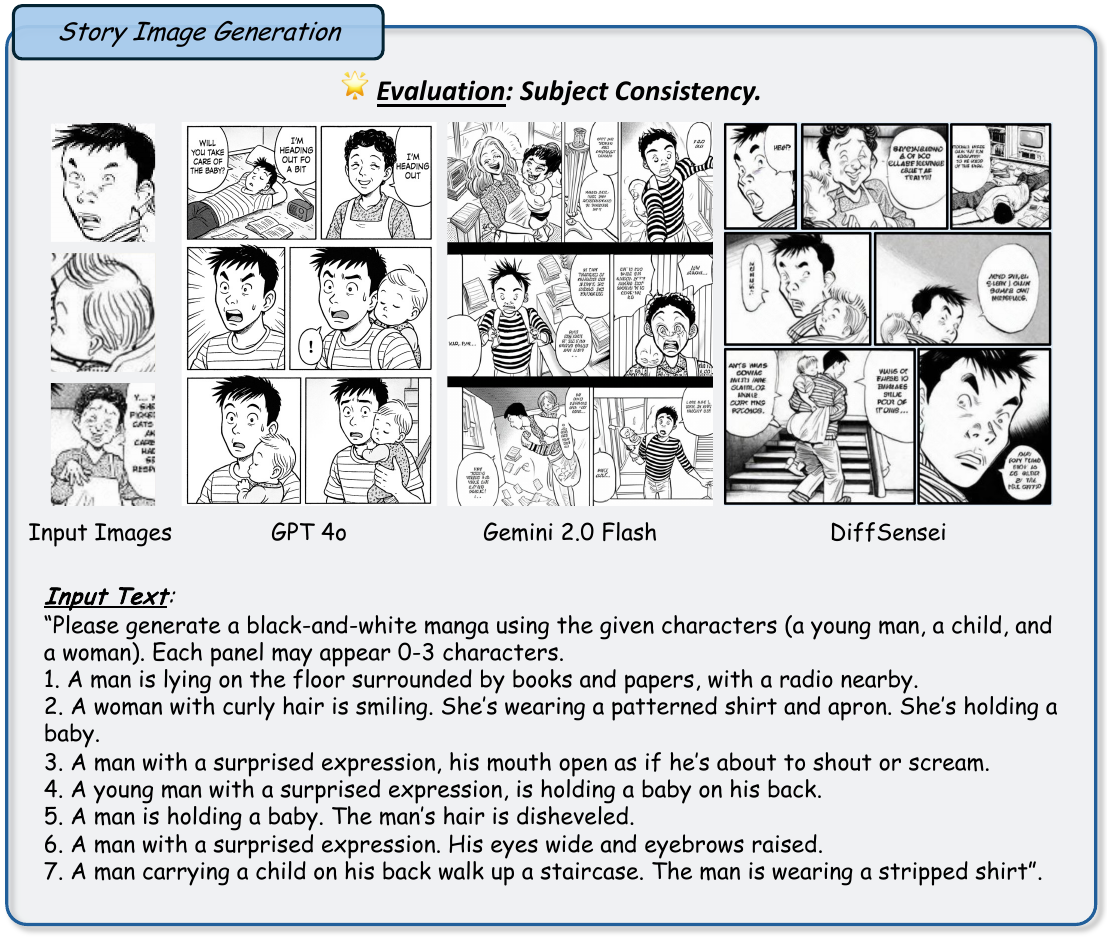}
    \caption{
    \textit{\textbf{Task:}} Story image generation. The goal is to generate coherent story sequences based on narrative text, optionally conditioned on initial story frames or character images. \textit{\textbf{Setup:}} Each example combines an input narrative (and, when available, reference character images) with a series of generated story panels. We compare outputs from GPT-4o against baselines including Gemini 2.0 Flash~\cite{gemini-2-0-flash} and DiffSensei~\cite{DiffSensei}. \textit{\textbf{Observations:}} GPT-4o shows minor shortcomings in precise character consistency and panel count in specialized contexts, such as Japanese black-and-white manga, where dedicated models like DiffSensei deliver superior performance.
    }
    \label{fig:story_generation_2}
\end{figure}

Story image generation is a task to generate coherent stories based on input text narratives. The conditions may also include the first story frame or character images. We choose Gemini 2.0 Flash~\cite{gemini-2-0-flash}, StoryDiffusion~\cite{storydiffusion}, SEED-Story~\cite{SEED-Story}, and DiffSensei~\cite{DiffSensei} as baselines, due to their proven ability to generate coherent and expressive story images and their public availability. The results are shown in Figure~\ref{fig:story_generation_1} and Figure~\ref{fig:story_generation_2}.

In the first example, GPT-4o and StoryDiffusion successfully generate a three-panel short story about a fisherman, whereas Gemini 2.0 Flash fails by producing a single panel that appears to combine the three story narratives.
In the second example, the story narrative is longer, spanning 11 panels. To evaluate this scenario with GPT-4o, we instruct the model to generate story images sequentially—using the input image and all previously generated images along with the corresponding text prompts. As shown in the figure, GPT-4o is capable of generating a long story with consistency.
In the final example, we examine a Japanese black-and-white manga style with multiple input character images. GPT-4o is able to generate coherent stories, though it exhibits minor errors in character consistency (notably with the depiction of the woman) and misalignment with the input narrative (the narrative requires 7 panels, but only 6 are generated). The baseline Gemini 2.0 Flash performs worse, failing to preserve character status and the correct number of panels, as it also produces only 6 panels. Conversely, the DiffSensei model demonstrates superior performance, likely due to its specialized design and training for Japanese black-and-white manga generation.

In conclusion, while GPT-4o achieves comparable performance to current baselines in story image generation, it shows limitations in specific scenarios—such as Japanese black-and-white manga and precise character status preservation—when compared to methods specifically tailored for those tasks.

\clearpage

\subsubsection{Low-level Vision}

Low-level vision tasks aim to enhance the basic quality or detail of visual content by improving various aspects of an image. Initial methods often focused on optimizing single tasks, such as super-resolution \cite{ntire_sr,sun2024coser}, denoising \cite{scpgabnet, scpgan, bnnlan}, restoration \cite{dinoir, instructir, llrrnet, ram, udrs2former, chen2022snowformer, chen2024teaching}, color adjustment \cite{ctrlcolor}, and more \cite{latentinpainting, StrDiffusion, sgt, dream360, iclight}. As the technology progressed, subsequent approaches expanded these techniques to handle multiple low-level tasks simultaneously, which is called universal image restoration. Low-level tasks play a critical role in image generation and editing, allowing visual generative models to provide higher-quality outputs in real-world applications. By enabling models to adapt to diverse inputs, they ensure that the generated images perform well across different visual tasks. This is especially important in areas such as image restoration and video enhancement, where high-precision visual content optimization is crucial, such as in film post-production and autonomous driving.

\begin{figure}[h]
    \centering
    \includegraphics[width=0.92\textwidth]{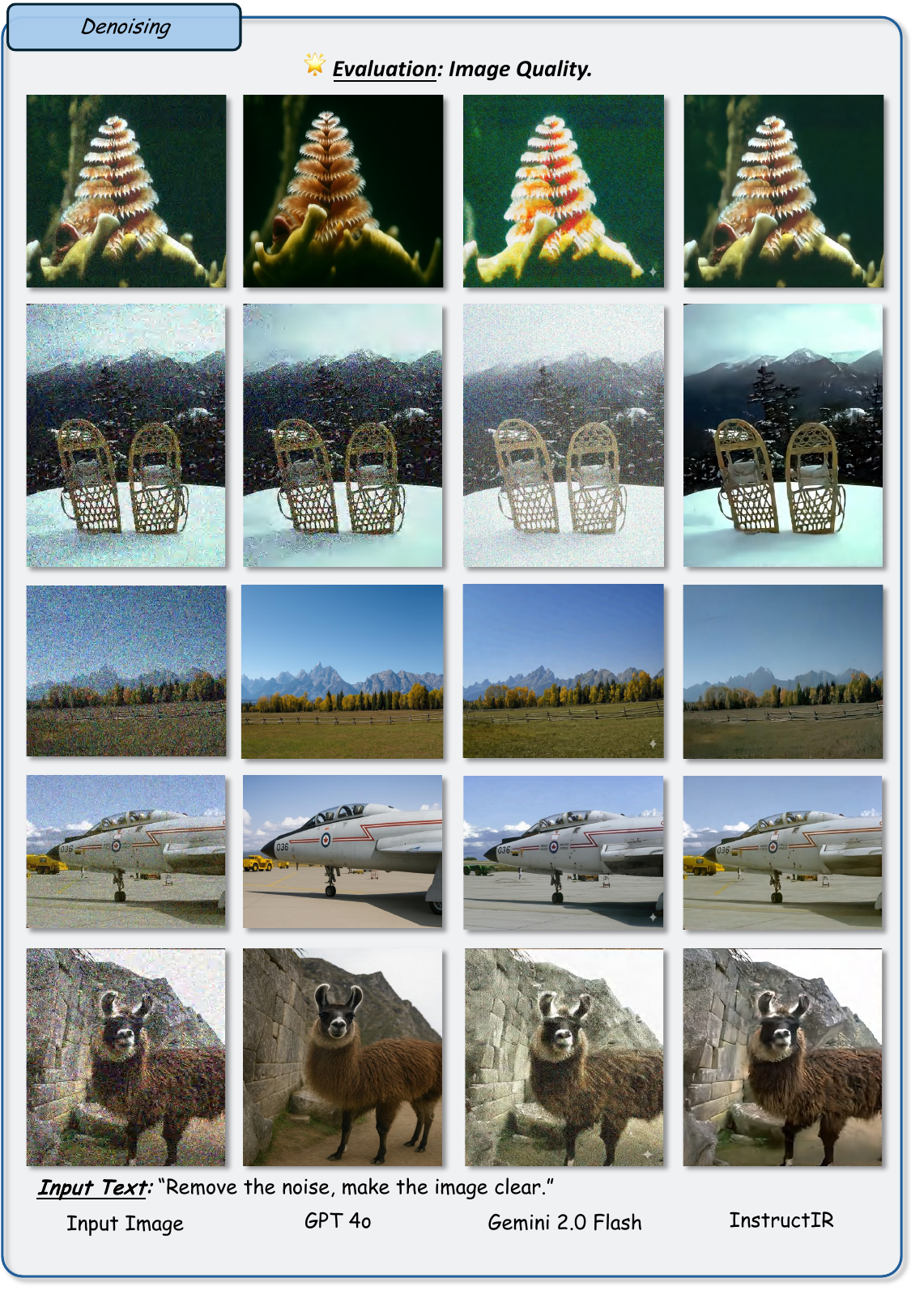}
    \caption{\textit{\textbf{Task:}} image denoising, aiming to remove the noise information and obtain high-quality clear version.
\textit{\textbf{Setup:}} We compare GPT-4o with InstructIR~\cite{instructir} and Gemini 2.0 Flash~\cite{gemini-2-0-flash} to evaluate the denoised images.
\textit{\textbf{Observations:}} GPT-4o can restore high-quality denoised images. Except for the second image, where the noise cannot be completely removed, the other images are free from noise. However, for low-level tasks, GPT-4o does not maintain content consistency well — the background colors and object shapes in many images have changed, such as the background color in the first image and the floor in the fourth image. }

    \label{fig:low_level_1}
\end{figure}

\begin{figure}[h]
    \centering
    \includegraphics[width=1\textwidth]{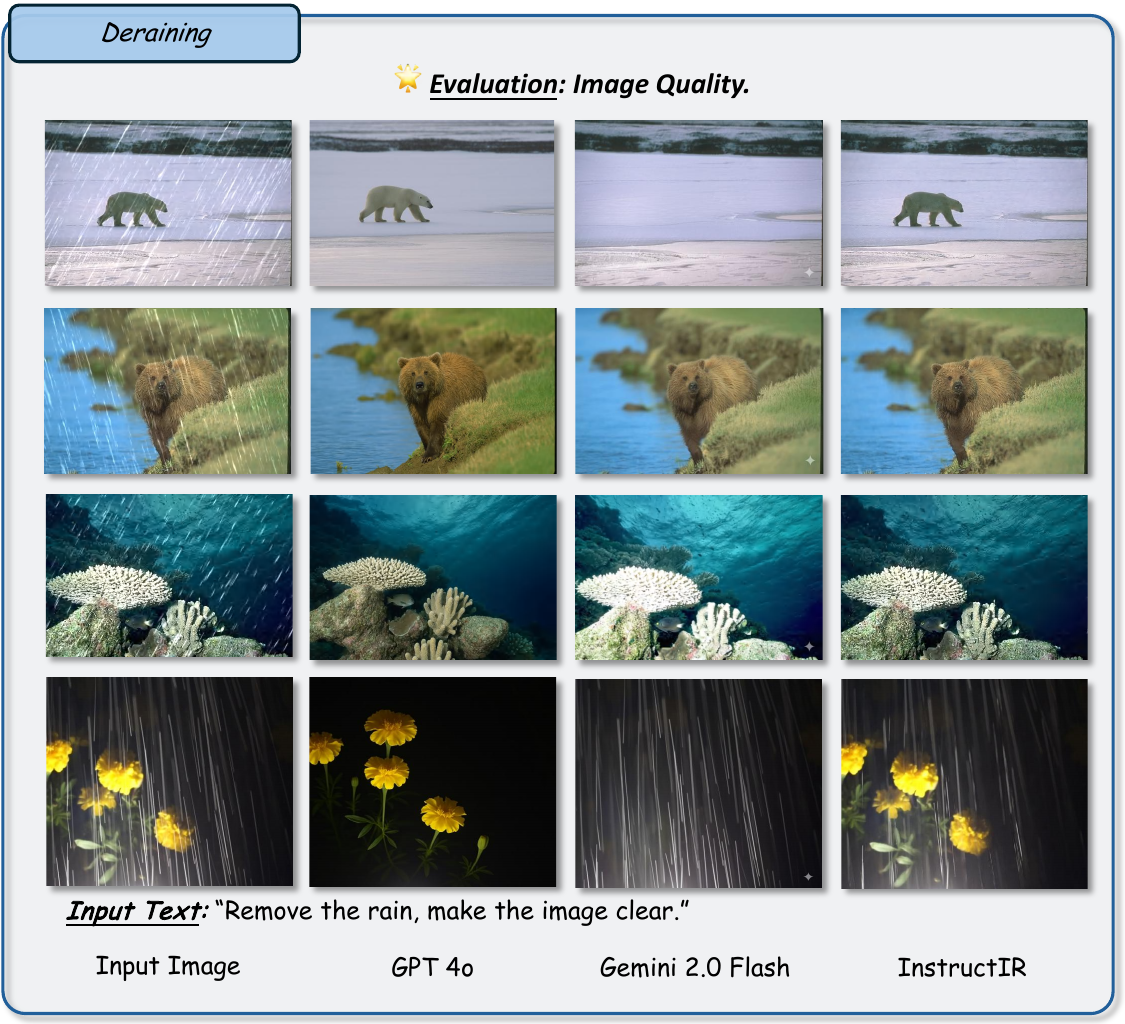}
    \caption{\textit{\textbf{Task:}} image deraining, aiming to remove the rain streak and get high-quality clear version.
\textit{\textbf{Setup:}} We compare GPT-4o with established baselines such as InstructIR~\cite{instructir} and Gemini 2.0 Flash~\cite{gemini-2-0-flash} to evaluate the derained images.
\textit{\textbf{Observations:}} The overall performance of the GPT-4o is well. However, the model struggles with maintaining content consistency in low-level visual details — for instance, the polar bear's background in the first image becomes unnaturally pink, and the underwater scene loses depth and clarity. The flowers also appear altered in color and arrangement. In contrast, InstructIR demonstrates the most consistent performance across all examples, effectively removing rain while preserving the original scene's structure, color, and composition. Overall, InstructIR is the most balanced and accurate model for image restoration in this comparison.}

    \label{fig:low_level_2}
\end{figure}

\begin{figure}[h]
    \centering
    \includegraphics[width=1\textwidth]{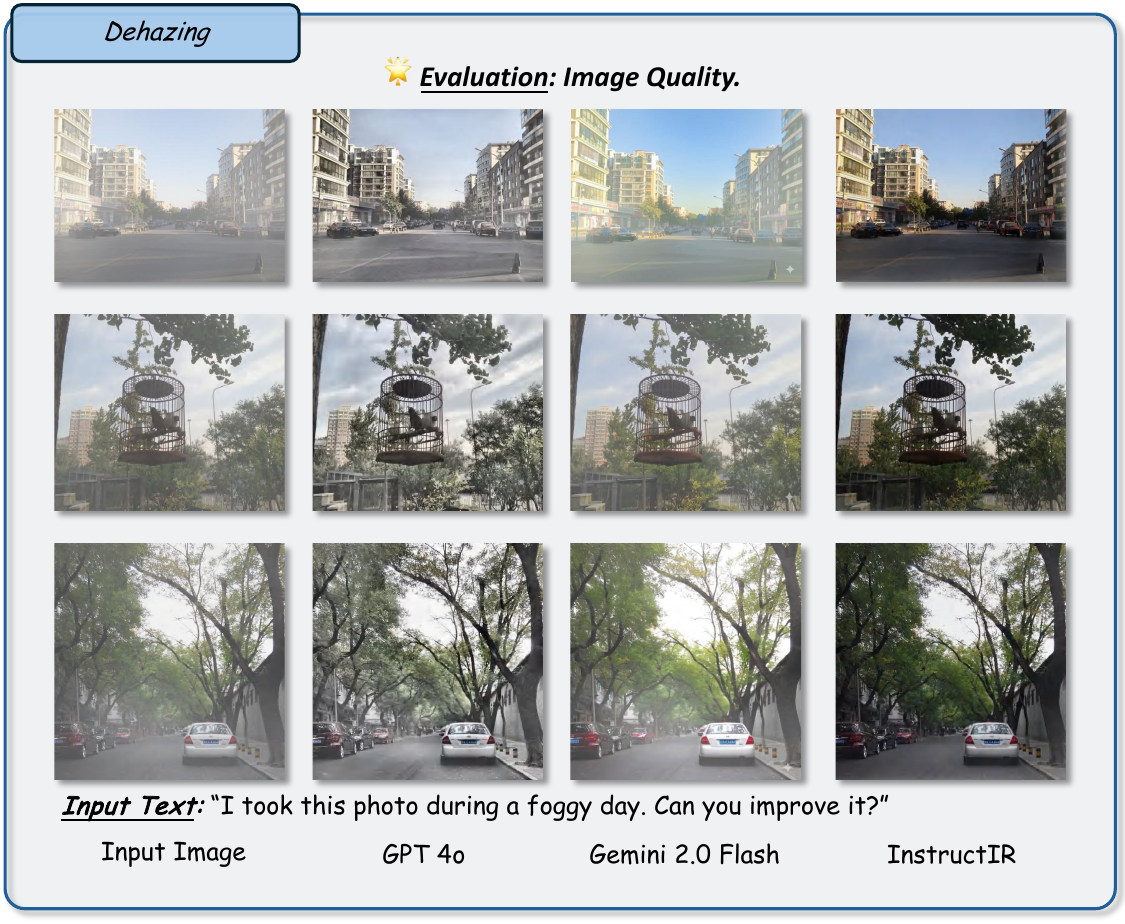}
    \caption{\textit{\textbf{Task:}} image dehazing, aiming to remove the haze information and get high-quality clear version.
\textit{\textbf{Setup:}} We compare GPT-4o with established baselines such as InstructIR~\cite{instructir} and Gemini 2.0 Flash~\cite{gemini-2-0-flash} to evaluate the dehazed images.
\textit{\textbf{Observations:}} GPT-4o performs moderately well in dehazing, managing to restore clearer structures and contrast in most scenes. However, its outputs often have a grayish or desaturated tone, especially visible in the second and third rows. Gemini 2.0 Flash produces more colorful results but tends to leave some haze behind, leading to a less crisp output. InstructIR outperforms both, offering the most visually natural and sharp dehazing across all examples while preserving original colors and details. Overall, InstructIR demonstrates the strongest capability in removing haze while maintaining realism.}

    \label{fig:low_level_3}
\end{figure}

\begin{figure}[h]
    \centering
    \includegraphics[width=1\textwidth]{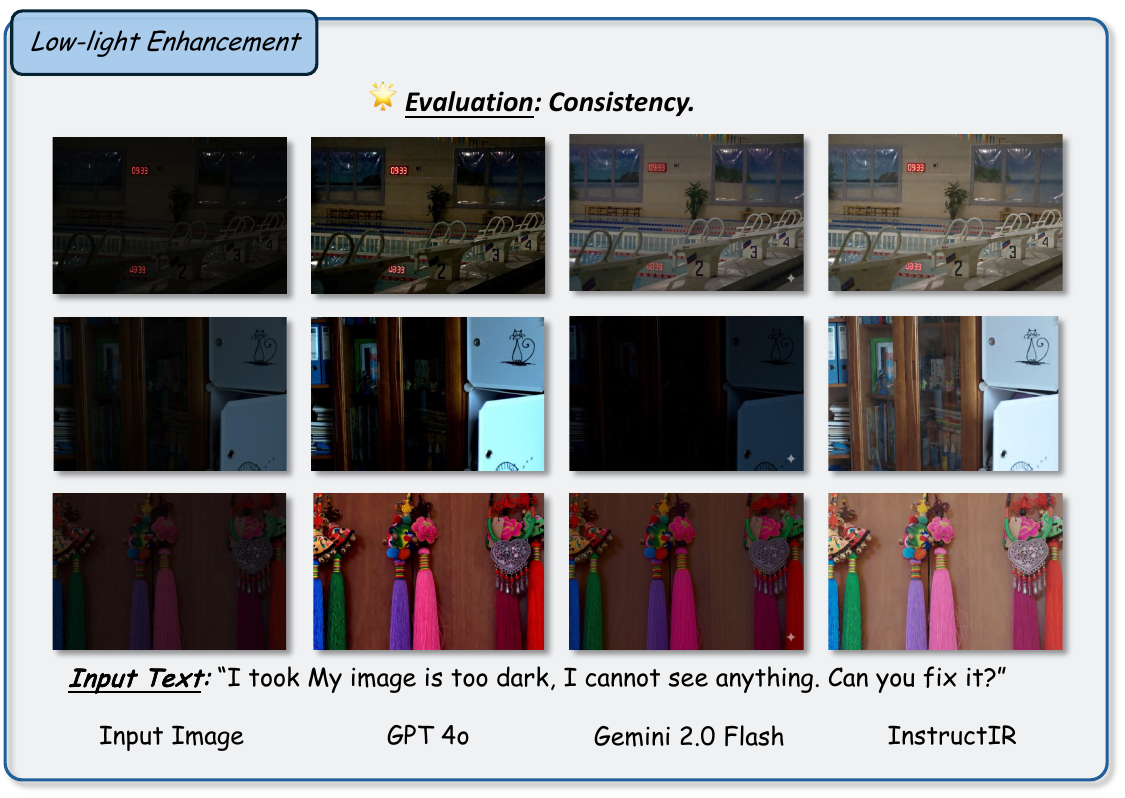}
    \caption{\textit{\textbf{Task:}} low-light image enhancement, aiming to increase the brightness of the image to obtain a high brightness image.
\textit{\textbf{Setup:}} We compare GPT-4o with established baselines such as InstructIR~\cite{instructir} and Gemini 2.0 Flash~\cite{gemini-2-0-flash} to evaluate the brightness images.
\textit{\textbf{Observations:}} In low-light enhancement tasks, GPT-4o can brighten images and recover basic visibility, but often introduces unnatural lighting and loses detail, especially in the second row, where the image remains overly dark. InstructIR consistently delivers the most balanced results, enhancing visibility while preserving true colors and textures, making it the best performer across all three examples.}
    \label{fig:low_level_4}
\end{figure}

\begin{figure}[h]
    \centering
    \includegraphics[width=1\textwidth]{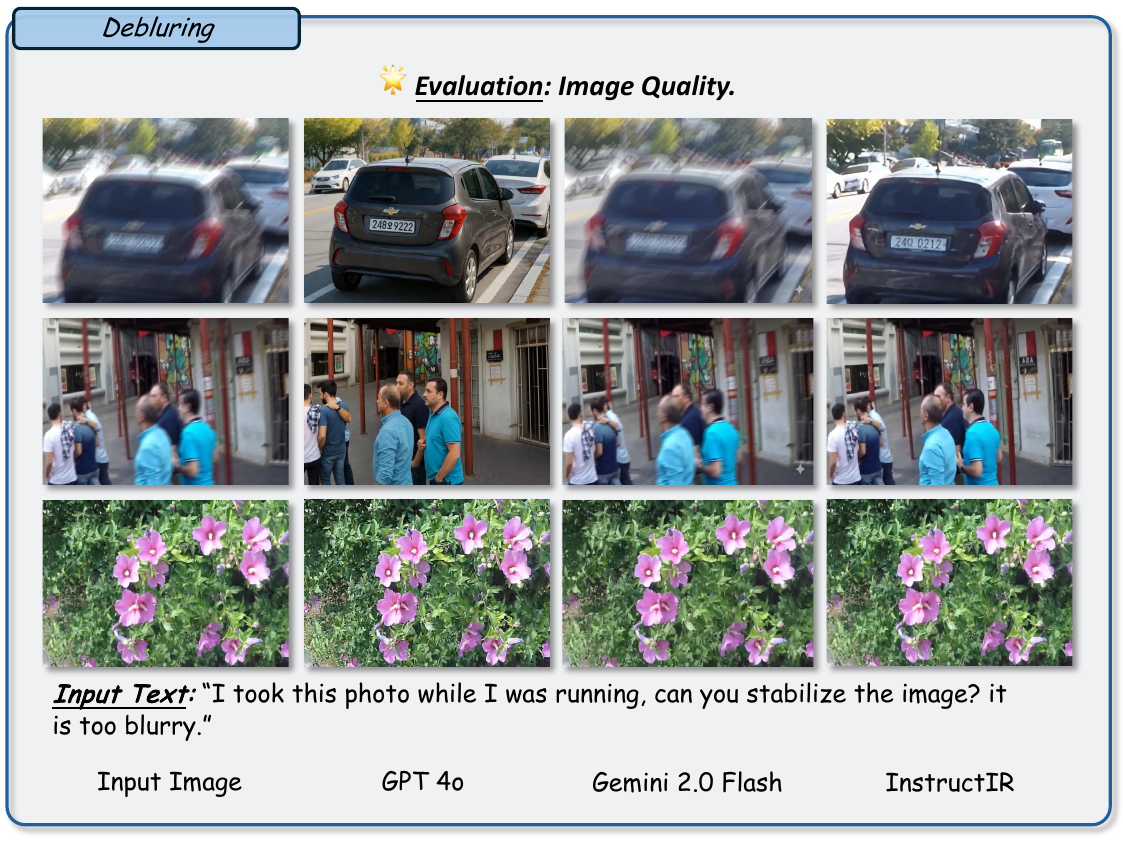}
    \caption{\textit{\textbf{Task:}} image deblurring, aiming to remove the blur information to obtain a clear image.
\textit{\textbf{Setup:}} We compare GPT-4o with established baselines such as InstructIR~\cite{instructir} and Gemini 2.0 Flash~\cite{gemini-2-0-flash} to evaluate the deblurred images.
\textit{\textbf{Observations:}} For motion deblurring, GPT-4o recovers some sharpness, especially in fine details like text or faces, but the content is not matched with the original image. Gemini 2.0 Flash sharpens the image slightly better in some cases but can introduce over-smoothing, making the result look artificial. InstructIR demonstrates the best deblurring performance overall — restoring clear edges, facial features, and text while maintaining natural textures. It consistently produces the most stable and visually convincing results across all examples.}
    \label{fig:low_level_5}
\end{figure}

\begin{figure}[h]
    \centering
    \includegraphics[width=1\textwidth]{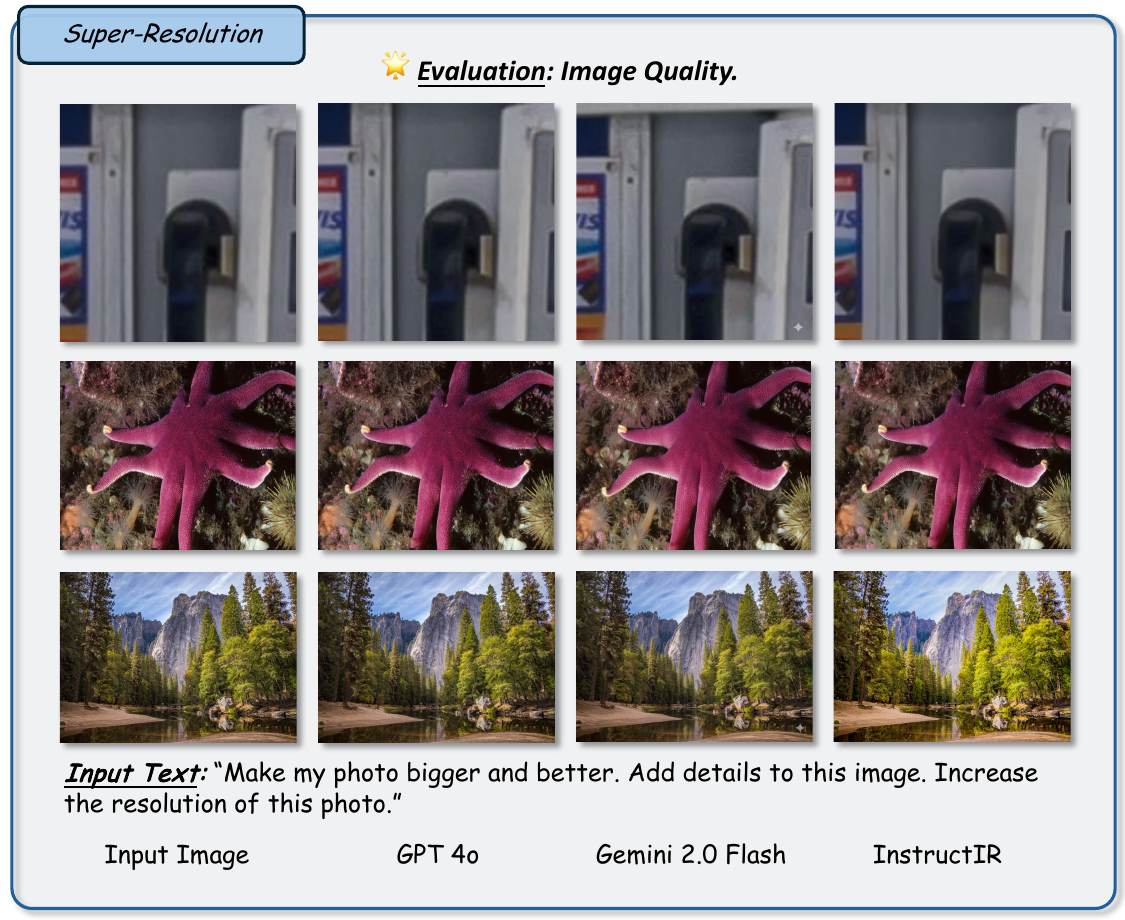}
    \caption{\textit{\textbf{Task:}} image super-resolution, aiming to improve the image resolution.
    \textit{\textbf{Setup:}} We compare GPT-4o with established baselines such as InstructIR~\cite{instructir} and Gemini 2.0 Flash~\cite{gemini-2-0-flash} to evaluate the deblurred images.
\textit{\textbf{Observations:}} In super-resolution, InstructIR delivers the most natural and detailed results across all examples—restoring fine edges in the card reader, realistic texture on the octopus, and sharp trees in the landscape. GPT-4o enhances clarity but misses details like the octopus surface and tree leaves. Gemini 2.0 Flash produces sharper outputs than GPT-4o but introduces unnatural textures and artifacts, especially in organic regions like the octopus and foliage.}
    \label{fig:low_level_6}
\end{figure}

\begin{figure}[h]
    \centering
    \includegraphics[width=1\textwidth]{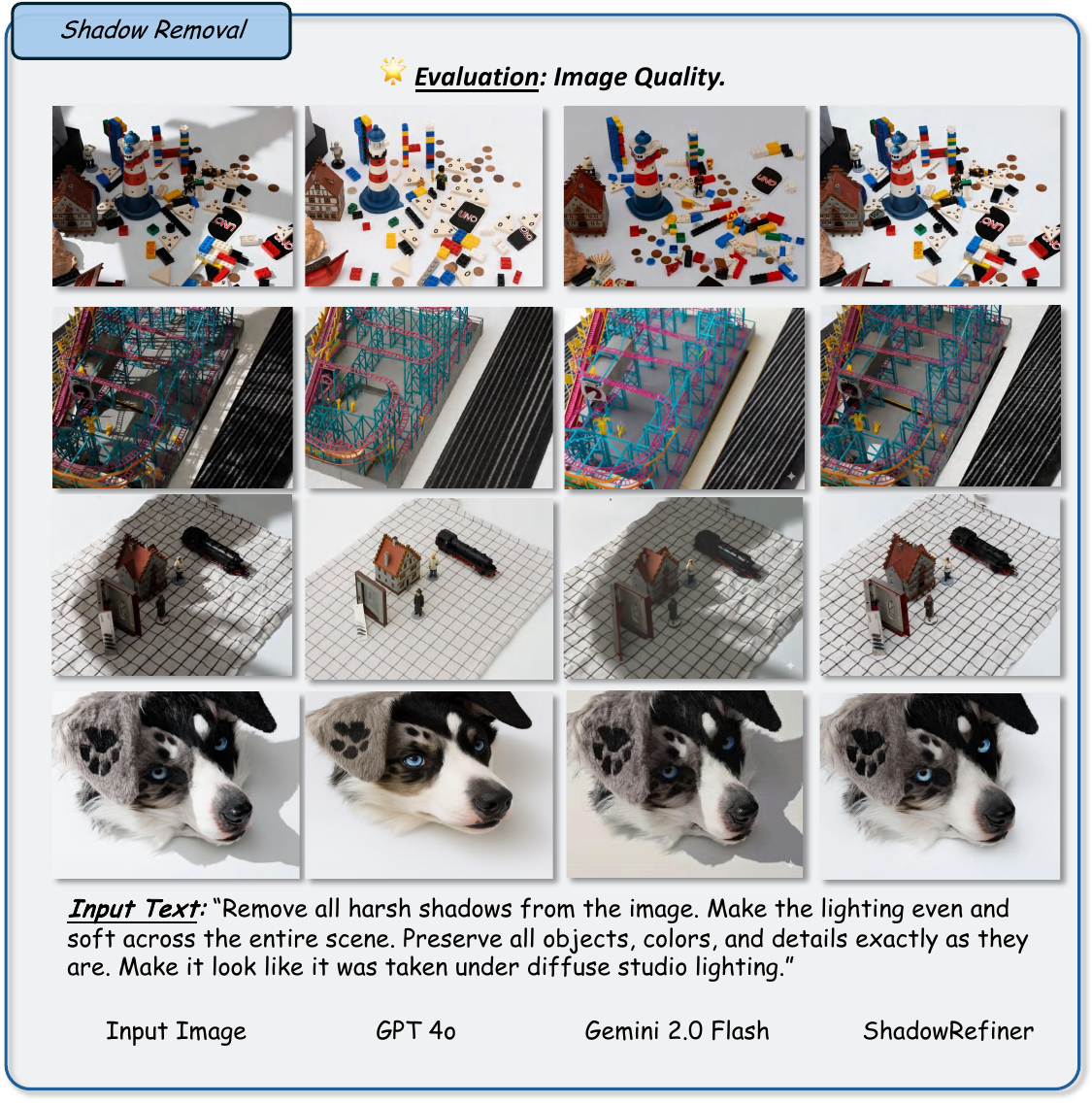}
    \caption{\textit{\textbf{Task:}} Shadow removal, aiming to eliminate harsh shadows while preserving the integrity of the scene, textures, and lighting balance.
\textit{\textbf{Setup:}} We compare GPT-4o with Gemini 2.0 Flash~\cite{gemini-2-0-flash} and ShadowRefiner \cite{shadowrefiner} to evaluate how well each method removes shadows and retains original object fidelity and lighting consistency.
\textit{\textbf{Observations:}} ShadowRefiner consistently achieves the most natural and effective shadow removal. It produces even, diffuse lighting across all scenes—e.g., softening shadows without distorting textures in complex scenes like the miniatures and dog portrait. Gemini 2.0 Flash removes shadows reasonably but occasionally leaves faint traces or flattens contrast, as seen in the second and fourth rows. GPT-4o shows stronger shadow reduction than Gemini 2.0 Flash but sometimes alters surface brightness or loses detail fidelity. ShadowRefiner best preserves the original color tones and textures while eliminating harsh shadows.}
    \label{fig:low_level_10}
\end{figure}

\begin{figure}[h]
    \centering
    \includegraphics[width=1\textwidth]{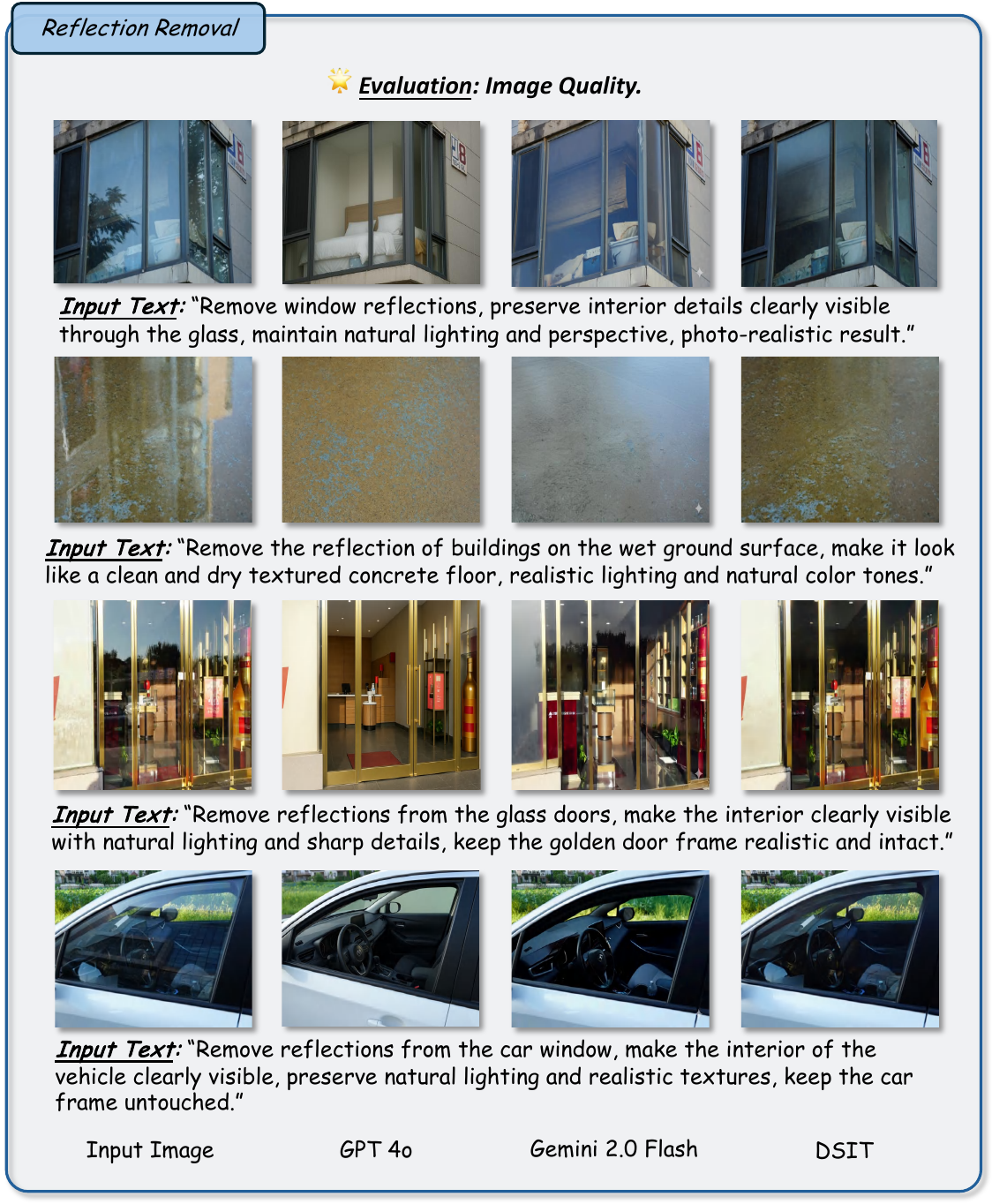}
    \caption{\textit{\textbf{Task:}} Reflection removal, aiming to eliminate unwanted reflections from transparent or reflective surfaces while preserving original content and realistic lighting.
\textit{\textbf{Setup:}} We compare GPT-4o with Gemini 2.0 Flash~\cite{gemini-2-0-flash} and DSIT \cite{dsit}, assessing their ability to remove reflections while maintaining scene realism, texture fidelity, and lighting consistency.
\textit{\textbf{Observations:}} DSIT shows the most effective and natural reflection removal across all examples. It restores interior visibility through windows (e.g., bed and car interior) while preserving lighting and geometry. Gemini 2.0 Flash removes some reflections but often leaves faded traces or dulls textures, especially on glass doors and wet pavement. GPT-4o performs better than Gemini 2.0 Flash in preserving background details but sometimes alters color tones and sharpness. Overall, DSIT provides the cleanest and most photorealistic results, especially for transparent surfaces like glass and reflective wet ground.}
    \label{fig:low_level_11}
\end{figure}

We evaluate the performance of GPT-4o in this challenging task. Firstly, for some image restoration tasks, such as super resolution, denoising, deraining, low-light enhancement, deblurring and dehazing. We collect reference images from previous relevant works Gemini 2.0 Flash and a universal image restoration model, InstrucIR \cite{instructir}, as shown in Figures~\ref{fig:low_level_1},~\ref{fig:low_level_2}, \ref{fig:low_level_3}, \ref{fig:low_level_4}, \ref{fig:low_level_5}, \ref{fig:low_level_6}, \ref{fig:low_level_10}, \ref{fig:low_level_11}. In most scenarios, GPT-4o guarantees high-quality output images, outperforming Gemini 2.0 Flash. However, there are still some degradation issues that are difficult to remove, as seen in the second image of the image denoising task. On the other hand, for low-level image restoration tasks, maintaining pixel consistency between the output and input images is crucial. GPT-4o does not perform well in this regard, as the content of many images changes. In contrast, InstructIR, designed specifically for image restoration, performs better, effectively removing degradation while maintaining pixel consistency throughout.

\begin{figure}[h]
    \centering
    \includegraphics[width=1\textwidth]{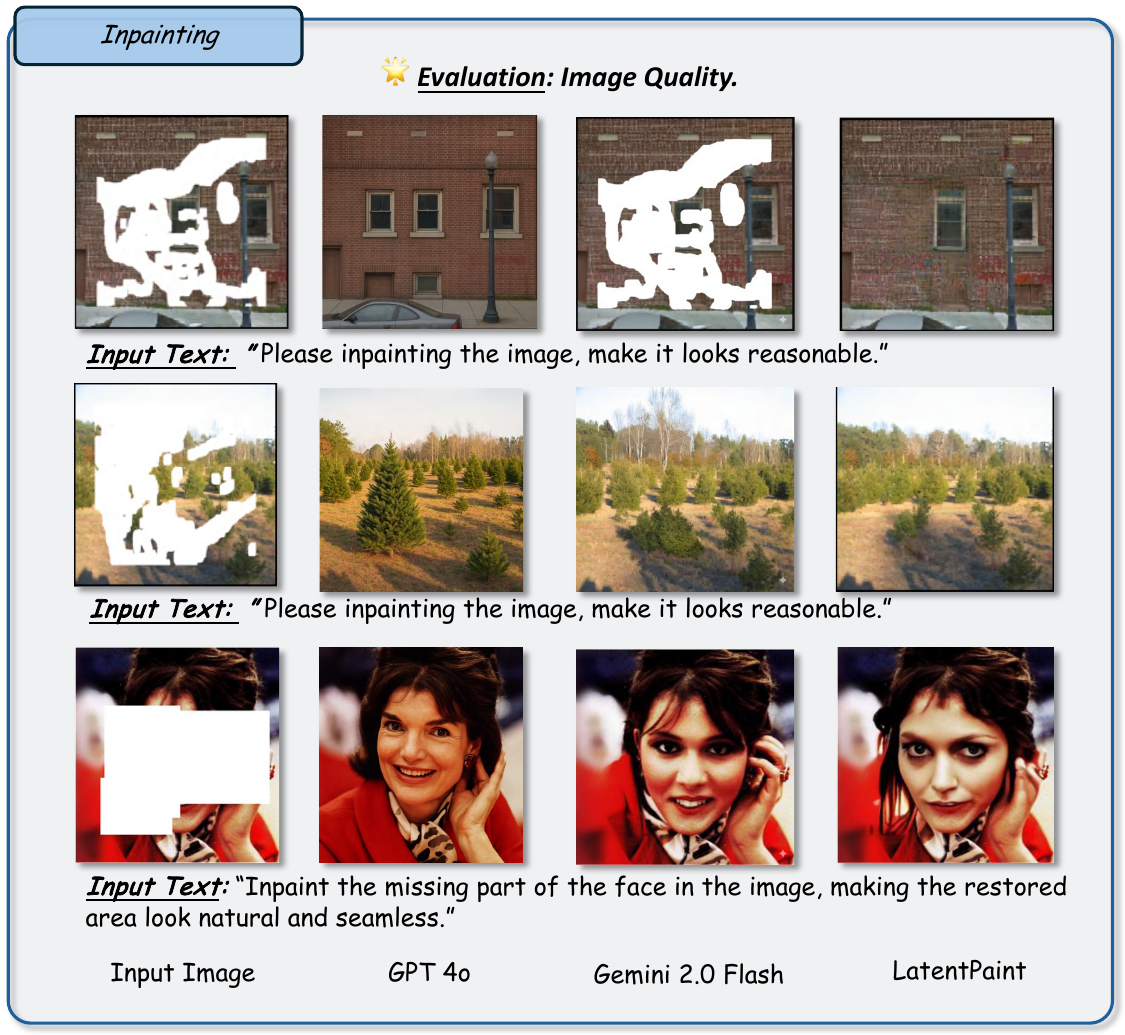}
    \caption{\textit{\textbf{Task:}} Image inpainting, aiming to restore missing or masked regions in an image to appear natural and consistent with the context.
\textit{\textbf{Setup:}} We compare GPT-4o with baselines such as Gemini 2.0 Flash~\cite{gemini-2-0-flash} and LatentPaint \cite{latentinpainting}, evaluating their ability to fill in masked regions realistically.
\textit{\textbf{Observations:}} GPT-4o produces plausible completions but often lacks fine structure and texture alignment—e.g., the bricks in the first row appear flat and misaligned. Gemini 2.0 Flash generates more visually coherent textures, especially in natural scenes like the second row, but can introduce slight over-smoothing. LatentPaint performs the best, accurately reconstructing facial details and complex textures such as hair and expression in the third row, demonstrating superior semantic understanding and visual consistency.}
    \label{fig:low_level_7}
\end{figure}

\begin{figure}[h]
    \centering
    \includegraphics[width=0.85\textwidth]{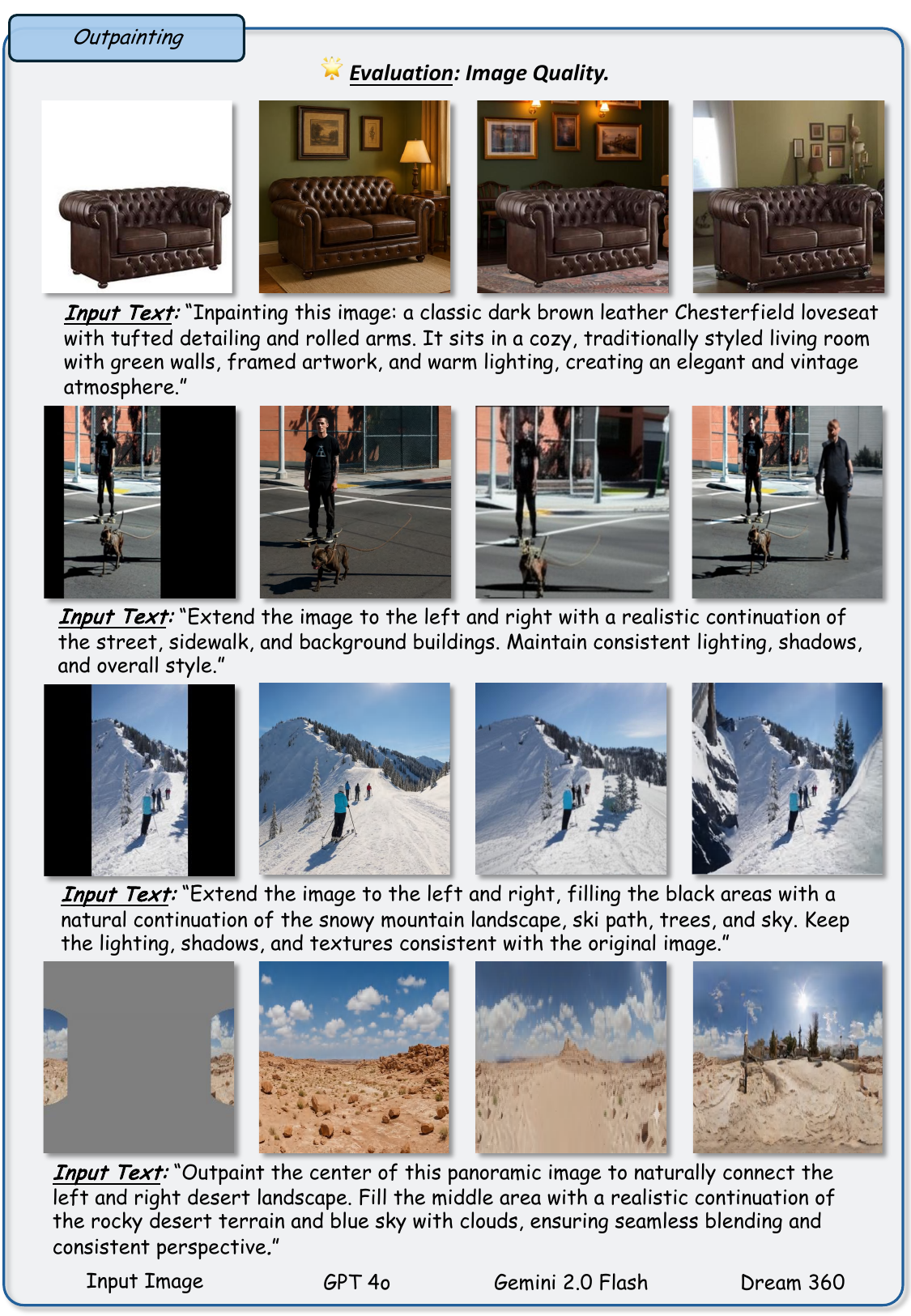}
    \caption{\textit{\textbf{Task:}} Image outpainting, aiming to extend the visual content of an image beyond its original boundaries coherently and realistically.
\textit{\textbf{Setup:}} We compare GPT-4o with Gemini 2.0 Flash~\cite{gemini-2-0-flash}, and some Specialized outpainting methods (SGT+ \cite{sgt}, StrDiffusion \cite{StrDiffusion} and Dream360 \cite{dream360}), evaluating their ability to extend content while maintaining visual consistency in lighting, texture, and semantics.
\textit{\textbf{Observations:}} The Specialized outpainting methods consistently produces the most coherent extensions — for example, it accurately maintains the room’s lighting and decor in the first row, continues architectural lines and street perspective in the second, and creates seamless snowy landscapes in the third. GPT-4o offers plausible structure but often lacks fine detail and texture continuity, such as mismatched snow gradients or missing shadows. Gemini 2.0 Flash performs slightly better in semantic extension than GPT-4o but can introduce lighting inconsistencies and abrupt transitions, particularly in wide scenes like the desert in the final row.}
    \label{fig:low_level_8}
\end{figure}

\begin{figure}[h]
    \centering
    \includegraphics[width=1\textwidth]{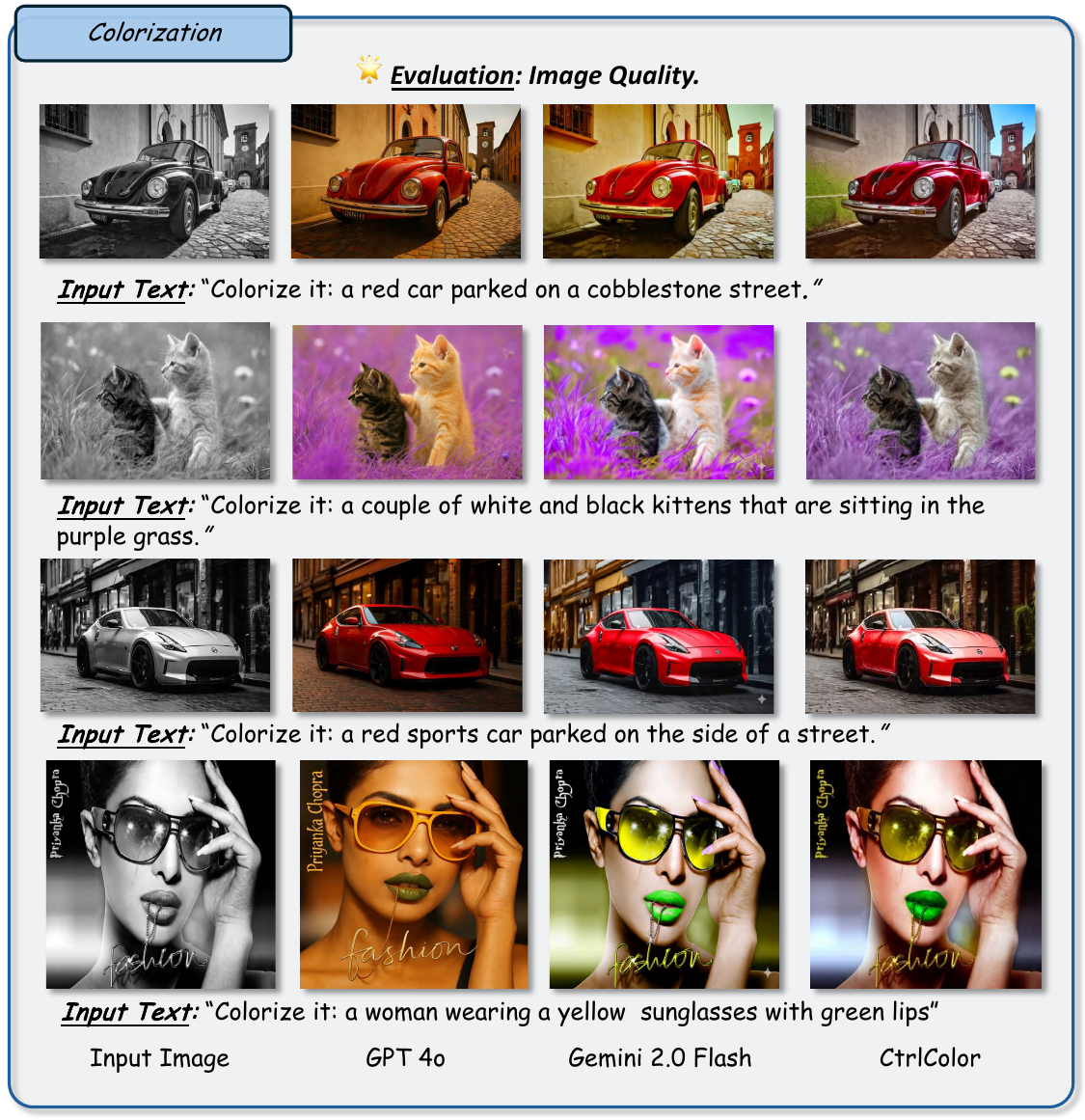}
    \caption{\textit{\textbf{Task:}} Image colorization, aiming to add realistic and semantically consistent color to grayscale images based on textual prompts.
\textit{\textbf{Setup:}} We compare GPT-4o with Gemini 2.0 Flash~\cite{gemini-2-0-flash} and CtrlColor \cite{ctrlcolor}, focusing on their ability to follow instructions and produce visually natural colorized outputs.
\textit{\textbf{Observations:}} CtrlColor performs the best overall, generating vivid and accurate colors that precisely match the prompts—such as green lips and yellow sunglasses in the last row, or the purple grass and kitten hues in the second. GPT-4o provides reasonably faithful colorization but often lacks richness or misinterprets tones (e.g., slightly dull red in the third row or inconsistent purple grass). Gemini 2.0 Flash is more vivid than GPT-4o but tends to oversaturate or produce stylized effects, especially on human features.}
    \label{fig:low_level_9}
\end{figure}

For image inpainting and outpainting in Figure \ref{fig:low_level_7}, \ref{fig:low_level_8}. We compared Gemini 2.0 Flash with the latest inpainting and outpainting methods \cite{StrDiffusion, sgt, latentinpainting, dream360}. Only the missing information needs to be completed, but GPT-4o still changes the undesired content of the image. Although the output image quality is higher, this is not ideal for evaluating the task itself. For human face inpainting, compared to the other two methods, the overall artistic style is more natural.
For the colorization, we choose the latest colorization model CtrlColor \cite{ctrlcolor}. The overall style is somewhat dark in Figure \ref{fig:low_level_9}. Compared to Gemini 2.0 Flash, GPT-4o's colors are more natural and consistent with the style. However, there are some inaccuracies in color control. For example, in the second image, the cat's color is not white as specified in the text. Additionally, GPT-4o still exhibits issues with changes in image content, such as the shape of the human's face in the fourth image.

\begin{figure}[h]
    \centering
    \includegraphics[width=0.88\textwidth]{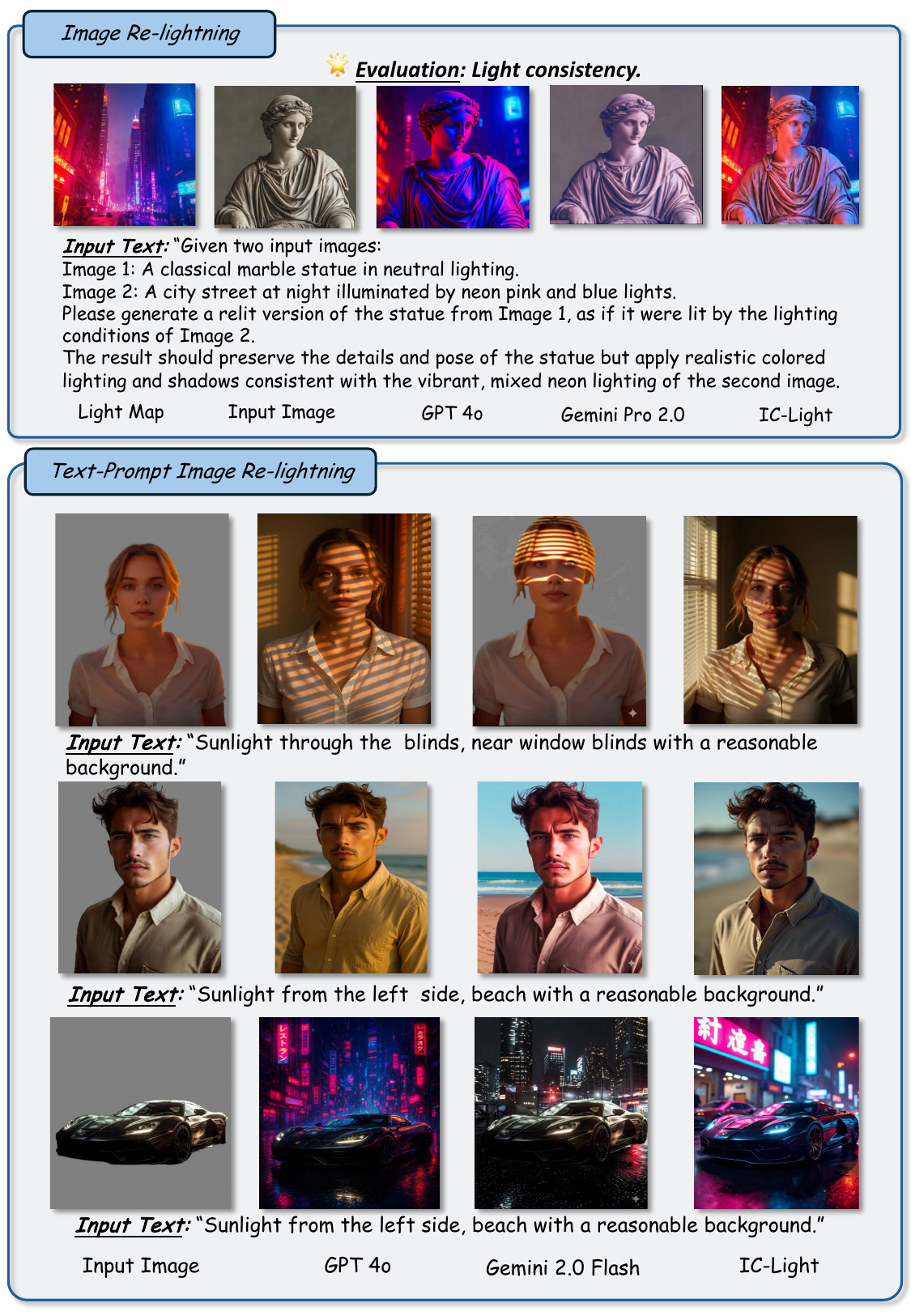}
    \caption{\textit{\textbf{Task:}} Image relighting, aiming to modify the lighting of a given image based on either a reference light map or a textual description, while preserving identity, texture, and spatial consistency.
\textit{\textbf{Setup:}} We compare GPT-4o with Gemini 2.0 Flash~\cite{gemini-2-0-flash} and IC-Light \cite{iclight} on two subtasks: reference-based and text-based relighting. Evaluations focus on lighting realism, directionality, shadow accuracy, and semantic preservation.
\textit{\textbf{Observations:}} IC-Light achieves the most realistic and consistent relighting across both tasks—accurately applying neon lighting from a reference image and generating sharp shadows and natural light from text prompts. Gemini 2.0 Flash preserves content well but produces softer, less directional lighting. GPT-4o offers more vivid lighting than Gemini 2.0 Flash but sometimes lacks shadow accuracy or background coherence.}
    \label{fig:low_level_12}
\end{figure}

For the image re-lighting task in Figure \ref{fig:low_level_12}, GPT-4o performs well in applying realistic lighting and shadows, with natural color tones that match the scene. However, it occasionally struggles with maintaining light consistency, particularly in complex lighting scenarios, such as neon or vibrant lights. Compared to Gemini 2.0 Flash, GPT-4o produces more natural and consistent results, but it doesn't always accurately replicate the lighting effects as seen in the second image, where the neon lighting could have been better captured. IC-Light \cite{iclight} is effective in applying realistic lighting, but tends to lose detail in some complex objects or faces under different light conditions. Overall, GPT-4o is a strong contender for the image re-light task, providing good light consistency but leaving room for improvement in some specific scenarios.

In summary, GPT-4o demonstrates strong performance in various low-level vision tasks, often surpassing Gemini 2.0 Flash in output quality with more natural and visually appealing results. However, it struggles with maintaining pixel consistency and avoiding undesired changes to image content, which are critical for tasks like restoration and inpainting. While its adaptability and realism are impressive, there is room for improvement in precision and task-specific consistency compared to specialized models like InstructIR and IC-Light. 

\clearpage

\subsubsection{Spatial Control}

Spatial control aims to generate visual outputs that not only reflect the content described in the prompt, but also precisely adhere to additional structural conditions (e.g., canny edge maps, depth maps, sketches, poses, and masks). This task evaluates a model’s ability to faithfully align text guidance with visual constraints—an essential capability for real-world creative applications such as illustration, animation, digital content creation, and visual storytelling.

\begin{figure}[h]
    \centering
    \includegraphics[width=1\textwidth]{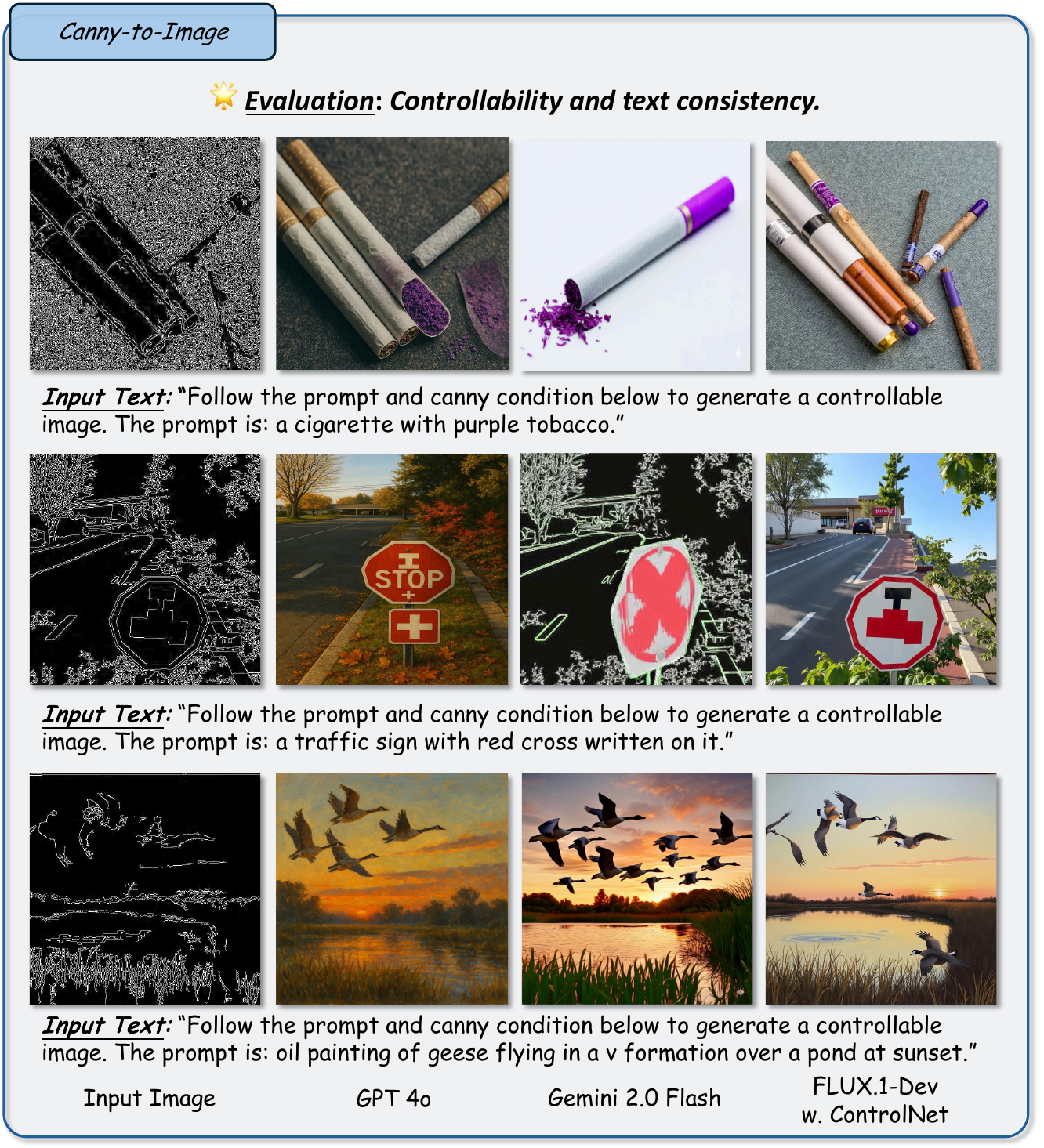}
    \caption{\textit{\textbf{Task:}} Canny-to-Image generation. The goal is to generate prompt-aligned images guided by canny maps. 
\textit{\textbf{Setup:}} Each row shows an input canny map and a text prompt, with outputs from GPT-4o, Gemini 2.0 Flash~\cite{gemini-2-0-flash}, and FLUX.1-Dev w. ControlNet~\cite{flux}. 
\textit{\textbf{Observations:}} GPT-4o performs worse than FLUX.1-Dev~\cite{flux} in structural fidelity, often introducing additional visual details that deviate from the input edge map. However, it produces more semantically aligned and aesthetically pleasing results overall. Compared to Gemini 2.0 Flash, GPT-4o significantly outperforms in both structure preservation and prompt consistency.
}
    \label{fig:control_1}
\end{figure}

\begin{figure}[h]
    \centering
    \includegraphics[width=1\textwidth]{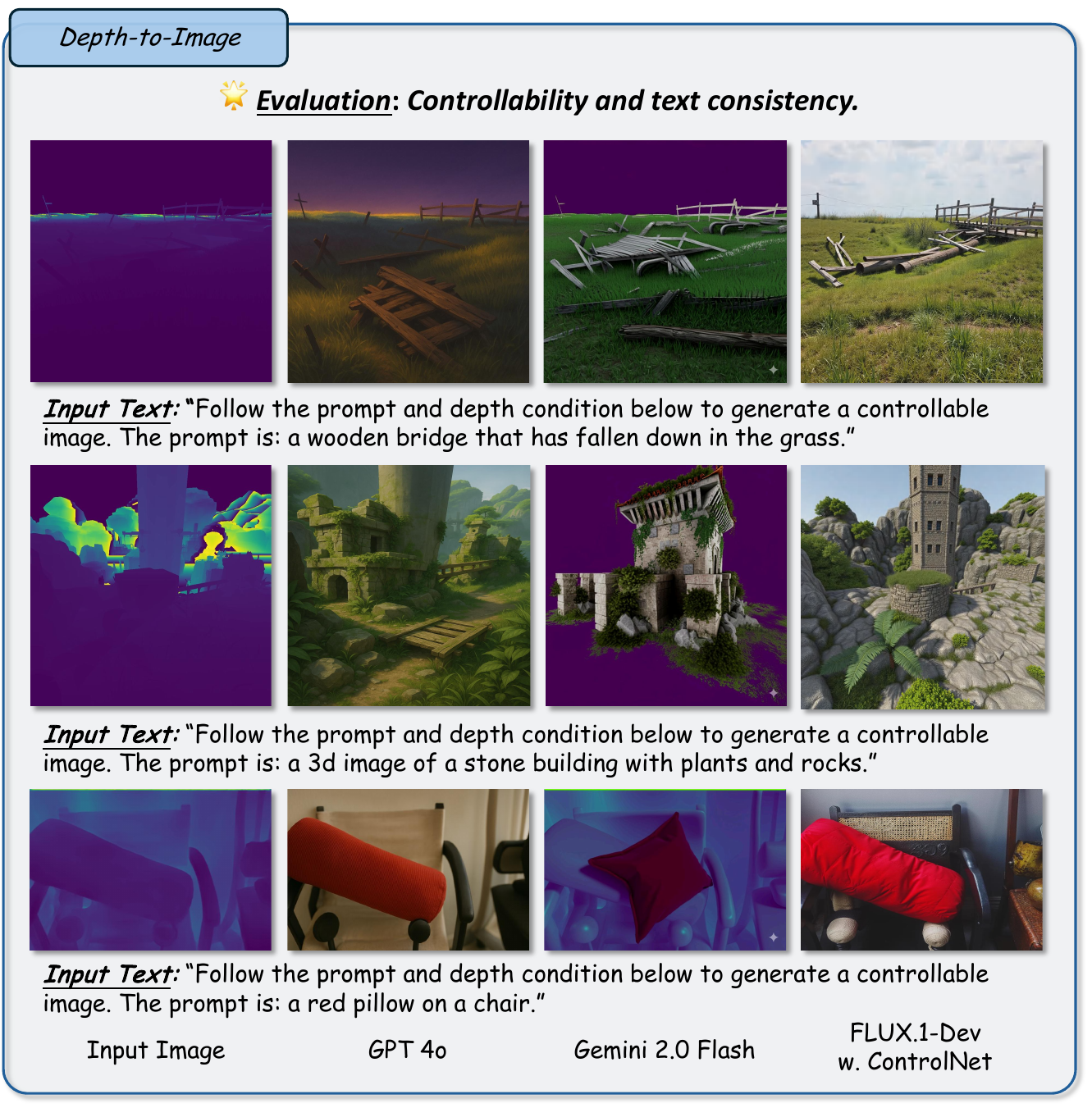}
    \caption{\textit{\textbf{Task:}} Depth-to-image generation, aiming to synthesize controllable and visually coherent images based on a text prompt and a given depth map.
\textit{\textbf{Setup:}} We compare GPT-4o with Gemini 2.0 Flash~\cite{gemini-2-0-flash} and FLUX.1-Dev w. ControlNet~\cite{flux}, focusing on controllability, text-prompt alignment, and the visual quality of generated scenes.
\textit{\textbf{Observations:}} GPT-4o generates visually appealing and stylistically consistent images that align reasonably with text and depth cues—such as the bridge scene and stone ruins with rich lighting and artistic tone. However, its controllability is weaker than FLUX.1-Dev w. ControlNet~\cite{flux}, which shows more precise depth alignment and object placement, as seen in the accurate layout of the bridge and red pillow. GPT-4o leans toward stylized coherence, while FLUX emphasizes photorealism with sharper spatial fidelity. Gemini 2.0 Flash lags behind both, often showing depth misalignment, shape distortion, and weaker semantic grounding.}

    \label{fig:control_2}
\end{figure}

\begin{figure}[h]
    \centering
    \includegraphics[width=1\textwidth]{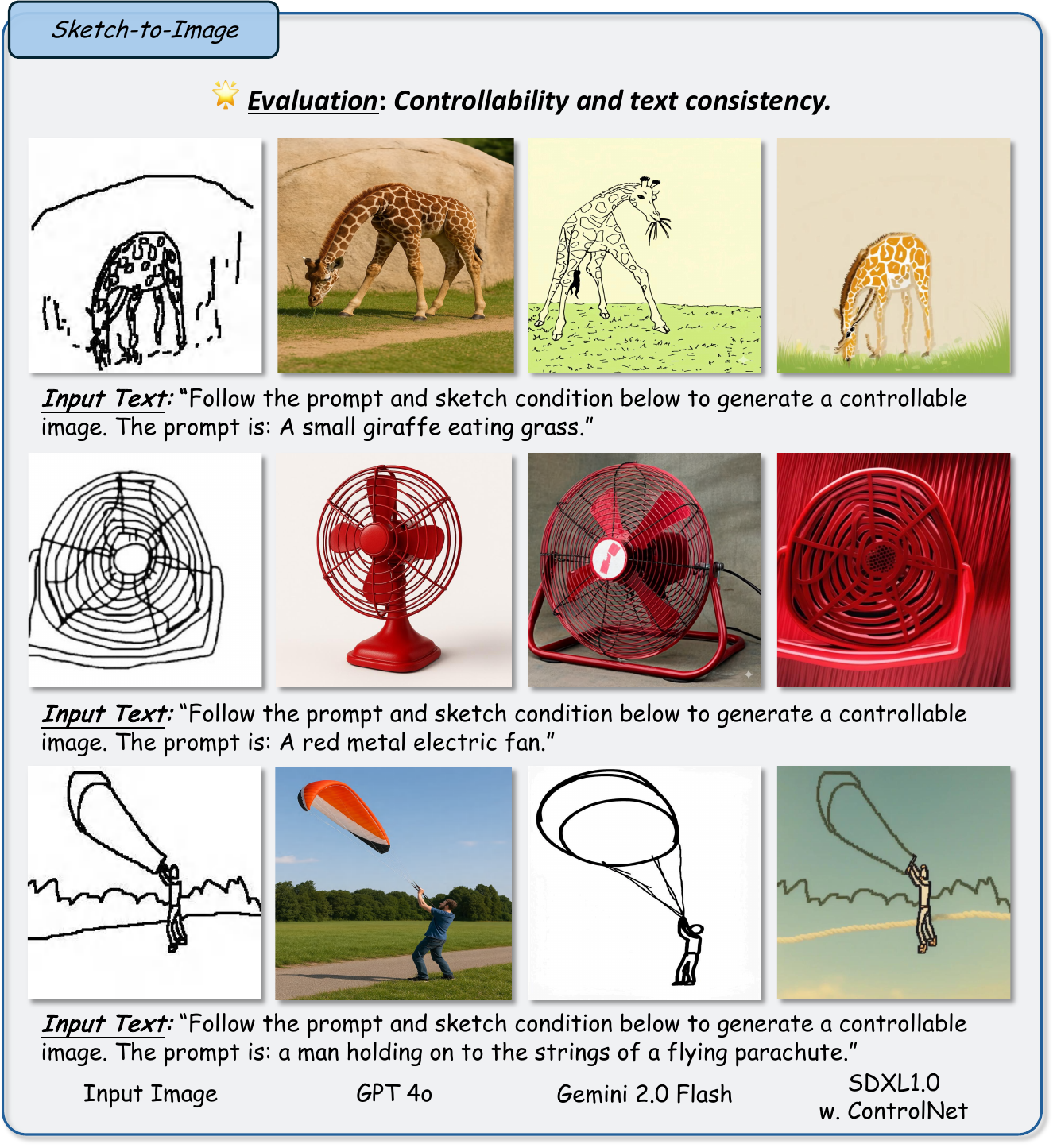}
    \caption{\textit{\textbf{Task:}} Sketch-to-image generation, which requires translating rough line drawings into realistic and semantically accurate images guided by text prompts.
\textit{\textbf{Setup:}} We evaluate GPT-4o against Gemini 2.0 Flash~\cite{gemini-2-0-flash} and SDXL1.0 w. ControlNet~\cite{sdxl}, focusing on how well each model respects the provided sketch while reflecting the described content.
\textit{\textbf{Observations:}} GPT-4o excels at generating lifelike scenes that match the prompt, often delivering visually pleasing and contextually grounded outputs—like the natural posture and setting of the giraffe or the dynamic movement in the parachute example. However, it tends to soften or reinterpret sketch lines, leading to slight mismatches in fine structure. In contrast, SDXL1.0 w. ControlNet~\cite{sdxl} offers stronger adherence to the input sketch, capturing geometric details more accurately (e.g., fan blades and figure outlines), albeit with slightly more synthetic textures. Gemini 2.0 Flash shows limited understanding of both sketch and prompt, often producing less realistic or structurally off-target images.}

    \label{fig:control_3}
\end{figure}

\begin{figure}[h]
    \centering
    \includegraphics[width=1\textwidth]{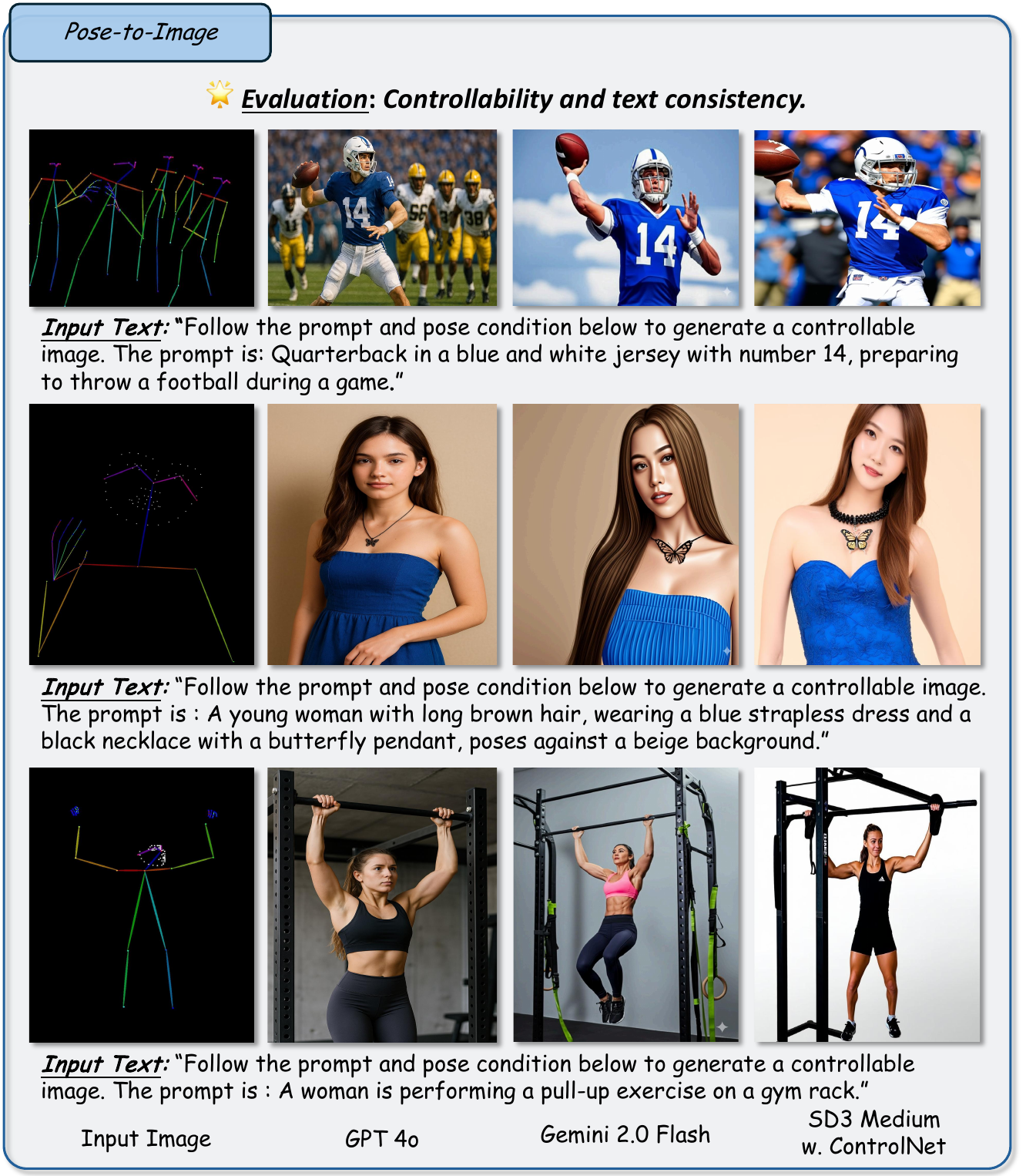}
    \caption{\textit{\textbf{Task:}} Pose-to-image generation, aiming to synthesize realistic images that reflect both the human pose and descriptive prompt.
\textit{\textbf{Setup:}} We benchmark GPT-4o against Gemini 2.0 Flash~\cite{gemini-2-0-flash} and SD3 Medium w. ControlNet~\cite{sd3}, evaluating their ability to follow pose conditions while generating semantically accurate and coherent images.
\textit{\textbf{Observations:}} GPT-4o performs well in complex scenes—such as the football example—where it effectively integrates pose, clothing, and background with strong realism, contextual and pose accuracy. In simpler cases like the pull-up exercise, it shows occasional pose drift, especially in limbs. SD3 Medium w. ControlNet~\cite{sd3} offers better pose fidelity overall, though its visual quality can be inconsistent. Gemini 2.0 Flash underperforms in both structure and coherence, often generating anatomically incorrect or visually weak results. Overall, GPT-4o balances text understanding and generation quality, especially in detailed prompts.
}
    \label{fig:control_4}
\end{figure}

\begin{figure}[h]
    \centering
    \includegraphics[width=1\textwidth]{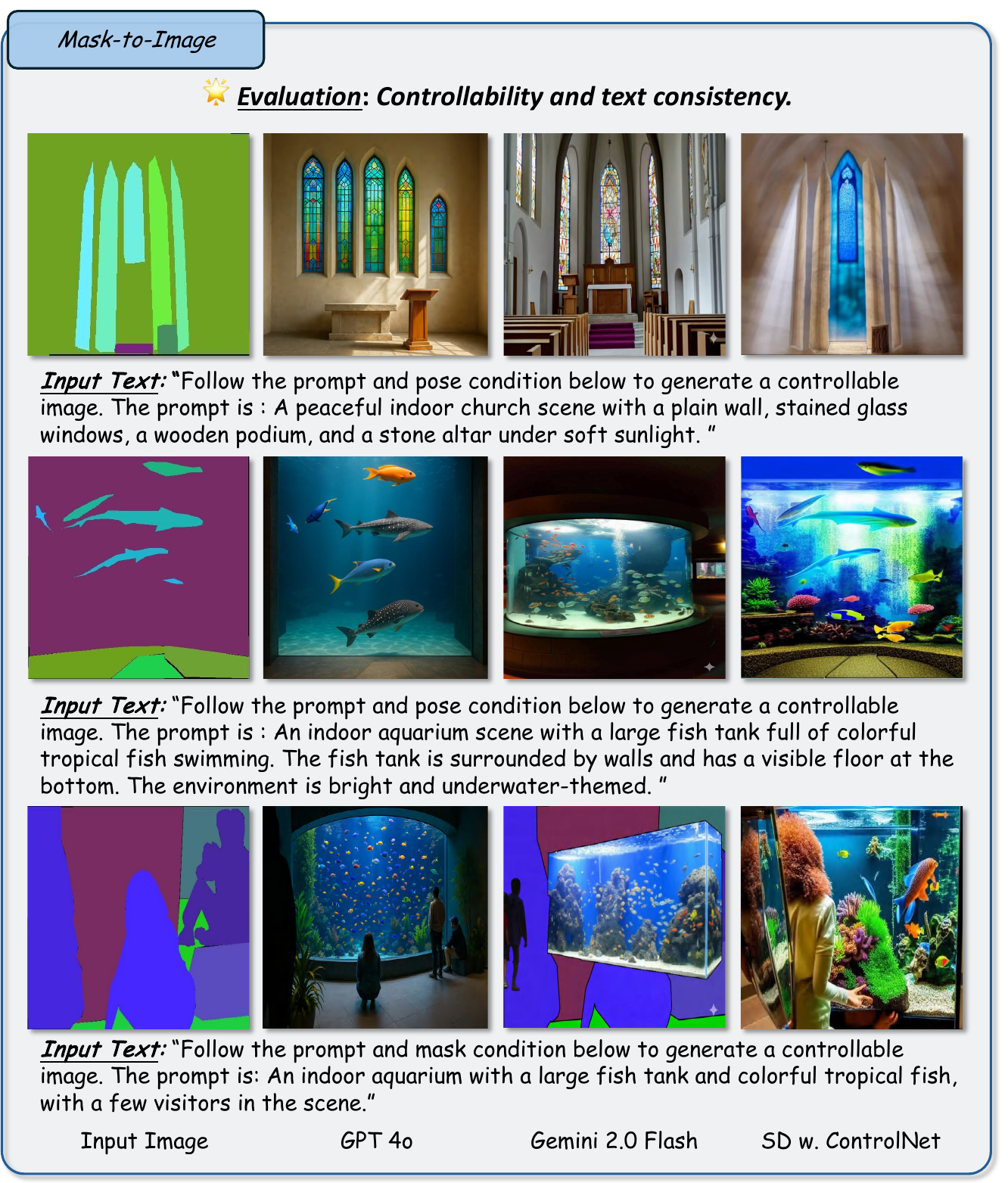}
    \caption{\textit{\textbf{Task:}} Mask-to-image generation, which requires translating semantic segmentation maps and textual prompts into coherent and realistic images.
\textit{\textbf{Setup:}} We compare GPT-4o with Gemini 2.0 Flash~\cite{gemini-2-0-flash} and SD1.5 w. ControlNet~\cite{sd1.5}, focusing on their ability to combine spatial layout from the mask with deeper scene understanding from the prompt.
\textit{\textbf{Observations:}} Compared to previous control tasks, this setting demands more from the model in terms of semantic reasoning and compositional understanding. GPT-4o excels in this regard, producing visually consistent scenes that align with the prompt’s intent—such as the serene church interior and the immersive aquarium setting with visitors. However, in fine-grained spatial control, especially with small or tightly shaped objects like tropical fish, SD1.5 w. ControlNet~\cite{sd1.5} performs better in preserving shape and positioning. Gemini 2.0 Flash continues to struggle in both fidelity and adherence to masks, often missing key scene elements or producing oversimplified outputs.}
    \label{fig:control_5}
\end{figure}

In this section, we examine GPT-4o’s performance across five representative types of controllable conditions: canny, depth, sketch, pose, and mask. For each setting, we compare its outputs with those from Gemini 2.0 Flash~\cite{gemini-2-0-flash} and a strong baseline method using ControlNet-based \cite{controlnet} diffusion backbones (FLUX.1-Dev~\cite{flux}, SDXL1.0~\cite{sdxl}, SD3 Medium~\cite{sd3} or SD1.5~\cite{sd1.5}). The results are illustrated in Figures~\ref{fig:control_1},~\ref{fig:control_2},~\ref{fig:control_3},~\ref{fig:control_4},~\ref{fig:control_5}.

Overall, GPT-4o achieves performance that is on par with ControlNet-based methods in many cases, especially under common or moderately complex conditions. In particular, GPT-4o is capable of handling semantically rich or contextually complex prompts, where its strong foundation model understanding can help preserve both high-level semantics and visual plausibility. This is especially evident in tasks like pose-to-image or mask-to-image, where the structural signal may be sparse or ambiguous.
However, GPT-4o’s strong generative prior can sometimes lead to overly detailed or hallucinated elements, which compromises structural fidelity. For instance, in canny-to-image or depth-to-image tasks that require fine-grained geometric alignment, GPT-4o may deviate from the input layout more noticeably than traditional diffusion-based methods. In contrast, ControlNet exhibits more stable and accurate control in these low-level structure-guided scenarios, making it better suited for applications where spatial accuracy is critical.
That said, ControlNet may struggle in more complex or open-ended cases, such as mask-to-image scenes involving multiple objects or interactions (e.g., aquariums with visitors and fish). In these scenarios, GPT-4o's strong cross-modal understanding partially compensates for its weaker control, offering plausible but not fully precise outputs.
By comparison, Gemini 2.0 Flash lacks robust controllable generation capabilities across all evaluated control types. Its outputs often fail to match either the control condition or the textual prompt, reflecting limited capacity in multimodal alignment and structural grounding.

In summary, GPT-4o demonstrates performance comparable to SOTA methods in most cases, excelling in tasks that require rich semantic understanding and contextual complexity while maintaining a balance between high-level semantics and visual plausibility. Although it may exhibit structural deviations in tasks requiring precise geometric alignment, its strong generative prior gives it an advantage in handling complex or open-ended scenarios. 

\clearpage

\subsubsection{Camera Control}
Although recent visual generative models demonstrate remarkable capabilities in creating high-quality images, generating images with specific camera settings (e.g., bokeh blur parameters, focal length, shutter speed, color temperature) and making further adjustments remains a challenging task. 
We further explore GPT-4o's performance in camera control, evaluating its ability to generate images with desired photographic parameters in text instructions. 
This task is particularly significant as it bridges the gap between artistic creativity and technical precision, enabling users to simulate professional photography techniques and achieve greater control over the visual output. Such advancements have broad applications in fields like photography, cinematography, and visual design.

\begin{figure}[h]
    \centering
    \includegraphics[width=1\textwidth]{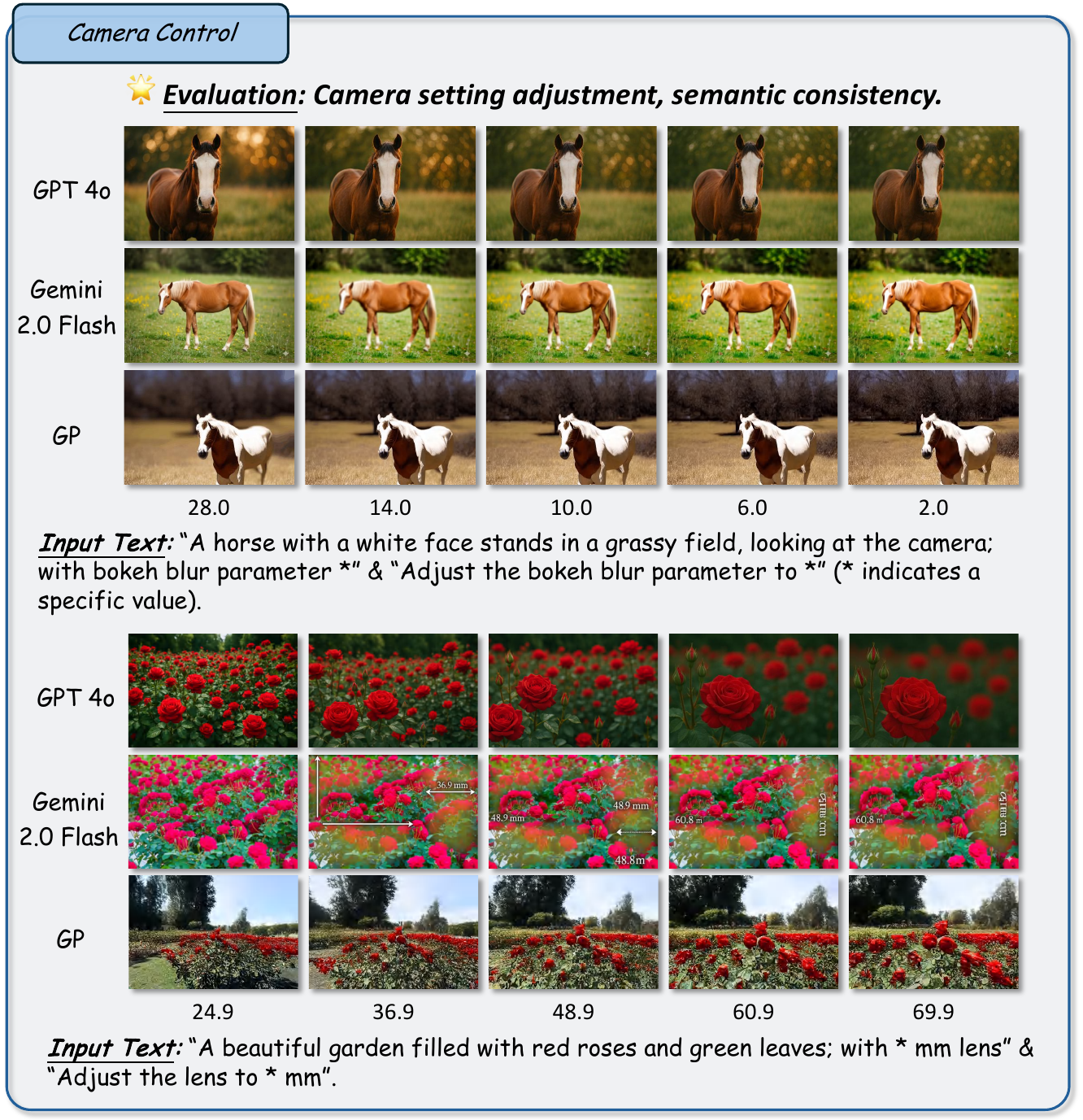}
    \caption{\textit{\textbf{Task:}} Camera control. The goal is to generate images aligned with specific photographic parameters, such as bokeh blur, focal length, shutter speed, and color temperature.  
    \textit{\textbf{Setup:}} Results are based on text prompts collected from \cite{yuan2024generative}, comparing outputs from GPT-4o, Gemini 2.0 Flash~\cite{gemini-2-0-flash}, and Generative Photography (GP)~\cite{yuan2024generative}. Each row includes the input text instructions and corresponding outputs.  
    \textit{\textbf{Observations:}} GPT-4o demonstrates strong performance in controlling bokeh blur, producing visually appealing and parameter-aligned results. However, it shows limitations in handling focal length, occasionally generating inconsistent or less accurate outputs. By contrast, Gemini 2.0 Flash struggles significantly in both aspects, often failing to produce coherent results. Overall, GPT-4o achieves better performance in this task but still requires further refinement to enhance focal length control.}
    \label{fig:custom_3}
\end{figure}

\begin{figure}[h]
    \centering
    \includegraphics[width=1\textwidth]{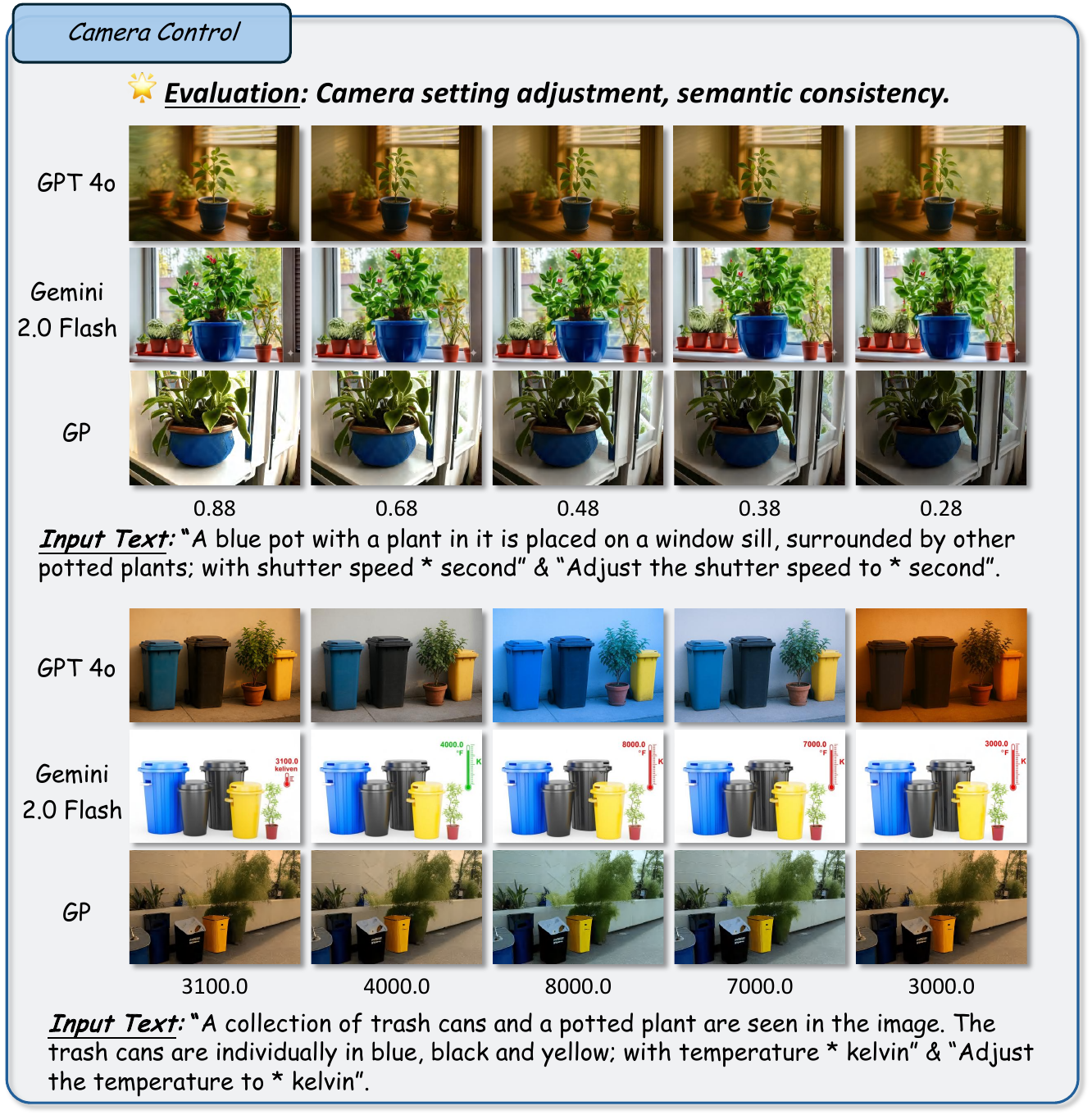}
    \caption{\textit{\textbf{Task:}} Camera control. The goal is to generate images aligned with specific photographic parameters, such as bokeh blur, focal length, shutter speed, and color temperature.  
    \textit{\textbf{Setup:}} Results are based on text prompts collected from \cite{yuan2024generative}, comparing outputs from GPT-4o, Gemini 2.0 Flash~\cite{gemini-2-0-flash}, and Generative Photography (GP)~\cite{yuan2024generative}. Each row includes the input text instructions and corresponding outputs.  
    \textit{\textbf{Observations:}} GPT-4o demonstrates strong performance in controlling color temperature, producing coherent and visually accurate results. However, it struggles with shutter speed, occasionally resulting in inconsistent or unrealistic motion effects. In contrast, Gemini 2.0 Flash fails to consistently handle either parameter, often producing outputs that lack alignment with the desired settings. Overall, GPT-4o outperforms Gemini 2.0 Flash in this task, but further improvements are needed for precise shutter speed control.}
    \label{fig:custom_4}
\end{figure}

Specifically, we collect text prompts from \cite{yuan2024generative}, and compare GPT-4o and Gemini 2.0 Flash~\cite{gemini-2-0-flash} with Generative Photography (GP) \cite{yuan2024generative}.
The results are reported in Figures~\ref{fig:custom_3},~\ref{fig:custom_4}.
We can observe that GPT-4o achieves decent results in controlling bokeh blur parameters and color temperature, demonstrating its strong generalizability to various photographic settings. However, it still falls short in adjusting focal length and shutter speed, occasionally leading to inconsistent visual semantics or incorrect visual effects. By comparison, Gemini 2.0 Flash struggles significantly across all camera control scenarios, failing to produce coherent or accurate outputs that align with the specified photographic parameters, highlighting its limited capability in this domain.

In this task, GPT-4o shows promising potential in camera control, outperforming Gemini 2.0 Flash and achieving competitive results in certain aspects. Nonetheless, there remains room for improvement in handling more complex adjustments, which could further enhance its applicability in professional photography and creative industries.

\clearpage

\subsubsection{In-context Visual Prompting}
The in-context visual prompting tasks aim at understanding and executing specific tasks on new query images by leveraging a pair of task-specific example images and accompanying text instructions. 
Previous works \cite{wang2023context, chen2023improving, lai2024unleashing} have explored this capability in the context of diffusion and autoregressive models, demonstrating its potential in enhancing model adaptability.
The significance of in-context visual prompting lies in its ability to enable models to generalize to novel tasks. 
This approach mirrors human-like learning, where new tasks can be understood and performed by observing relevant examples. 
This capability has broad implications across various domains, and paves the way for more flexible and efficient paradigms capable of adapting to a wide range of specific tasks.

\begin{figure}[h]
    \centering
    \includegraphics[width=0.9\textwidth]{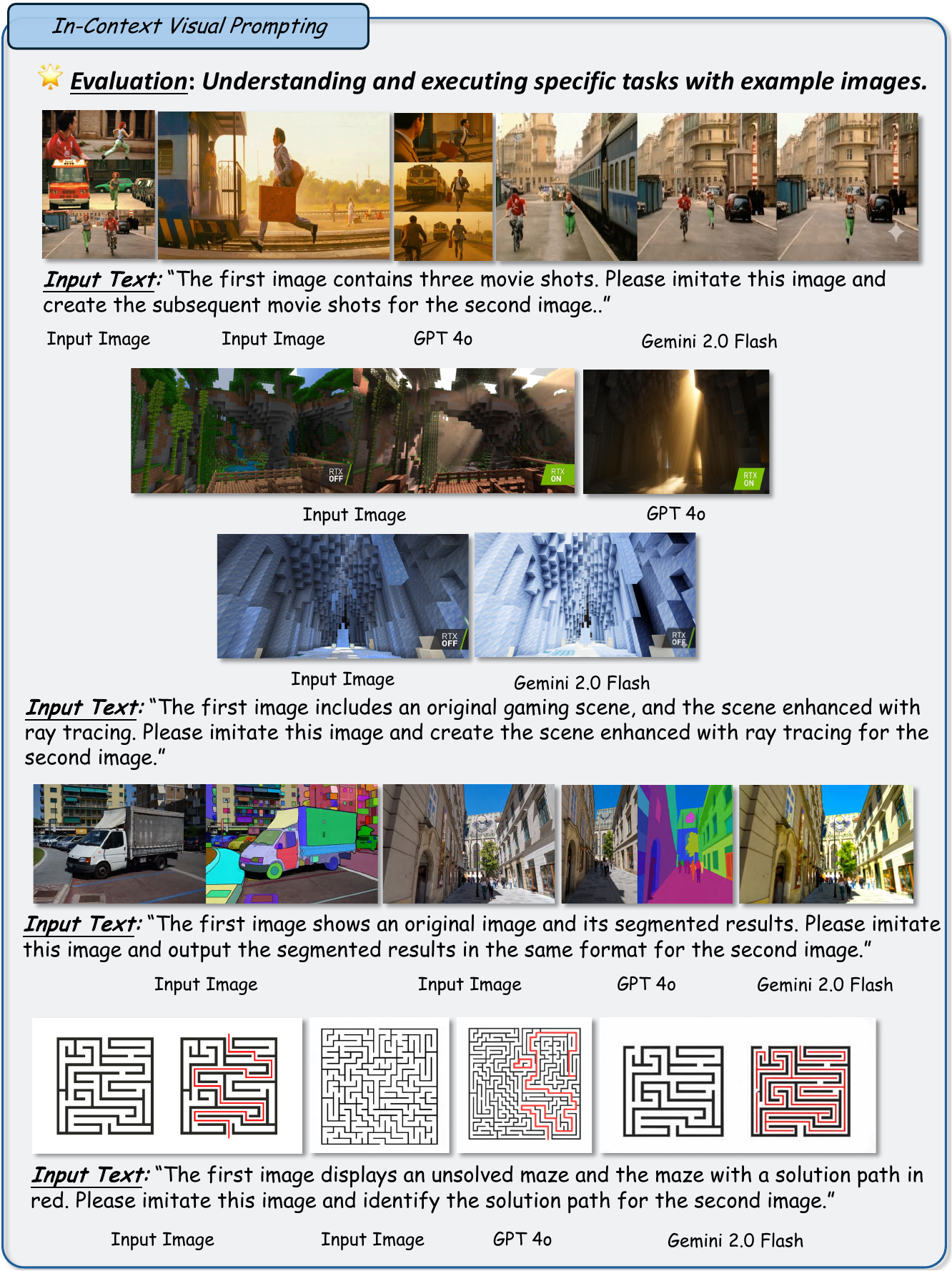}
    \caption{\textit{\textbf{Task:}} In-context visual prompting. The goal is to perform specific visual tasks on new query images based on task-specific example images and text instructions.  
    \textit{\textbf{Setup:}} Four representative tasks are evaluated: movie-shot generation, ray-tracing rendering, overlaid mask visualization, and maze solving. Each row includes example images, query images, and the corresponding outputs.  
    \textit{\textbf{Observations:}} GPT-4o excels in movie-shot generation and ray-tracing, producing coherent outputs but lacks consistency in visual semantics. It fails with overlaid mask visualization and maze solving, showing limits in complex task integration. While promising for in-context visual prompting, it needs refinement for more complex and reasoning-intensive tasks.} 
    \label{fig:custom_5}
\end{figure}

We curate four representative tasks to evaluate the performance of GPT-4o in in-context visual prompting. These tasks are designed to assess the model's ability to understand and adapt to specific visual tasks based on provided examples and guidance, including:
\noindent
\begin{itemize} 
\item \textbf{Movie-Shot Generation:} A three-shot image collected from \cite{huang2024context} is provided as an example, and the model is instructed to follow this format to generate similar movie shots for the query image.
\item \textbf{Ray-Tracing Rendering:} An example gaming scene is provided with and without ray tracing, and the model is expected to render a ray-traced version of the query image.
\item \textbf{Overlaid Mask Visualization:} The model receives an original image accompanied by its corresponding segmented results from \cite{kirillov2023segment} and is tasked with outputting the segmented results in the same format for the query image.
\item \textbf{Maze Solving:} A maze and its corresponding solution path are provided as examples, and the model is required to draw the solution path for a new maze presented in the query image.
\end{itemize}
All the results are illustrated in Figure~\ref{fig:custom_5}.
Compared with Gemini 2.0 Flash~\cite{gemini-2-0-flash}, GPT-4o demonstrates promising performance in movie-shot generation and ray-tracing rendering tasks, showcasing its ability to follow example formats and generate visually coherent outputs. However, it still struggles with maintaining consistent visual semantics across the generated outputs.
For the overlaid mask visualization task, GPT-4o falls short in effectively executing the instructions. The result fails to adhere to the required format, indicating that the model's ability to process and generate complex outputs remains limited.
For maze solving, a task that demands advanced visual reasoning and logical inference, GPT-4o struggles significantly.
This highlights the challenges in combining higher-level reasoning with visual generation capabilities, suggesting that more sophisticated reasoning mechanisms are needed for tasks of this nature.

In summary, GPT-4o shows considerable potential in in-context visual prompting, while it still underperforms in certain difficult tasks.
These observations suggest that further advancements are necessary to enhance its generation and reasoning capabilities for more complex and diverse visual tasks.

\clearpage

\subsection{Image-to-3D Tasks}

We evaluate the 3D understanding capabilities from 2D images of GPT-4o across three tasks: 2D image-to-3D modeling, 2D UV map-to-3D rendering, and novel view synthesis.

\subsubsection{Image to 3D modeling}

Generating 3D models from monocular images boosts a wide range of applications, including augmented reality, virtual reality, and the gaming industry. This capability not only facilitates the content creation process but also mitigates the reliance on specialized 3D artists for creating 3D assets, which is more time- and cost-effective. Therefore, there is a growing research interest in generating 3D models from 2D images.
Early methods on image-to-3D employ the learning-based approaches for single-view reconstruction~\cite{mescheder2019occupancy, niemeyer2020differentiable, wang2018pixel2mesh, pan2019deep}. Recent works leverage the diffusion model prior to perform image-conditioned 3D generative modeling~\cite{liu2023zero, liu2023one, qian2023magic123, xu2024instantmesh}.

\begin{figure}[h]
    \centering
    \includegraphics[width=1\textwidth]{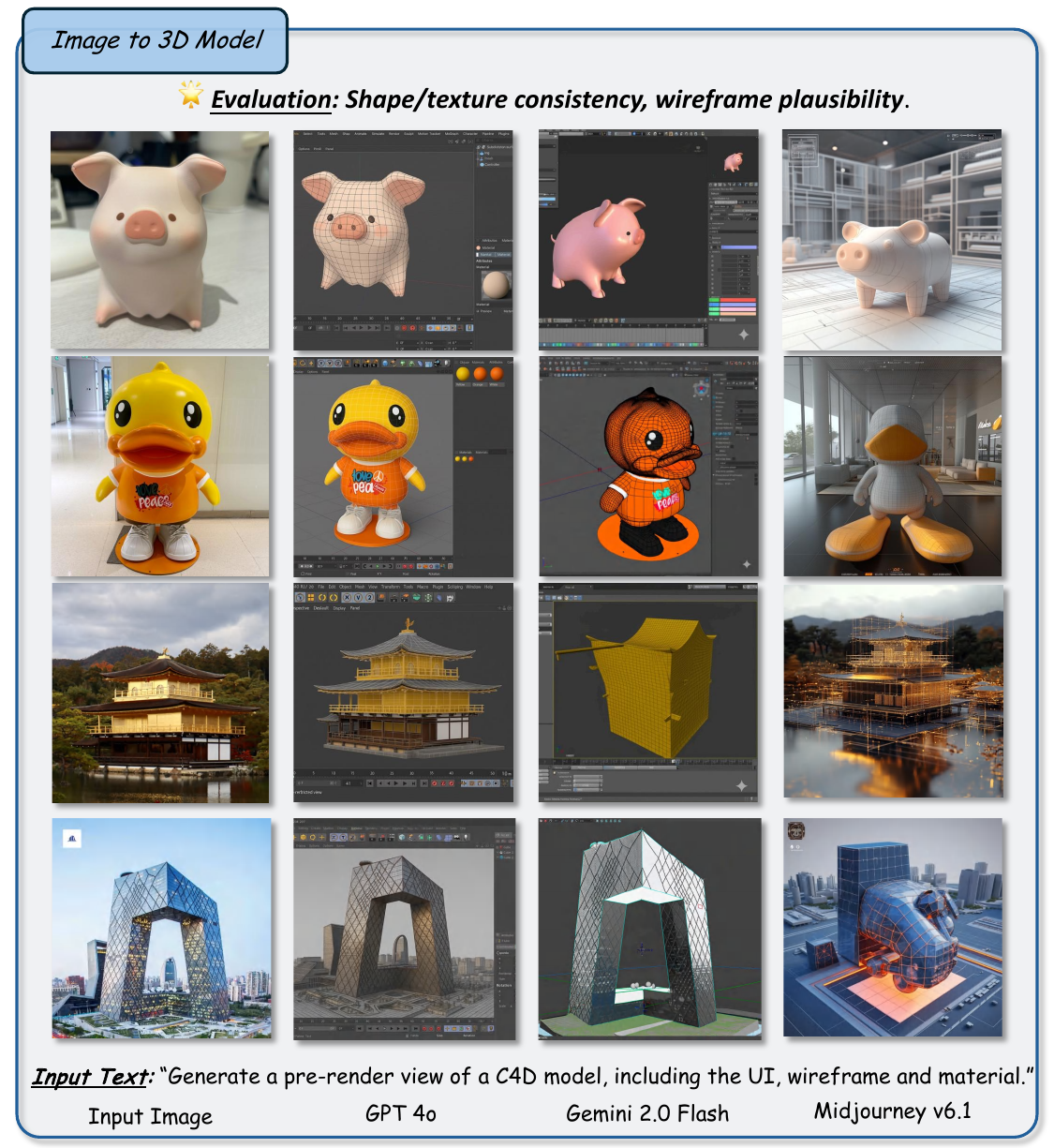}
    \caption{\textit{\textbf{Task:}} Image-to-3D model rendering. Evaluate the 3D modeling ability given a 2D image. \textit{\textbf{Setup:}} Each row shows an input image and a text prompt with outputs from GPT-4o, Gemini 2.0 Flash~\cite{gemini-2-0-flash}, and Midjourney v6.1~\cite{Midjourney}. \textit{\textbf{Observation:}} GPT-4o can generate better 3D model rendering with consistent shape, texture, and plausible wireframe than Gemini 2.0 Flash and Midjourney v6.1.}
    \label{fig:3D_1}
\end{figure}

In this section, we investigate the potential of GPT-4o for 3D modeling from 2D images. We begin by prompting GPT-4o to generate a Cinema 4D modeling interface to test its ability to produce coherent representations of structure, material, and wireframe based on the input image. As shown in Figure~\ref{fig:3D_1}, GPT-4o can generate high-quality 3D model renderings within the application interface. Notably, the generated models exhibit clear wireframes and textures consistent with the input images. In contrast, Gemini 2.0 Flash and Midjourney v6.1 fail to achieve comparable results under the same conditions, which produce inconsistent modelings. We then prompt the GPT-4o to generate corresponding 3D object and material files in .obj and .mtl formats to further evaluate its understanding of the underlying structure in the rendered images. However, the output 3D models are coarse and inconsistent with input images, indicating that although GPT-4o can produce visually coherent 3D renderings, its capability to transform these into accurate and usable 3D object files remains limited. Additionally, Gemini 2.0 Flash and Midjourney v6.1 do not support exporting 3D models.

\subsubsection{UV Map to 3D rendering}

UV maps are 2D images that store texture information for 3D models. In 3D modeling, geometric data is represented in 3D space, while texture data is defined in a 2D texture space. UV mapping is the process of projecting a 2D UV map onto a 3D model, accurately aligning texture with geometry. The UV mapping process can evaluate models' capability for 3D perception and spatial understanding. Moreover, this task has broad applications in design, helping to reduce the burden on designers to create product renderings from 2D maps manually and provide useful references.

\begin{figure}[h]
    \centering
    \includegraphics[width=1\textwidth]{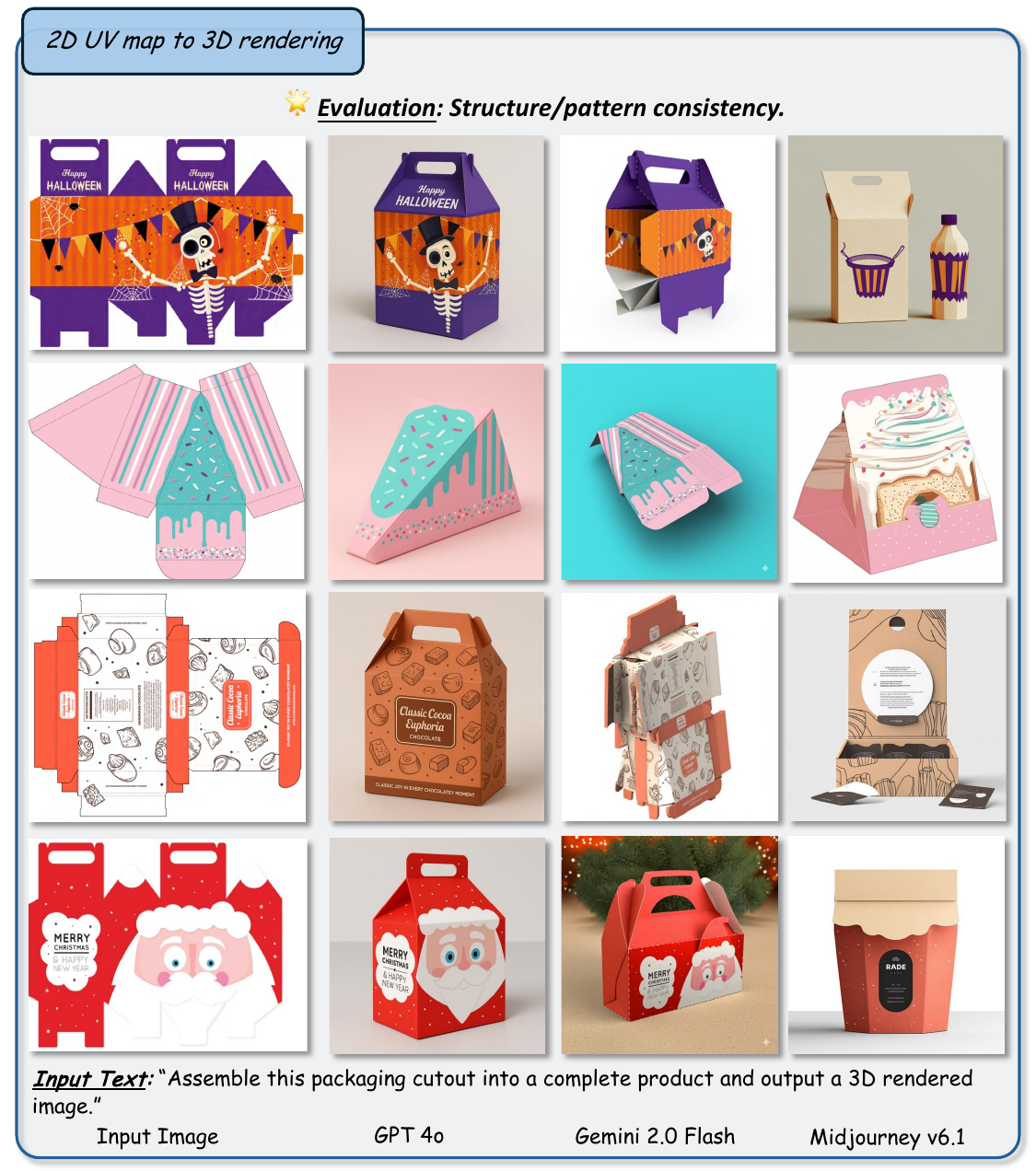}
    \caption{\textit{\textbf{Task:}} 2D UV map to 3D rendering. Evaluate the 3D perception and spatial understanding ability. \textit{\textbf{Setup:}} Each row shows an input image and a text prompt with outputs from GPT-4o, Gemini 2.0 Flash~\cite{gemini-2-0-flash}, and Midjourney v6.1~\cite{Midjourney}. \textit{\textbf{Observation:}} GPT-4o can generate better 3D renderings based on 2D maps than Gemini 2.0 Flash and Midjourney v6.1. However, structure and pattern inconsistencies still exist among these three models.}
    \label{fig:3D_2}
\end{figure}

As shown in Figure~\ref{fig:3D_2}, GPT-4o exhibits a superior ability to generate consistent 3D renderings from 2D maps compared to Gemini 2.0 Flash and Midjourney v6.1. However, some outputs remain unsatisfactory, displaying inconsistencies in patterns and structure (see row 3 in Figure~\ref{fig:3D_2}). Gemini 2.0 Flash struggles to correctly wrap the 3D model, though it maintains pattern consistency. Midjourney v6.1 tends to introduce additional, imagined features, which reduce controllability in this task.

\subsubsection{Novel View Synthesis}

From a monocular view, humans can imagine an object's 3D shape and appearance since humans have collected enough prior knowledge for different objects throughout their daily lives. This ability to infer novel views of objects is essential for a wide range of tasks, from object manipulation to artistic creation such as painting. Early works achieve image-to-3D reconstruction using category-specific priors or large-scale pre-training~\cite{huang2022planes, pavlakos2019expressive, reizenstein2021common, gao2022get3d, zuffi2018lions}. Recent studies have shown that large diffusion models contain rich 3D prior information of the visual world, enabling them to perform novel view synthesis~\cite{liu2023zero, liu2023one, qian2023magic123, long2024wonder3d}. These novel views can then be used for zero-shot 3D reconstruction using different 3D representations such as NeRF~\cite{mildenhall2021nerf}, mesh, or SDF.

\begin{figure}[h]
    \centering
    \includegraphics[width=1\textwidth]{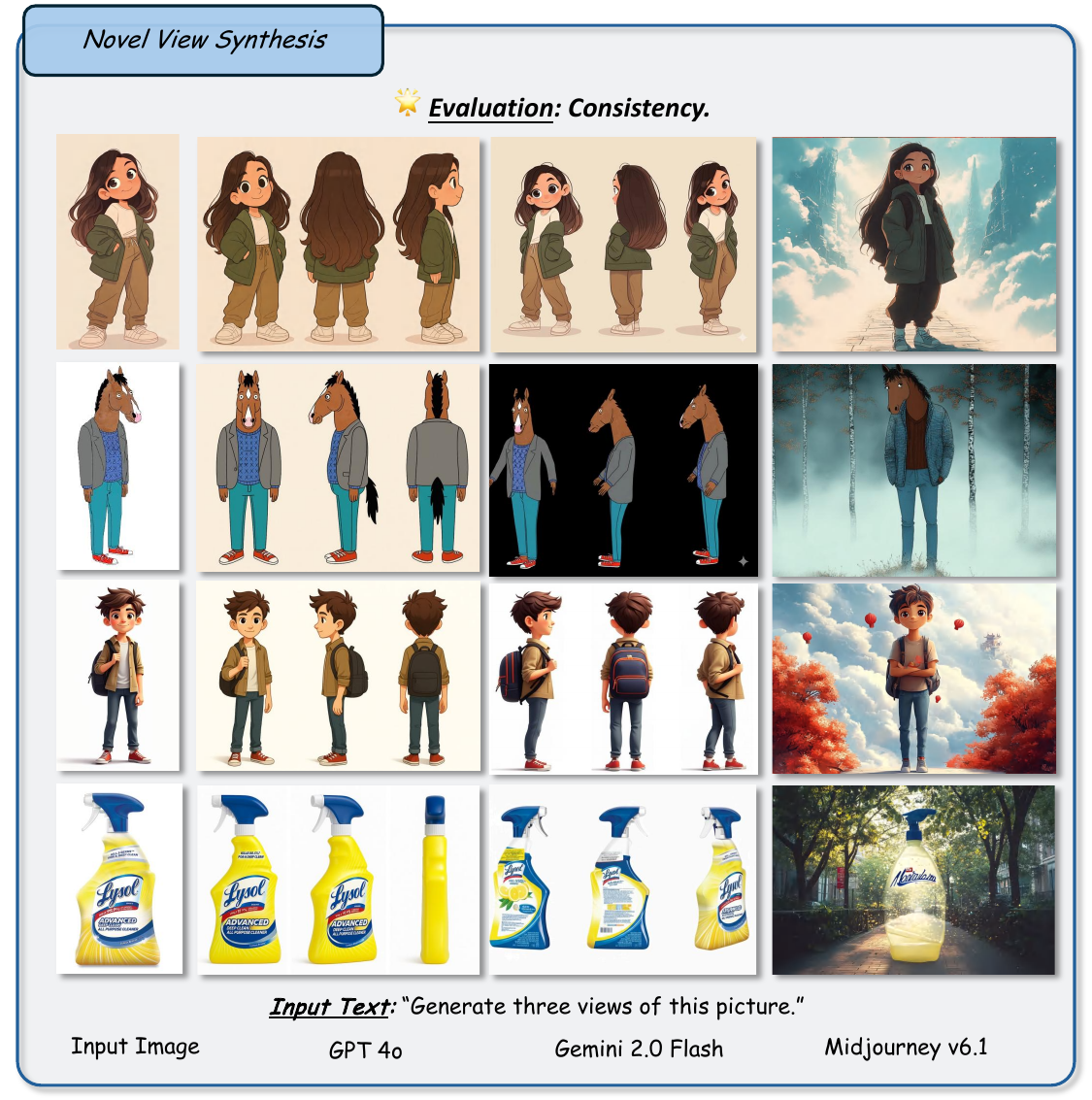}
    \caption{\textit{\textbf{Task:}} Novel view synthesis. Evaluate the 3D perception and spatial understanding ability. \textit{\textbf{Setup:}} Each row shows an input image and a text prompt with outputs from GPT-4o, Gemini 2.0 Flash~\cite{gemini-2-0-flash}, and Midjourney v6.1~\cite{Midjourney}. \textit{\textbf{Observation:}} GPT-4o can generate better style and structure-consistent novel views for both artistic painting and asymmetric objects.}
    \label{fig:3D_4}
\end{figure}

In this section, we evaluate the ability of GPT-4o for novel view synthesis on objects with artistic styles and asymmetric geometry. As shown in Figure~\ref{fig:3D_4}, for artistically styled objects, GPT-4o and Gemini 2.0 Flash largely preserve structural consistency with the input image, although they may change some elements or fine details. For the asymmetric object, GPT-4o can preserve the object scale and size better than Gemini 2.0 Flash. However, Midjourney v6.1 fails to generate consistent novel views, instead producing visually appealing images that do not align with the given prompt of this task.

\clearpage

\subsection{Image-to-X Tasks}
\label{sec:image_to_x}

In this section, we further evaluate both GPT-4o and Gemini 2.0 Flash for several dense image understanding tasks, including segmentation-related tasks, depth estimation, normal estimation, matting, salient object detection, edge detection, layout detection, text detection, and object tracking.

\subsubsection{Image Segmentation}
\label{sub_sec:image_seg}

Image segmentation tasks group pixels of the given image or video into semantic regions. It is a fundamental problem in computer vision and involves numerous real-world applications, such as robotics, automated surveillance, and image/video editing. With the development of recent deep learning methods, this domain has achieved rapid progress.
Early works mainly adopt CNN-based methods with large kernels or respective fields.
Recently, transformer-based methods have also worked well and surpassed previous CNN-based methods on various benchmarks.
In particular, we test three segmentation tasks, including referring segmentation, semantic segmentation, and panoptic segmentation.

\begin{figure}[h]
    \centering
    \includegraphics[width=0.95\textwidth]{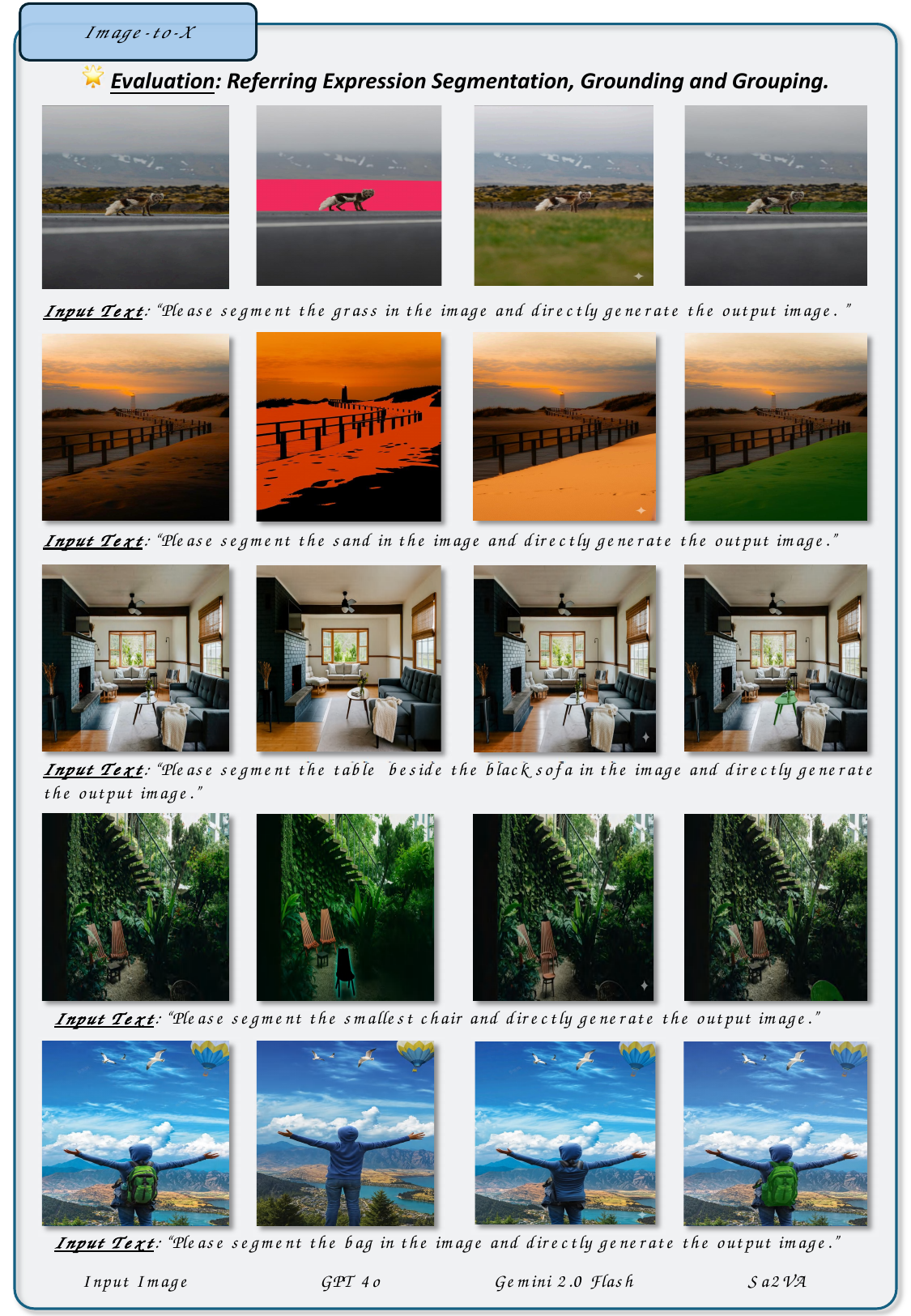}
    \caption{\textit{\textbf{Task:}} Image to X: Referring expression segmentation. Evaluate the grounding and grouping ability.
    \textit{\textbf{Setup:}} Each row shows an input image and a text prompt with outputs from GPT-4o, Gemini 2.0 Flash~\cite{gemini-2-0-flash}, and Sa2VA~\cite{sa2va}.
    \textit{\textbf{Observation:}} These examples indicate that current GPT-4o has weak pixel-level grounding ability.}
    \label{fig:res_seg}
\end{figure}

\noindent
\textbf{Referring Segmentation}. This task outputs the corresponding mask according to the input texts, and the goal is to test the pixel-level grounding ability of the model. 
In Figure~\ref{fig:res_seg}, we compare GPT-4o, Gemin 2.0 Flash and recent state-of-the art method, Sa2VA~\cite{sa2va} (8B model \footnote{https://huggingface.co/ByteDance/Sa2VA-8B}). 
We show five open-world test cases. For the first two cases, GPT-4o shows the coarse localization ability on the background region. For example, it can mark the grass region despite the unfavorable boundaries.
However, compared to the SOTA method, Sa2VA, GPT-4o mistakenly merges both large regions. 
In the third row, both GPT-4o and Gemini 2.0 Flash cannot perform grounding with complex text inputs. 
In the fourth row, all models perform badly.
GPT-4o generates an unseen chair in the images while Gemin 2.0 Flash performs image editing functions by replacing the smallest chair with a normal chair.
Sa2VA also segments the wrong object (the nearest chair).
In the last example, GPT-4o also cannot segment smaller objects (``bag'').
For all examples, both GPT-4o and Gemini 2.0 Flash modify the image contents.
These examples indicate that GPT-4o has weak pixel grounding ability.

\begin{figure}[h]
    \centering
    \includegraphics[width=1\textwidth]{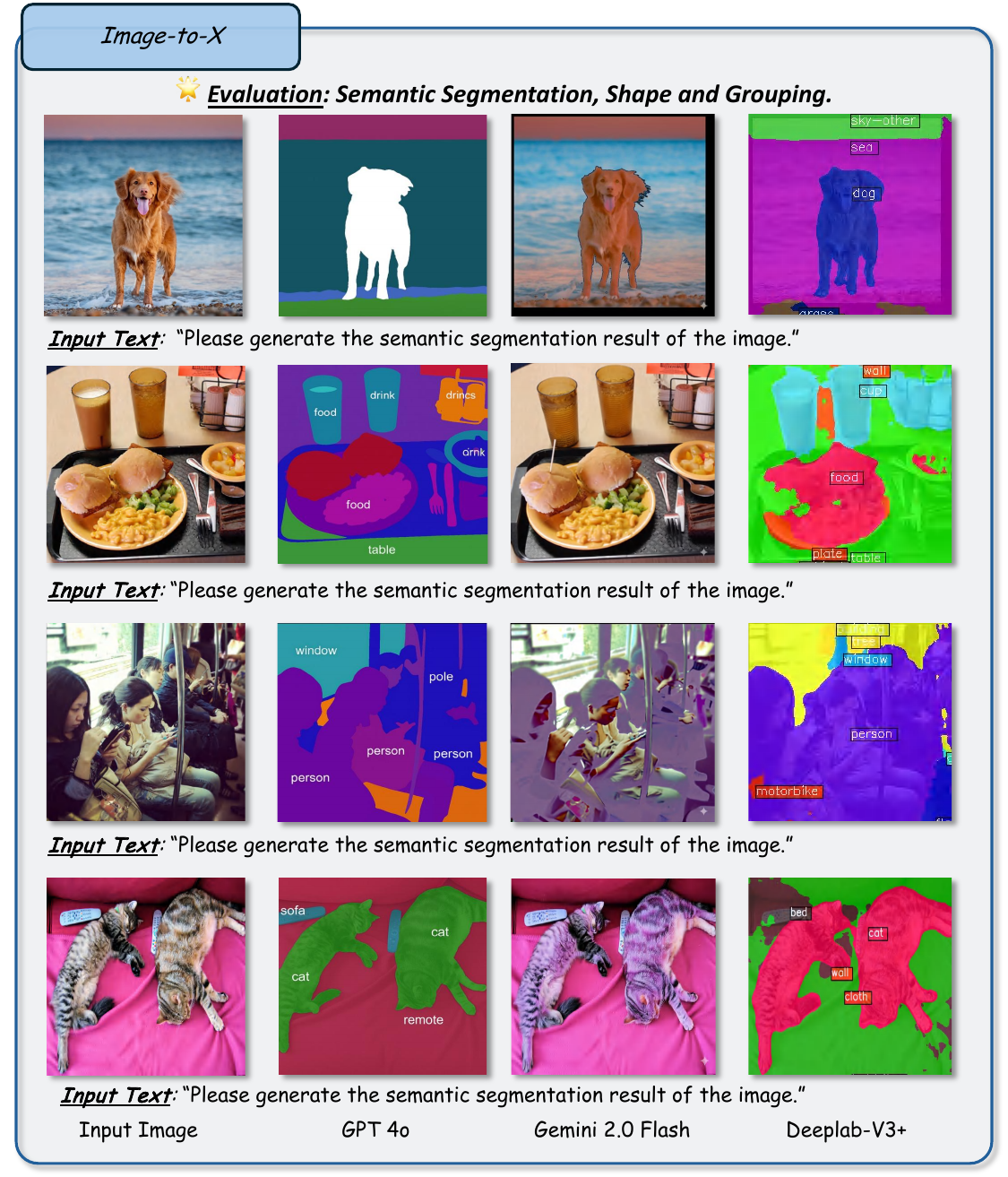}
    \caption{\textit{\textbf{Task:}} Image to X: Semantic segmentation. Evaluate the shape and grouping ability.
    \textit{\textbf{Setup:}} Each row shows an input image and a text prompt with outputs from GPT-4o, Gemini 2.0 Flash~\cite{gemini-2-0-flash}, and Deeplab-V3+~\cite{deeplab-v3}.
    \textit{\textbf{Observation:}} Compared with Gemin-2.0, the mask quality of GPT-4o is good. However, there are still huge gaps in the standard semantic segmentation format.}
    \label{fig:semseg}
\end{figure}

\noindent
\textbf{Semantic Segmentation.} Semantic segmentation assigns each pixel a semantic label, which is one basic vision task. 
In Figure~\ref{fig:semseg}, we show several test cases on the semantic segmentation task.
In particular, we adopt Deeplab-V3+~\cite{deeplab-v3} (ResNet101 as backbone, trained on Pascal-Context) as one expert model for reference.
Surprisingly, the mask quality of GPT-4o is good on four examples, even comparable with an expert model, Deeplab-V3+.
During the testing, we find the texts may be randomly appended to the masks. 
This is why the first row differs from the remaining examples. 
For the second and third examples, GPT-4o misaligns the text and mask regions.
Compared to Gemin 2.0 Flash, GPT-4o has a much stronger ability in semantic segmentation, particularly for mask shape.
However, there is still a lot of room for this task, including a unified semantic segmentation format, enhanced text and mask alignments, and more correct mask labels.

\begin{figure}[h]
    \centering
    \includegraphics[width=0.95\textwidth]{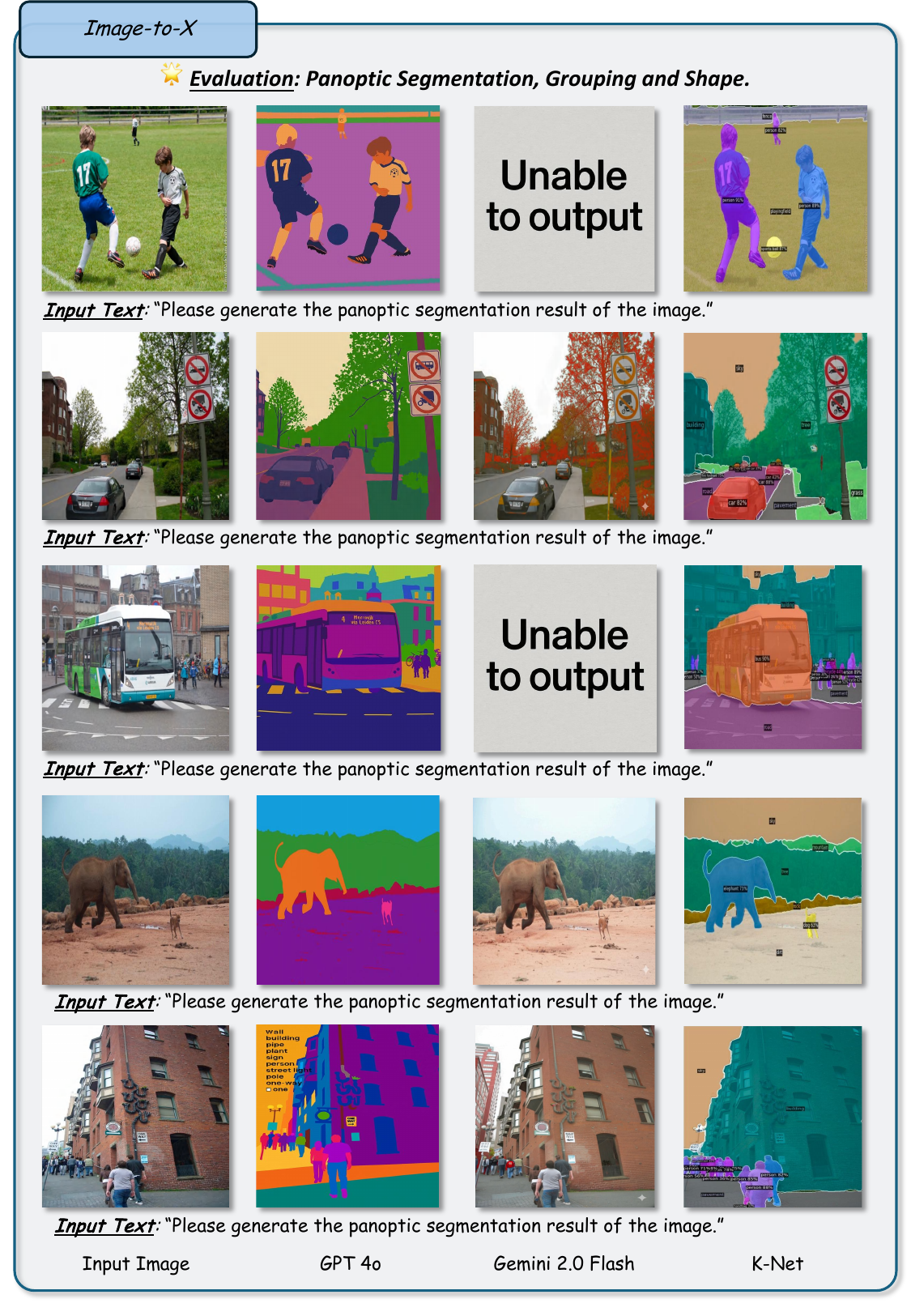}
    \caption{\textit{\textbf{Task:}} Image to X: Panoptic segmentation. Evaluate the shape and grouping ability.
    \textit{\textbf{Setup:}} Each row shows an input image and a text prompt with outputs from GPT-4o, Gemini 2.0 Flash~\cite{gemini-2-0-flash}, and K-Net~\cite{k-net}.
    \textit{\textbf{Observation:}} GPT-4o can understand the panoptic segmentation task, while Gemini 2.0 Flash cannot do this task in the first and third cases.}
    \label{fig:panseg}
\end{figure}

\noindent
\textbf{Panoptic Segmentation.} This task assigns the foreground region a semantic label and assigns one mask label and one instance ID to each instance, which is a unified task format of semantic segmentation and instance segmentation.
In Figure~\ref{fig:panseg}, we compare the panoptic segmentation ability of GPT-4o, Gemini 2.0 Flash, and one expert model, K-Net~\cite{k-net}(trained on the COCO panoptic segmentation dataset, with ResNet50 as backbone).
Overall, the mask shapes of GPT-4o are good. 
The model can understand the panoptic segmentation task, while the Gemini 2.0 Flash cannot do this task in the first and third cases.
However, the spatial locations have been changed for all cases. 
The generated masks are in part-whole formats and are even finer-grained than K-Net.
For example, in the first example, the jersey number (17) of the person and the hair of the people are also marked.
Meanwhile, we also find a similar issue: several examples have text, while several do not have text, even though they adopt the same text prompt.
In addition, GPT-4o can distinguish different instances with different colors, despite most of them not being good (see the last example).

\clearpage

\subsubsection{Edge Detection}
\label{sub_sec:edge_det}

\noindent
\textbf{Edge Detection.} As a classic vision task, edge detection aims to identify the boundaries or edges of objects within an image. 
These edges represent the locations with significant changes in image intensity, color, or other visual features. Common edge detection operators include the Sobel, Prewitt, and Canny operators. Recent works adopt deep learning-based approaches.

\begin{figure}[h]
    \centering
    \includegraphics[width=1\textwidth]{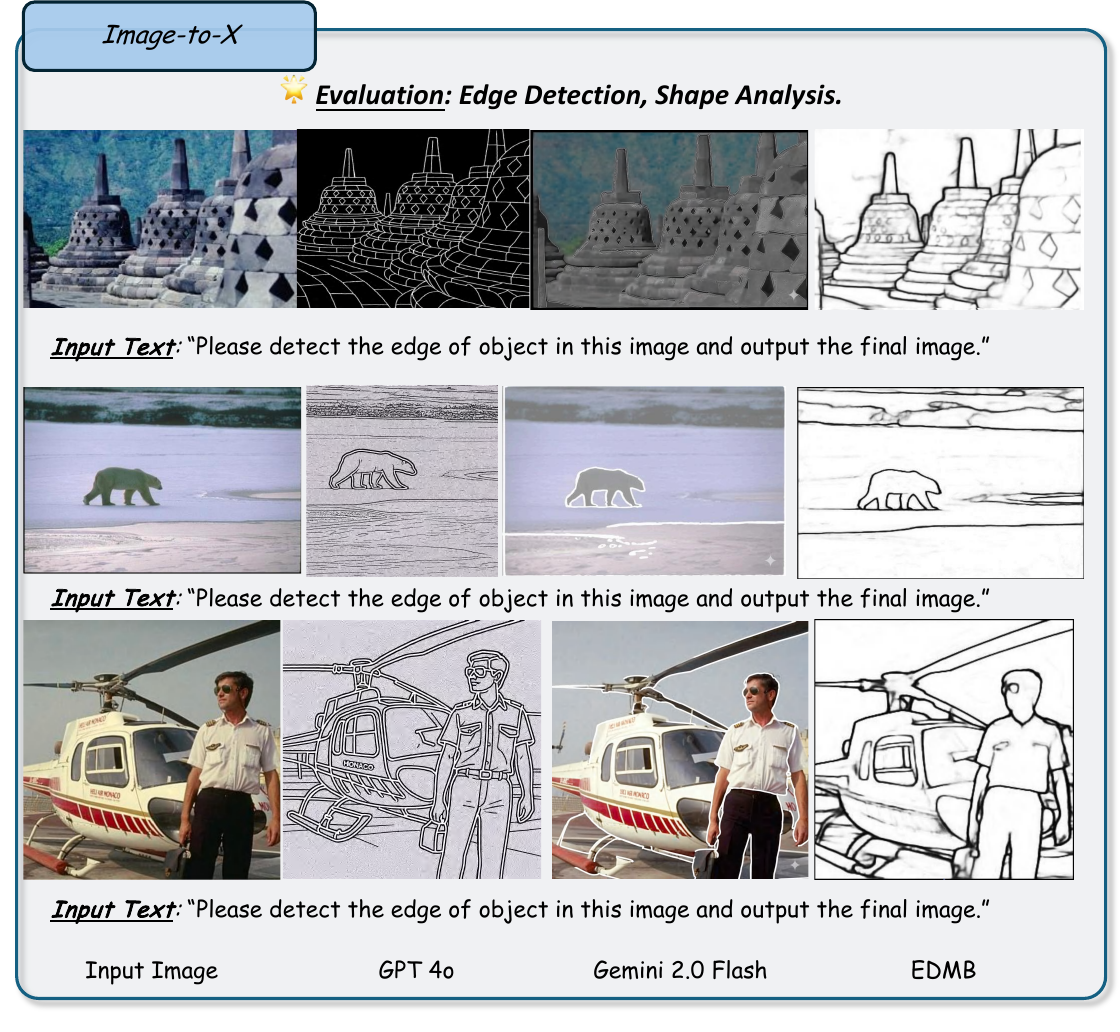}
    \caption{\textit{\textbf{Task:}} Image to X: Edge detection. Evaluate the shape analysis ability.
    \textit{\textbf{Setup:}} Each row shows an input image and a text prompt with outputs from GPT-4o, Gemini 2.0 Flash~\cite{gemini-2-0-flash}, and EDMB~\cite{li2025edmb}.
    \textit{\textbf{Observation:}} We find both GPT-4o and Gemini 2.0 Flash can detect object edges for both foreground and background objects.}
    \label{fig:edge_det}
\end{figure}

In Figure~\ref{fig:edge_det}, we compare this ability with a recent SOTA deep learning based approach, EMDB~\cite{li2025edmb}. 
For four examples, we find both GPT-4o and Gemini 2.0 Flash can detect object edges for both foreground and background objects.
In addition, the details are even good using GPT-4o.
We find two critical issues: 1) The spatial localization of GPT-4o is changed as observed by the segmentation tasks. 
2) The content of GPT-4o is also changed. For example, in the first example, the road is generated, which does not exist in the input image.

\noindent
\textbf{Image Matting.} Image matting is a technique in image processing that aims to separate a foreground object from its background and obtain a detailed alpha matte, which indicates the transparency or opacity of each pixel in the foreground. 
It goes beyond simple segmentation by providing more precise information about the boundaries and fine details of the object, especially for complex objects like hair or smoke.

\begin{figure}[h]
    \centering
    \includegraphics[width=1\textwidth]{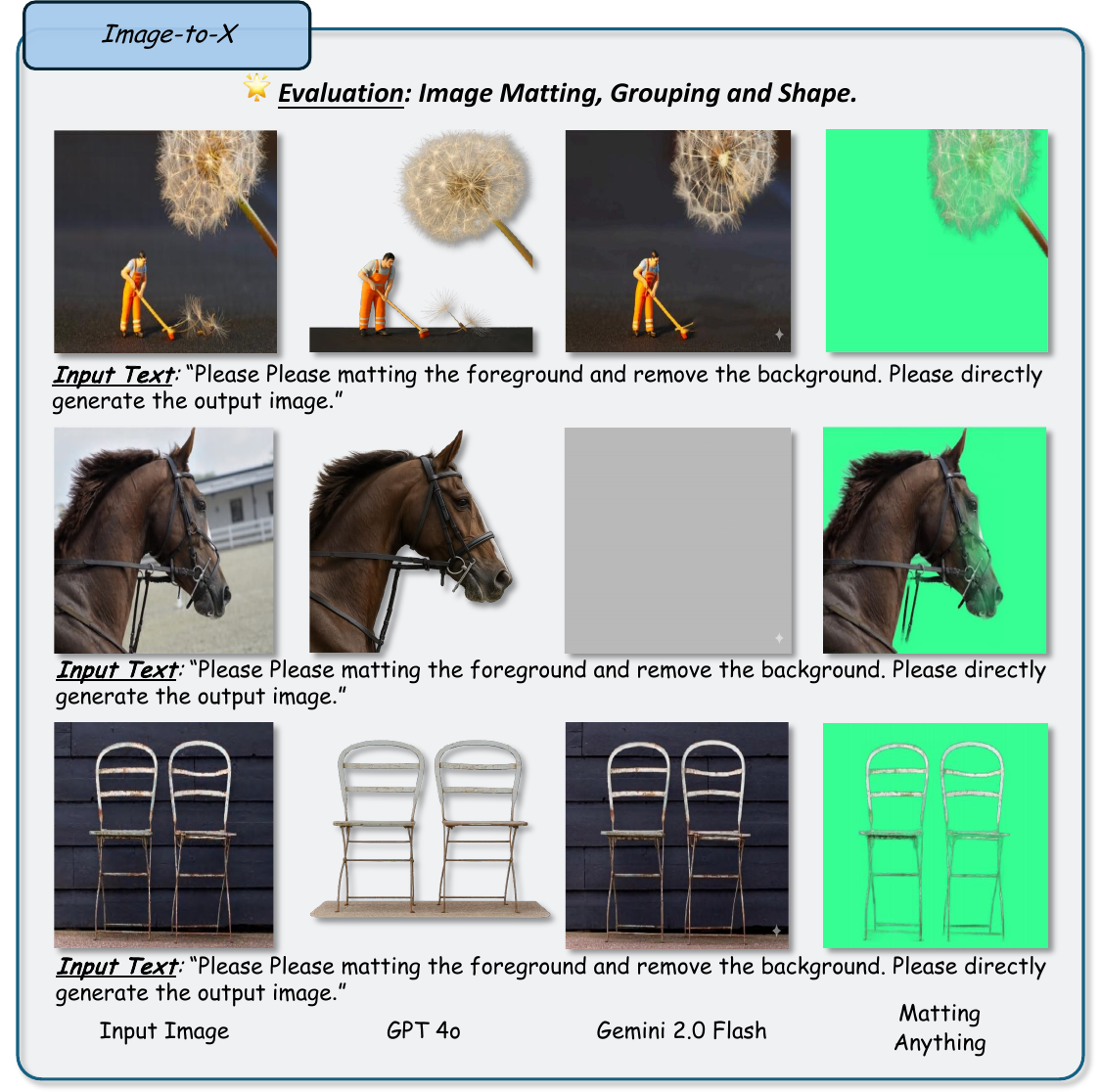}
    \caption{\textit{\textbf{Task:}} Image to X: Image matting. Evaluate the grouping and shape analysis ability.
    \textit{\textbf{Setup:}} Each row shows an input image and a text prompt with outputs from GPT-4o, Gemini 2.0 Flash~\cite{gemini-2-0-flash}, and Matting Anything~\cite{mattinganything}.
    \textit{\textbf{Observation:}} Compared with Gemini, GPT-4o can handle the simple cases, as shown in the third row. However, considering the strict requirements of image matting (fine-grained and aligned details), the overall quality is bad.}
    \label{fig:matting}
\end{figure}

In Figure~\ref{fig:matting}, we show three testing examples, with one expert model, Matting Anything~\cite{li2023matting}.
Compared with Gemini, GPT-4o can handle the simple cases, as shown in the third row.
Thus, it can understand the task goal. For example, it can even keep the fine-grained details of a horse hair.
However, considering the strict requirements of image matting (fine-grained and aligned details), the overall quality is bad.
Compared with Matting Anything, both GPT-4o and Gemini work poorly.
We find nearly the same issues: 1) Wrong spatial localization, 2) Changed contents.


\clearpage

\subsubsection{Salient Object}

\noindent
\textbf{Salient Object Detection.} Salient object detection is a crucial technique in the field of computer vision and image processing. It aims to identify and locate the most visually prominent objects within an image or a video sequence.

\begin{figure}[h]
    \centering
    \includegraphics[width=1\textwidth]{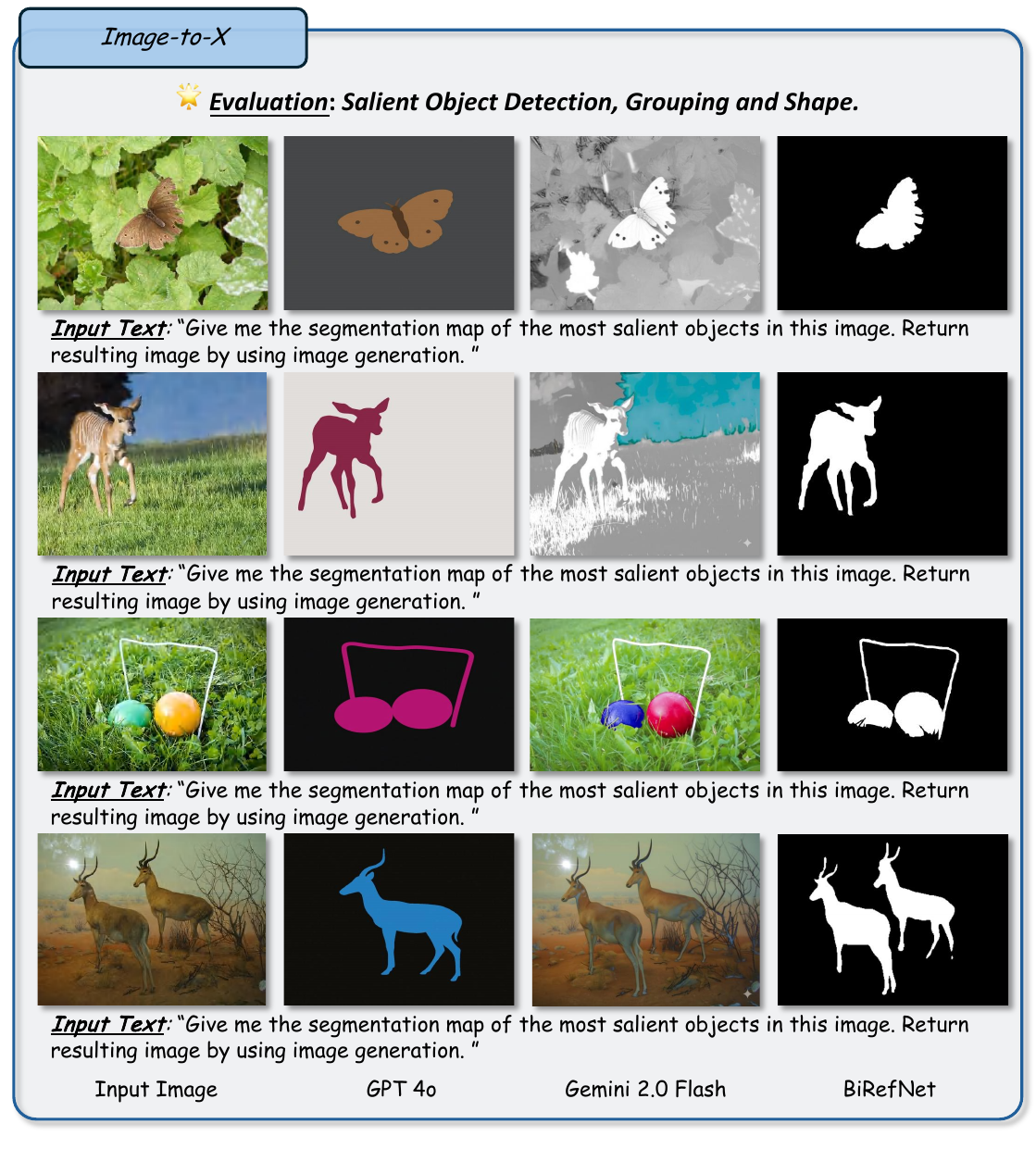}
    \caption{\textit{\textbf{Task:}} Image to X: Salient object detection. Evaluate the grouping and shape analysis ability.
    \textit{\textbf{Setup:}} Each row shows an input image and a text prompt with outputs from GPT-4o, Gemini 2.0 Flash~\cite{gemini-2-0-flash}, and BiRefNet~\cite{zheng2024birefnet}.
    \textit{\textbf{Observation:}} For all examples, compared with Gemini, GPT-4o can detect related salient objects with the text prompts while Gemini can not achieve this function.}
    \label{fig:sailent_object_det}
\end{figure}

In Figure~\ref{fig:sailent_object_det}, we adopt one expert model, BiRefNet~\cite{zheng2024birefnet}, as reference. 
For all examples, compared with Gemini 2.0 Flash, GPT-4o can detect relevant salient objects with the text prompts while Gemini can not achieve this.
The second example shows that the GPT-4o can generate the aligned salient masks.
However, for other examples, the spatial location is not changed where the results are generated according to the input image and potential classes.
In the last examples, GPT-4o cannot generate multiple salient object masks, which is also a limitation when dealing with multiple objects.

\noindent
\textbf{Mirror Detection.} Mirror detection is a task in computer vision that focuses on identifying mirror surfaces within an image or a scene. Previous works explore this direction by adopting visual cues and geometric cues.

\begin{figure}[h]
    \centering
    \includegraphics[width=1\textwidth]{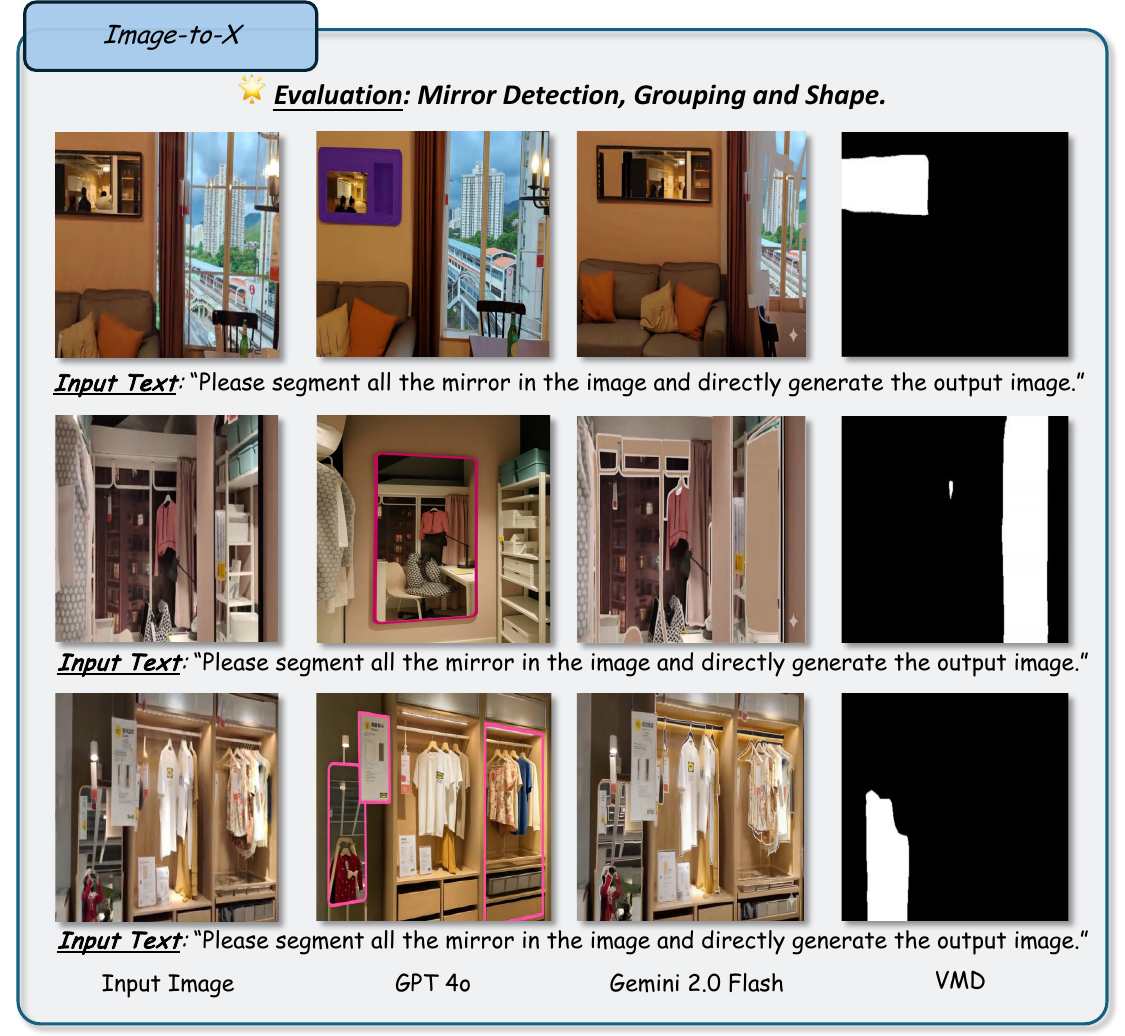}
    \caption{\textit{\textbf{Task:}} Image to X: Mirror detection. Evaluate the grouping and shape analysis ability.
    \textit{\textbf{Setup:}} Each row shows an input image and a text prompt with outputs from GPT-4o, Gemini 2.0 Flash~\cite{gemini-2-0-flash}, and VMD~\cite{warren2024effective}.
    \textit{\textbf{Observation:}} For simple cases, we find that GPT-4o can carry out mirror detection, as shown in the first example. For the complex scene, it cannot work as well as VMD.}
    \label{fig:mirror_det}
\end{figure}

In Figure~\ref{fig:mirror_det}, we also explore this ability for both GPT-4o and Gemini 2.0 Flash. 
As for comparison, we adopt a recent SOTA expert model, VMD~\cite{warren2024effective}.
For simple cases, we find that GPT-4o can carry out mirror detection, as shown in the first example.
%
For the complex scene, it cannot work as well as the expert model, VMD.
As shown in the second example, it generates a fake mirror and leads to a wrong image output with a line to mark the boundaries of the fake mirror.
As shown in the last row, GPT-4o treats several rectangular objects as mirrors, leading to several false positive examples.

\noindent
\textbf{Shadow Detection.} Shadow detection is a significant process in computer vision and image processing that aims to identify and localize shadow regions in an image or a video. This technique is crucial, as shadows can otherwise disrupt object detection, recognition, and scene analysis.

\begin{figure}[h]
    \centering
    \includegraphics[width=0.98\textwidth]{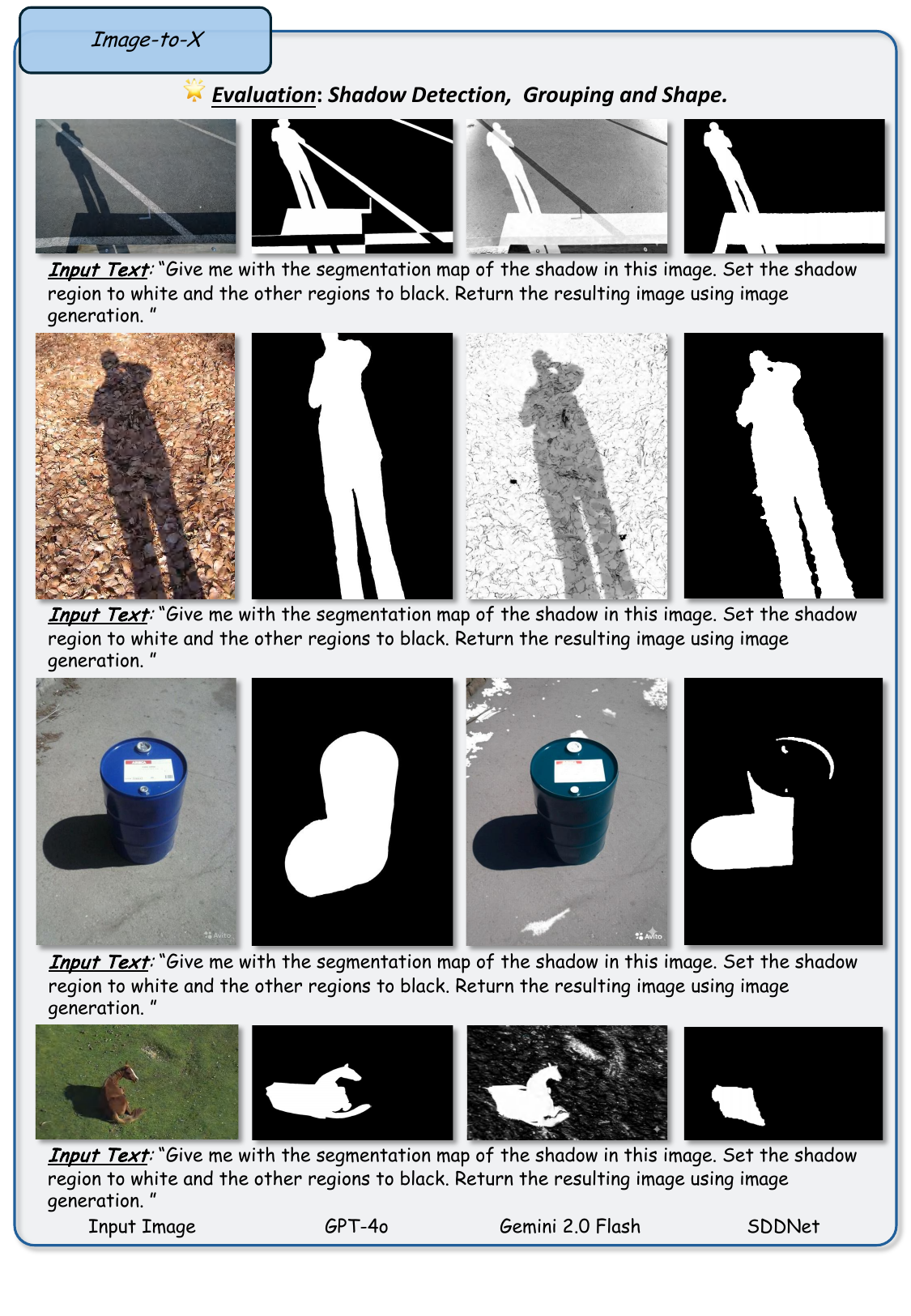}
    \caption{\textit{\textbf{Task:}} Image to X: Shadow detection. Evaluate the grouping and shape analysis ability.
    \textit{\textbf{Setup:}} Each row shows an input image and a text prompt with outputs from GPT-4o, Gemini 2.0 Flash~\cite{gemini-2-0-flash}, and SDDNet~\cite{cong2023sddnet}.
    \textit{\textbf{Observation:}} For more complex examples, both models detect both objects and their shadows with one mask output, as shown in the last two rows, leading to false positive predictions.}
    \label{fig:shadow_det}
\end{figure}

In Figure~\ref{fig:shadow_det}, we compare and test this ability for GPT-4o.
We adopt the SOTA model, SDDNet~\cite{cong2023sddnet} for reference.
For the simple examples (single objects and no objects in the image), both GPT-4o 
and Gemini can localize the shadow, as shown in the first two rows.
For more complex examples, both models detect both objects and their shadows with one mask output, as shown in the last two rows.
Thus, GPT-4o cannot handle these inputs.
In addition, the spatial misalignments also happen for all the cases.

\noindent
\textbf{Camouflage Object Detection.} Camouflage object detection is a challenging task in computer vision. It aims to identify objects that are designed to blend into their backgrounds, making them difficult to distinguish by human eyes or traditional detection methods. 
This has a wide application for the military, security, and wildlife conservation.

\begin{figure}[h]
    \centering
    \includegraphics[width=1\textwidth]{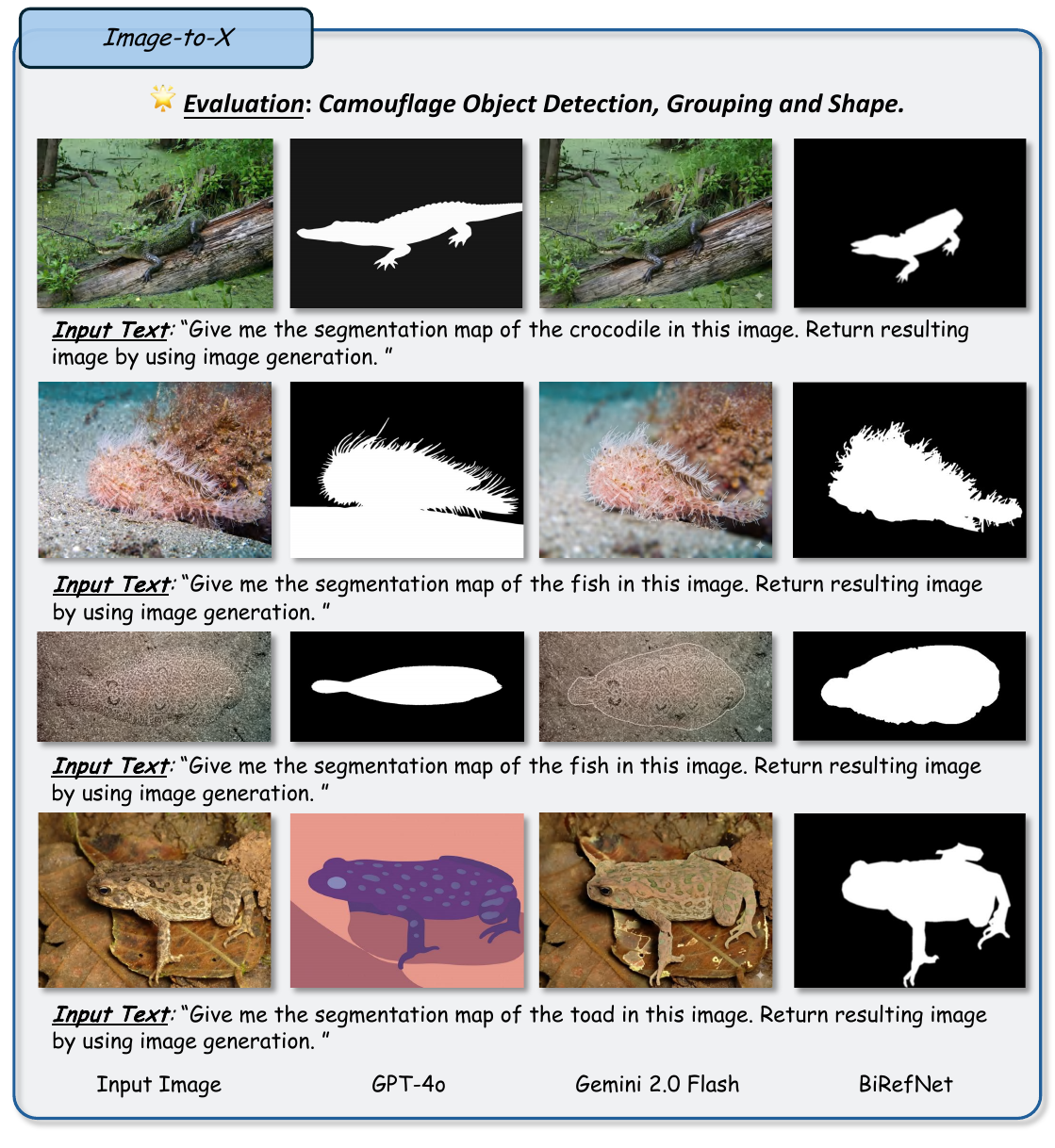}
    \caption{\textit{\textbf{Task:}} Image to X: Camouflage object detection. Evaluate the grouping and shape analysis ability.
    \textit{\textbf{Setup:}} Each row shows an input image and a text prompt with outputs from GPT-4o, Gemini 2.0 Flash~\cite{gemini-2-0-flash}, and BiRefNet~\cite{zheng2024birefnet}.
    \textit{\textbf{Observation:}} Both GPT-4o and Gemini 2.0 Flash can detect and segment the camouflage animals for simple cases. However, the spatial misalignments still exist.}
    \label{fig:camouflage_object}
\end{figure}

As shown in Figure~\ref{fig:camouflage_object}, we also include one expert model, BiRefNet~\cite{zheng2024birefnet} for reference.
For all examples, both GPT-4o and Gemini 2.0 Flash can detect and segment the camouflage animals for simple cases, as shown in the last two rows.
GPT-4o can also detect the specific object, given the text prompt, as shown in the first row.
However, the same misalignment issues still exist.
In addition, it also mixes segmentation maps (in binary masks or color masks), as shown in the last row.

\newpage

\clearpage

\subsubsection{Depth Estimation}
\begin{figure}[h]
    \centering
    \includegraphics[width=1\textwidth]{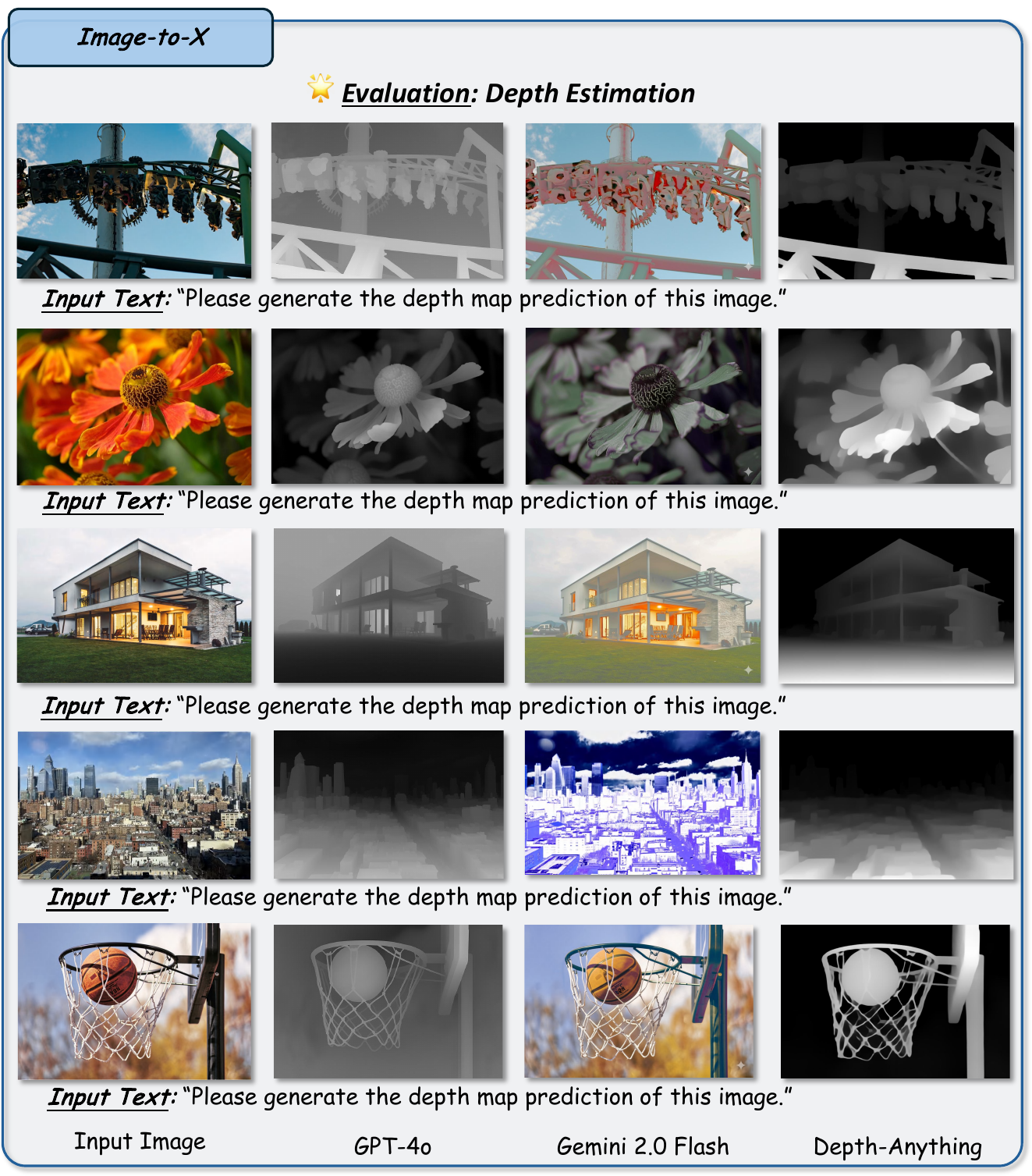}
    \caption{\textit{\textbf{Task:}} Image to X: Depth estimation.
    \textit{\textbf{Setup:}} Each row shows an input image and a text prompt with outputs from GPT-4o, Gemini 2.0 Flash~\cite{gemini-2-0-flash}, and Depth-Anything~\cite{depthanything}.
    \textit{\textbf{Observation:}} We convert the depth map generated by Depth-Anything into a visualization map similar to GPT-4o. This evaluation shows that GPT-4o has the capability of distinguishing the depth relationship of different parts in the image, but its understanding of the background is insufficient.}
    \label{fig:depth}
\end{figure}

The depth estimation task involves predicting the distance from the camera to objects within a scene. In this paper, we focus on monocular depth estimation, which takes a single image as input.
In Figure~\ref{fig:depth}, we compare GPT-4o, Gemini 2.0 Flash, and a recent SOTA method, Depth-Anything~\cite{depthanything}. We first notice that Gemini cannot produce reasonable depth estimations. For GPT-4o, although it can output a fancy depth map visualization, we want to point out that this output is a grayscale visualization of depth estimation and cannot be directly converted to the depth of each pixel.
We show mainly five cases.
In the first test case, we notice that GPT-4o is good at capturing details in images, which Depth-Anything may not be good at. 
Although we cannot directly determine the accuracy of the depth value, we can judge from the visualization that the depth relationship between objects is accurate.
What GPT-4o cannot do well is the background. Since the background in the image is the sky, we can infer from common sense that these areas are infinitely far away from the camera. However, the depth map output of GPT-4o does not handle these areas correctly.
GPT-4o performs similarly in the second, fourth, and fifth examples.
Among them, we would like to emphasize the fourth test case, since for buildings farther away, GPT-4o has no way to effectively analyze the distance between each building and the camera.
In the third example, although the output of GPT-4o is very confusing, it completely misunderstands the depth relationship of the entire image. Therefore, we believe that the depth estimation performance of GPT-4o is still unstable.

\clearpage

\subsubsection{Normal Estimation}
\begin{figure}[h]
    \centering
    \includegraphics[width=1\textwidth]{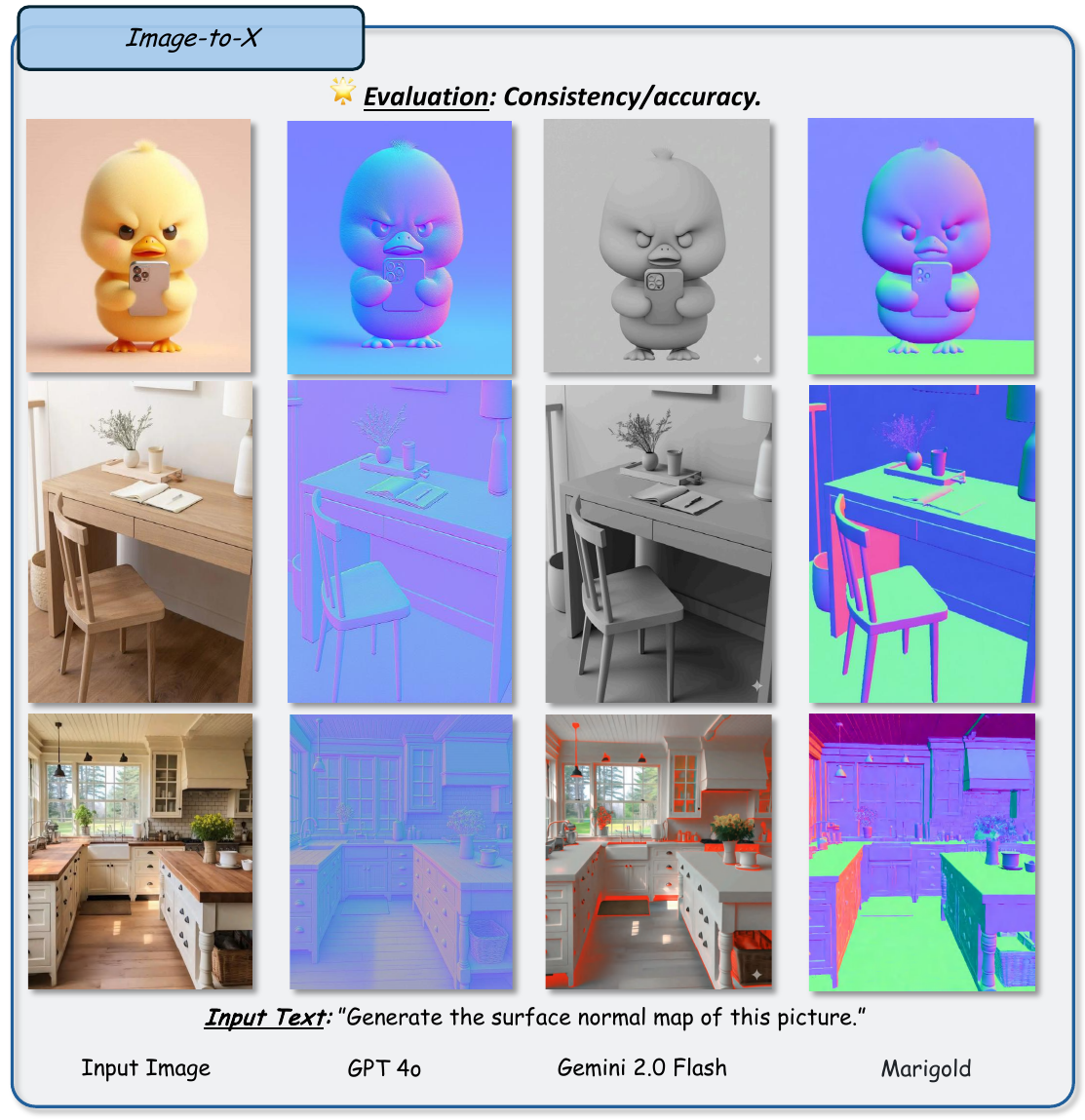}
    \caption{\textit{\textbf{Task:}} Image to X: Normal estimation.
    \textit{\textbf{Setup:}} Each row shows an input image and a text prompt with outputs from GPT-4o, Gemini 2.0 Flash~\cite{gemini-2-0-flash}, and Marigold~\cite{ke2023repurposing}.
    \textit{\textbf{Observation:}} This evaluation shows that GPT-4o has the capability of generating a visualization map of the surface normal, but the understanding of the details is still insufficient.}
    \label{fig:normal}
\end{figure}

The surface normal estimation task involves predicting the orientation of surfaces at each pixel in an image, typically represented as 3D vectors. 
In Figure~\ref{fig:normal}, we compare GPT-4o, Gemini 2.0 Flash, and Marigold normals~\cite{ke2023repurposing}. The results show that GPT-4o can generate reasonable results.
However, since GPT-4o's output is an appealing normal map visualization, we want to clarify that this output is a color-coded visualization and does not directly provide the exact normal vector for each pixel.
Thus, we cannot use lighting or other methods to verify the accuracy of the normal maps, and downstream tasks cannot use the output results.
However, we also find some unreasonable details. In the third test case, common sense suggests that the ground should be flat, but GPT-4o predicts normals for these textured areas that differ from the surrounding areas.

\clearpage

\subsubsection{Layout Detection}
\begin{figure}[h]
    \centering
    \includegraphics[width=1\textwidth]{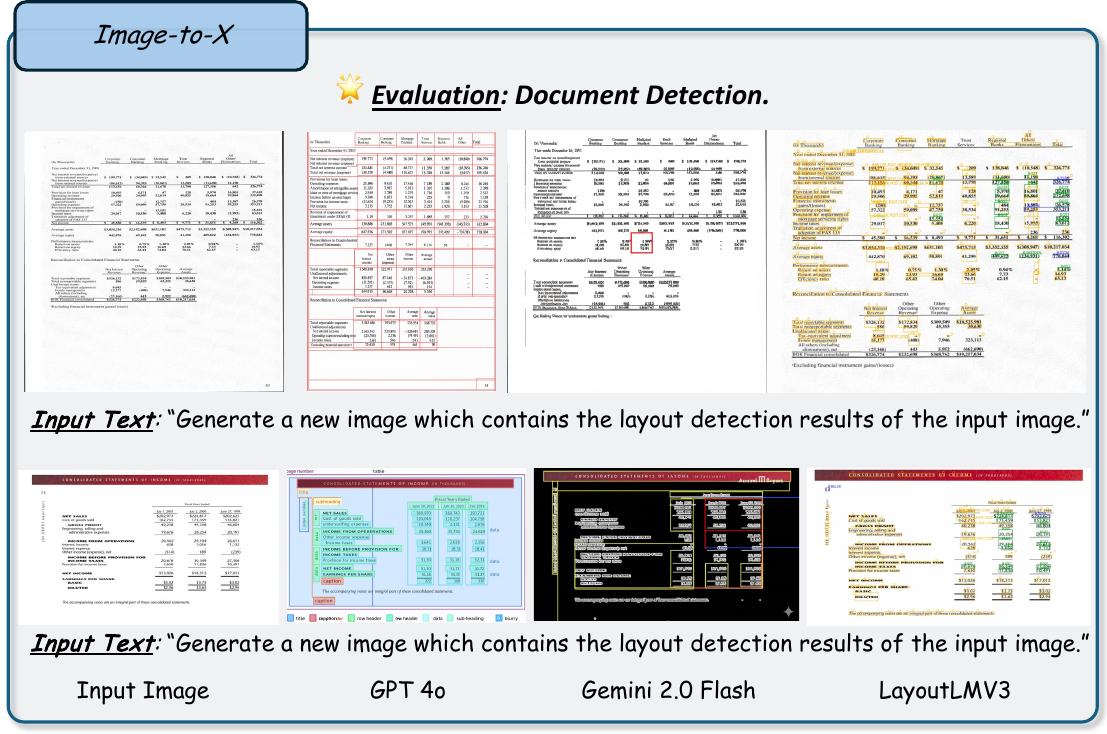}
    \caption{\textit{\textbf{Task:}} Image to X: Layout detection.
    \textit{\textbf{Setup:}} Each row shows an input image and a text prompt with outputs from GPT-4o, Gemini 2.0 Flash~\cite{gemini-2-0-flash}, and LayoutLMV3~\cite{huang2022layoutlmv3}.
    \textit{\textbf{Observation:}} The results show that GPT-4o and Gemini frequently generate a different document but a correct detected layout.}
    \label{fig:layout_det}
\end{figure}

The layout detection task requires the model to identify structural components (e.g., titles, paragraphs, tables, images) in the given image. In Figure~\ref{fig:layout_det}, we compare the performance of GPT-4o, Gemini 2.0 Flash, and LayoutLMV3~\cite{huang2022layoutlmv3} on the layout detection task.
In the test cases, GPT-4o hallucinates layout elements that do not exist, although the final output is another document with ``layout detection'' results.
If we consider the use in downstream tasks, such results are meaningless. Therefore, we conclude that GPT-4o is not capable of the layout detection task.

\clearpage

\subsubsection{Text Detection}
\begin{figure}[h]
    \centering
    \includegraphics[width=1\textwidth]{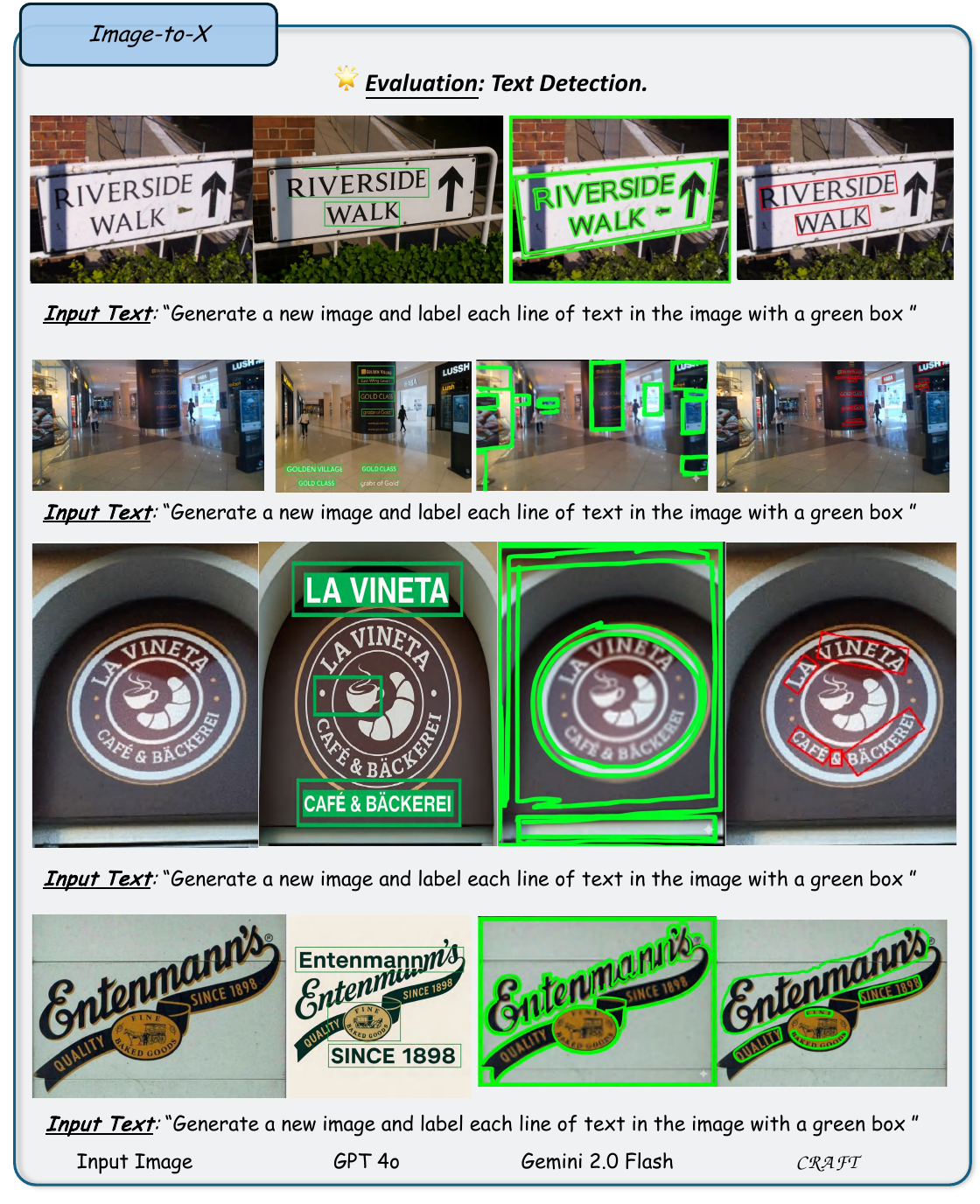}
    \caption{\textit{\textbf{Task:}} Image to X: Text detection.
    \textit{\textbf{Setup:}} Each row shows an input image and a text prompt with outputs from GPT-4o, Gemini 2.0 Flash~\cite{gemini-2-0-flash}, and CRAFT~\cite{baek2019character}.
    \textit{\textbf{Observation:}} The results show that GPT-4o frequently generates text that does not exist.}
    \label{fig:text_det}
\end{figure}

The text detection task requires the model to detect the texts in the given image. 
In Figure~\ref{fig:text_det}, we compare the performance of GPT-4o, Gemini 2.0 Flash~\cite{gemini-2-0-flash}, and CRAFT~\cite{baek2019character} regarding to text detection. 
We observe that CRAFT exhibits better performance compared to the other models.

In the first test case, GPT-4o demonstrates comparable performance to CRAFT. However, in other cases, GPT-4o continuously generates some nonexistent texts and labels them as ``text area''.
This issue becomes particularly evident in cluttered scenes or images with complex backgrounds. These false positives not only reduce detection precision but also make the output less reliable for downstream tasks such as OCR or document understanding.
On the other hand, Gemini does not generate nonexistent texts but tends to over-predict some areas as text areas.

\clearpage

\subsubsection{Object Tracking}
\begin{figure}[h]
    \centering
    \includegraphics[width=1\textwidth]{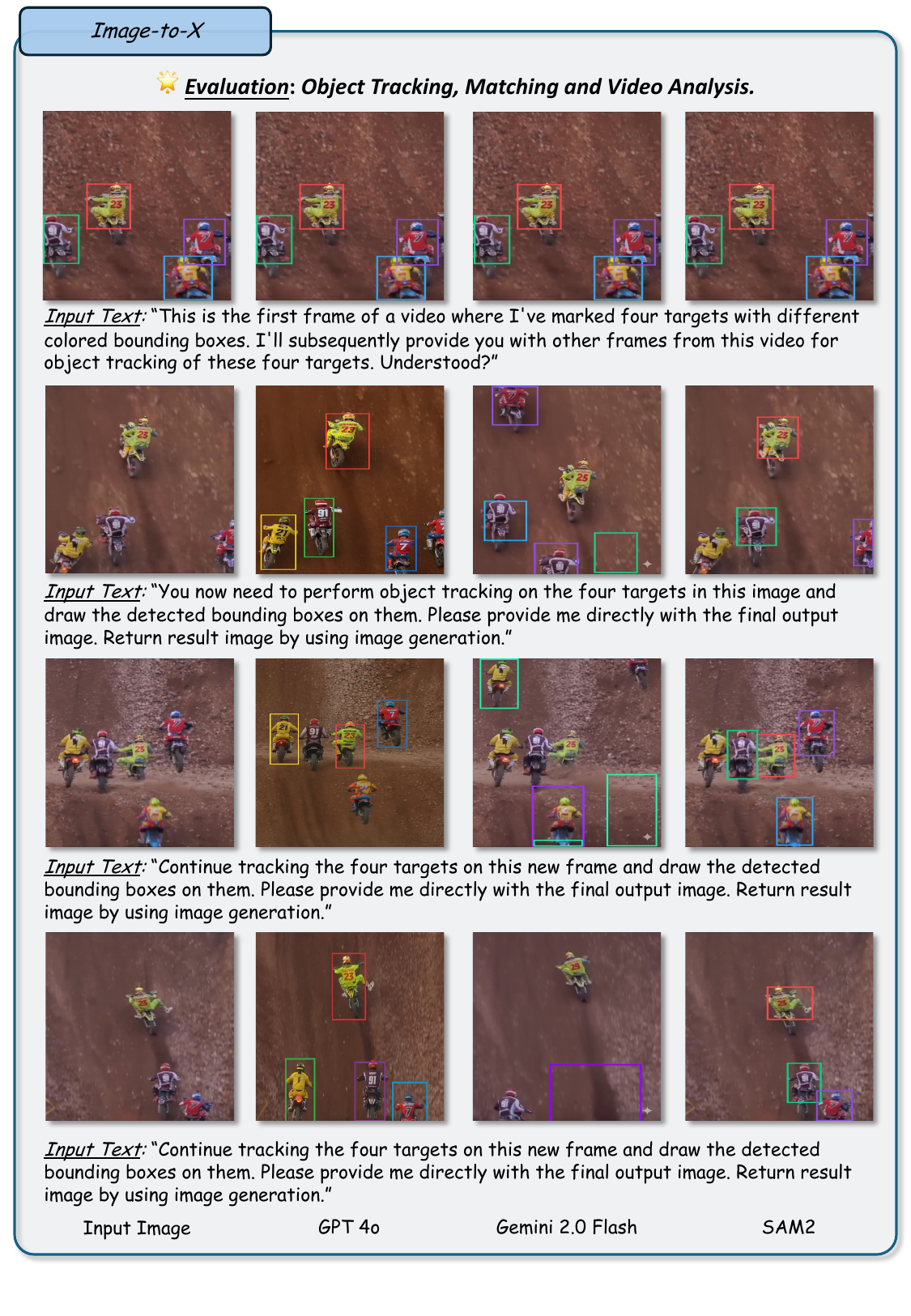}
    \caption{\textit{\textbf{Task:}} Image to X: Object tracking, matching, and video analysis (1/4).
    \textit{\textbf{Setup:}} Each row shows an input image and a text prompt with outputs from GPT-4o, Gemini 2.0 Flash~\cite{gemini-2-0-flash}, and SAM-2~\cite{ravi2024sam}.
    \textit{\textbf{Observation:}} This evaluation shows that GPT-4o has the capability of tracking objects, but it cannot generate a consistent image compared to the input image.}
    \label{fig:obj_track_1}
\end{figure}

\begin{figure}[h]
    \centering
    \includegraphics[width=1\textwidth]{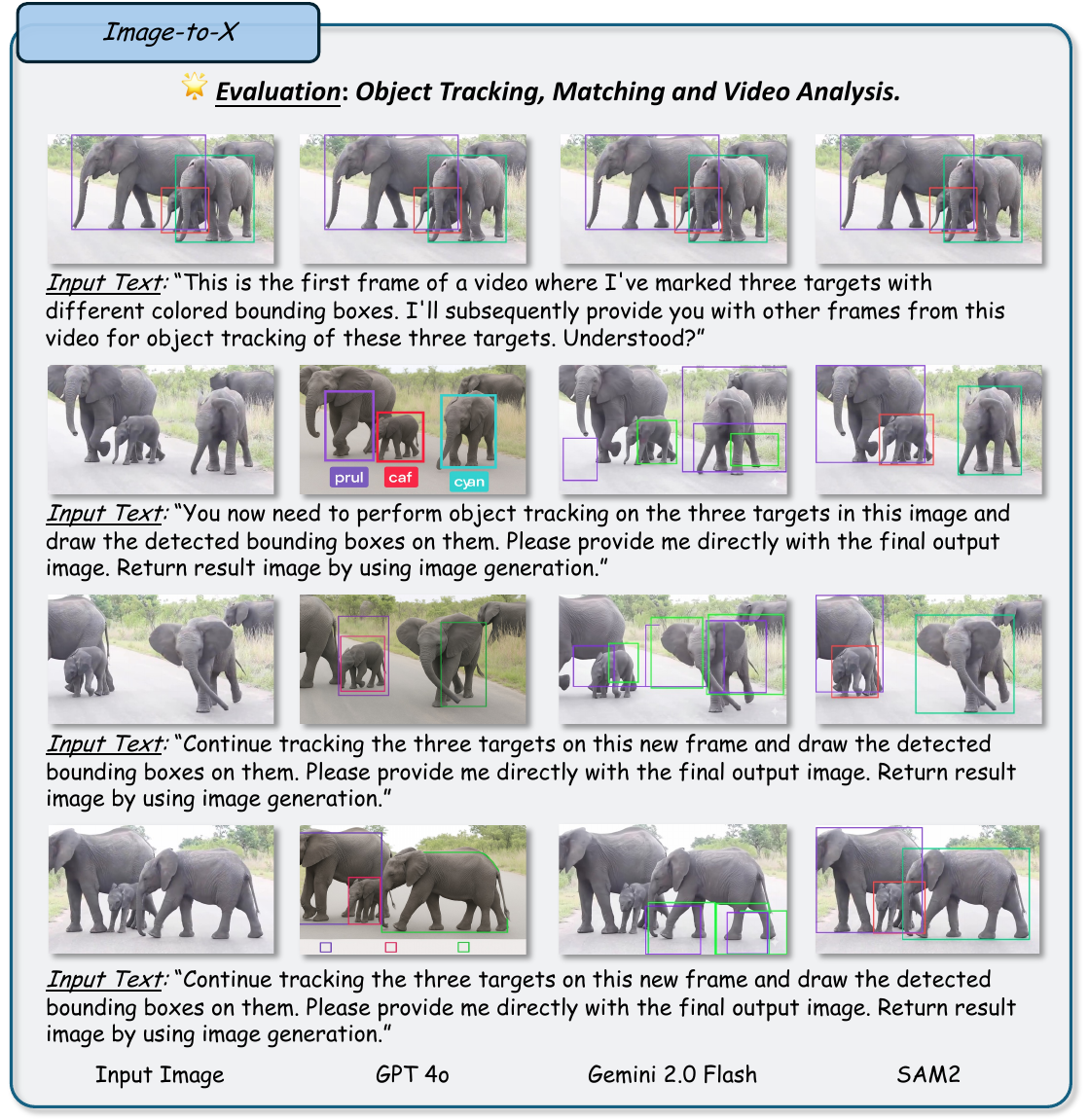}
    \caption{\textit{\textbf{Task:}} Image to X: Object tracking, matching, and video analysis (2/4).
    \textit{\textbf{Setup:}} Each row shows an input image and a text prompt with outputs from GPT-4o, Gemini 2.0 Flash~\cite{gemini-2-0-flash}, and SAM-2~\cite{ravi2024sam}.
    \textit{\textbf{Observation:}} This evaluation shows that GPT-4o has the capability of tracking objects, but it cannot generate a consistent image compared to the input image.}
    \label{fig:obj_track_2}
\end{figure}

\begin{figure}[h]
    \centering
    \includegraphics[width=1\textwidth]{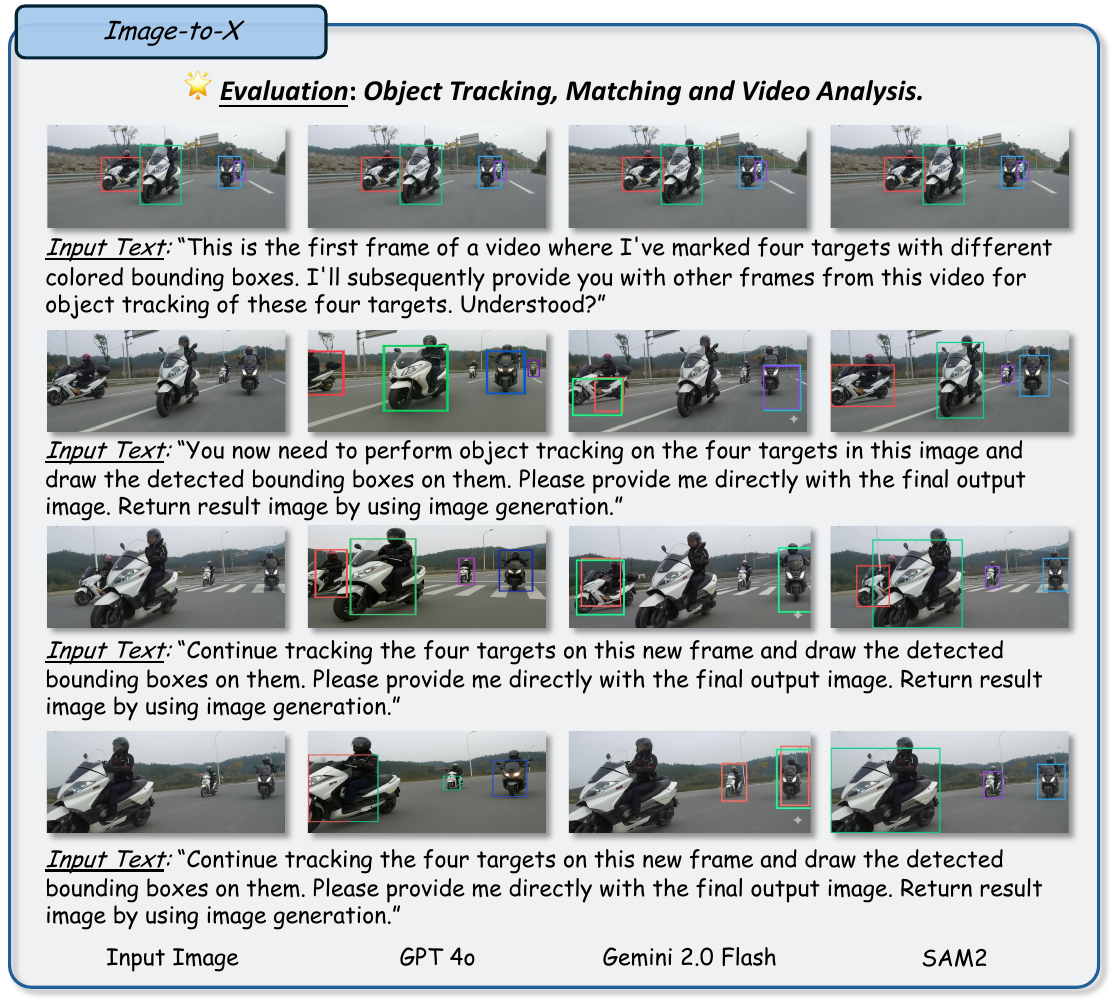}
    \caption{\textit{\textbf{Task:}} Image to X: Object tracking, matching, and video analysis (3/4).
    \textit{\textbf{Setup:}} Each row shows an input image and a text prompt with outputs from GPT-4o, Gemini 2.0 Flash~\cite{gemini-2-0-flash}, and SAM-2~\cite{ravi2024sam}.
    \textit{\textbf{Observation:}} This evaluation shows that GPT-4o has the capability of tracking objects, but it cannot generate a consistent image compared to the input image.}
    \label{fig:obj_track_3}
\end{figure}

\begin{figure}[h]
    \centering
    \includegraphics[width=1\textwidth]{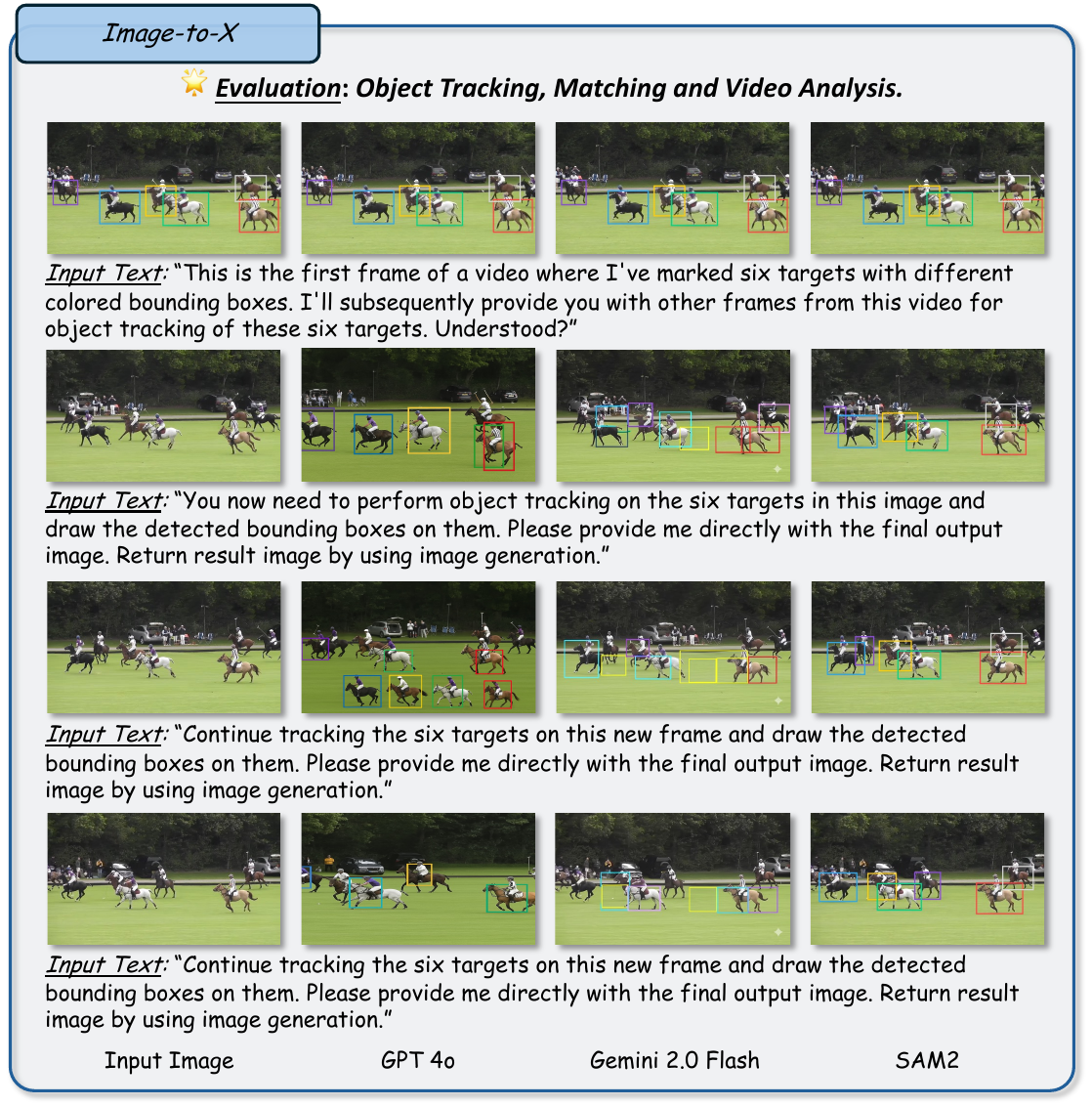}
    \caption{\textit{\textbf{Task:}} Image to X: Object tracking, matching, and video analysis (4/4).
    \textit{\textbf{Setup:}} Each row shows an input image and a text prompt with outputs from GPT-4o, Gemini 2.0 Flash~\cite{gemini-2-0-flash}, and SAM-2~\cite{ravi2024sam}.
    \textit{\textbf{Observation:}} This evaluation shows that GPT-4o has the capability of tracking objects, but it cannot generate a consistent image compared to the input image.}
    \label{fig:obj_track_4}
\end{figure}

The object tracking task requires the model to continuously locate and follow the specific object across the frames in a video sequence. We test the multi-object tracking, which requires the model to track several objects concurrently. 
We test four cases (Figure~\ref{fig:obj_track_1},~\ref{fig:obj_track_2},~\ref{fig:obj_track_3},~\ref{fig:obj_track_4}). We compare GPT-4o, Gemini 2.0 Flash, and a recent SOTA method SAM-2~\cite{ravi2024sam}. 
Our first observation is that GPT-4o seems unable to generate images that are consistent with the original image. This may be related to the nature of its generative model.
Even if we ignore this, for the tracking task, SAM-2 still performs better, while GPT-4o will have problems such as failing to maintain consistent tracking of the target, frequently drifting, or losing the object entirely.
In Figure~\ref{fig:obj_track_1}, the output of GPT-4o generally demonstrates the ability to track objects, but there are also some defects. For example, a new object is even created out of the existing objects in the last picture generated by GPT-4o.
We speculate that this is caused by the influence of the conversation context.
In Figure~\ref{fig:obj_track_2}, GPT-4o outputs some content that should not be in the output, such as the ``caf'' tag.
In Figure~\ref{fig:obj_track_3}, GPT-4o can track a relatively simple object, but it fuses two separate objects.
In Figure~\ref{fig:obj_track_4}, GPT-4o lacks the capability of tracking in the dense scenario.

\newpage

\clearpage

\section{Limitations}

Although GPT-4o demonstrates impressive capabilities across a wide range of image generation tasks, several limitations remain. These challenges highlight key areas for future improvement in developing unified foundation models for vision-language generation.

\subsection{Inconsistent Generation}

While GPT-4o often produces high-quality and semantically relevant images conditioned on textual prompts, it occasionally exhibits inconsistencies. Specifically, the model may generate visually compelling outputs that deviate from precise semantic cues of the input image, such as object count, spatial layout, specific shapes, or designated colors. These inconsistencies are especially problematic in tasks requiring partial image editing or compositional accuracy. Notably, such issues are less common in diffusion-based models or discrete denoising architectures like MaskGIT~\cite{chang2022maskgit,bai2024meissonic}, suggesting that GPT-4o operates under a distinct generative paradigm with inherent trade-offs in fidelity and control.

\subsection{Hallucination}

GPT-4o is also susceptible to hallucinations—producing content that is logically implausible, semantically inconsistent, or factually incorrect. These include fabricating non-existent objects or geographical features (e.g., imaginary islands or landmarks), and misrepresenting relationships between entities. Such errors are particularly prevalent in complex or underspecified prompts, where the model appears to rely on internal priors rather than grounded world knowledge. While hallucination is a common challenge across generative models, it poses notable limitations for real-world applications demanding precision, such as education, medical illustration, or scientific visualization.

\subsection{Data Bias}

Despite strong alignment between text and vision modalities, GPT-4o struggles with data bias issue, which fail in generating underrepresented cultural elements and rendering non-Latin scripts such as Chinese, Japanese, and Arabic. The generated characters are often incomplete, distorted, or replaced with Latin-like approximations. These artifacts reflect underlying challenges in multilingual representation, likely due to limited exposure to diverse scripts during training and the inherent difficulty of accurate typographic rendering in pixel space. 
This phenomenon is emblematic of a larger issue in AI systems—data bias. The training data used to develop models like GPT-4o may disproportionately represent certain languages, cultures, and writing systems, leading to disparities in performance across different linguistic groups. These biases are not only technical limitations but also ethical concerns, as they can contribute to the exclusion of underrepresented languages and cultures from AI applications. 
As vision-language models are increasingly deployed globally, improving support for multilingual text remains a crucial step toward inclusive and culturally competent AI systems.

\section{Conclusion}
\label{sec:conclusion}

In conclusion, this work presents a comprehensive study on the development of unified vision-language generative models, with a focus on evaluating GPT-4o across a wide range of image generation tasks.
Our analysis shows that GPT-4o demonstrates strong capabilities in aligning vision and language, achieving competitive results across text-to-image, image-to-image, image-to-3D, and image-to-X tasks.
However, limitations remain in inconsistent generation, hallucination, and data bias in underrepresented cultural elements and non-Latin scripts, highlighting current trade-offs in model design and training data coverage.
We also emphasize that architecture alone does not determine success; training data, model scale, and optimization strategies are equally critical components of progress.
We hope future work will provide deeper empirical insights into such proprietary systems and clarify their position within the broader landscape of unified generative modeling.

{\small
\bibliographystyle{plain}
  \bibliography{refbib}

\begin{thebibliography}{100}

\bibitem{dream360}
Hao Ai, Zidong Cao, Haonan Lu, Chen Chen, Jian Ma, Pengyuan Zhou, Tae-Kyun Kim, Pan Hui, and Lin Wang.
\newblock Dream360: Diverse and immersive outdoor virtual scene creation via transformer-based 360 image outpainting.
\newblock {\em IEEE transactions on visualization and computer graphics}, 2024.

\bibitem{ideogram-3-0}
Ideogram AI.
\newblock Ideogram.
\newblock \url{https://ideogram.ai/}, 2024.

\bibitem{baek2019character}
Youngmin Baek, Bado Lee, Dongyoon Han, Sangdoo Yun, and Hwalsuk Lee.
\newblock Character region awareness for text detection.
\newblock In {\em CVPR}, 2019.

\bibitem{bai2024humanedit}
Jinbin Bai, Wei Chow, Ling Yang, Xiangtai Li, Juncheng Li, Hanwang Zhang, and Shuicheng Yan.
\newblock Humanedit: A high-quality human-rewarded dataset for instruction-based image editing.
\newblock {\em arXiv preprint arXiv:2412.04280}, 2024.

\bibitem{bai2023integrating}
Jinbin Bai, Zhen Dong, Aosong Feng, Xiao Zhang, Tian Ye, Kaicheng Zhou, and Mike~Zheng Shou.
\newblock Integrating view conditions for image synthesis.
\newblock {\em arXiv preprint arXiv:2310.16002}, 2023.

\bibitem{bai2024meissonic}
Jinbin Bai, Tian Ye, Wei Chow, Enxin Song, Qing-Guo Chen, Xiangtai Li, Zhen Dong, Lei Zhu, and Shuicheng Yan.
\newblock Meissonic: Revitalizing masked generative transformers for efficient high-resolution text-to-image synthesis.
\newblock {\em arXiv preprint arXiv:2410.08261}, 2024.

\bibitem{barratt2018note}
Shane Barratt and Rishi Sharma.
\newblock A note on the inception score.
\newblock {\em arXiv preprint arXiv:1801.01973}, 2018.

\bibitem{betker2023improving}
James Betker, Gabriel Goh, Li~Jing, Tim Brooks, Jianfeng Wang, Linjie Li, Long Ouyang, Juntang Zhuang, Joyce Lee, Yufei Guo, et~al.
\newblock Improving image generation with better captions.
\newblock {\em Computer Science. https://cdn. openai. com/papers/dall-e-3. pdf}, 2023.

\bibitem{brack2023ledits}
Manuel Brack, Felix Friedrich, Katharina Kornmeier, Linoy Tsaban, Patrick Schramowski, Kristian Kersting, and Apolinário Passos.
\newblock Ledits++: Limitless image editing using text-to-image models.
\newblock 2023.

\bibitem{brooks2022instructpix2pix}
Tim Brooks, Aleksander Holynski, and Alexei~A Efros.
\newblock Instructpix2pix: Learning to follow image editing instructions.
\newblock {\em arXiv preprint arXiv:2211.09800}, 2022.

\bibitem{chang2022maskgit}
Huiwen Chang, Han Zhang, Lu~Jiang, Ce~Liu, and William~T Freeman.
\newblock Maskgit: Masked generative image transformer.
\newblock In {\em Proceedings of the IEEE/CVF conference on computer vision and pattern recognition}, pages 11315--11325, 2022.

\bibitem{posta}
Haoyu Chen, Xiaojie Xu, Wenbo Li, Jingjing Ren, Tian Ye, Songhua Liu, Ying-Cong Chen, Lei Zhu, and Xinchao Wang.
\newblock Posta: A go-to framework for customized artistic poster generation.
\newblock {\em arXiv preprint arXiv:2503.14908}, 2025.

\bibitem{chen2025multimodal}
Liang Chen, Shuai Bai, Wenhao Chai, Weichu Xie, Haozhe Zhao, Leon Vinci, Junyang Lin, and Baobao Chang.
\newblock Multimodal representation alignment for image generation: Text-image interleaved control is easier than you think.
\newblock {\em arXiv preprint arXiv:2502.20172}, 2025.

\bibitem{deeplab-v3}
Liang-Chieh Chen, George Papandreou, Florian Schroff, and Hartwig Adam.
\newblock Rethinking atrous convolution for semantic image segmentation.
\newblock {\em arXiv preprint arXiv:1706.05587}, 2017.

\bibitem{udrs2former}
Sixiang Chen, Tian Ye, Jinbin Bai, Erkang Chen, Jun Shi, and Lei Zhu.
\newblock Sparse sampling transformer with uncertainty-driven ranking for unified removal of raindrops and rain streaks.
\newblock In {\em Proceedings of the IEEE/CVF International Conference on Computer Vision}, pages 13106--13117, 2023.

\bibitem{chen2022snowformer}
Sixiang Chen, Tian Ye, Yun Liu, and Erkang Chen.
\newblock Snowformer: Context interaction transformer with scale-awareness for single image desnowing.
\newblock {\em arXiv preprint arXiv:2208.09703}, 2022.

\bibitem{chen2024teaching}
Sixiang Chen, Tian Ye, Kai Zhang, Zhaohu Xing, Yunlong Lin, and Lei Zhu.
\newblock Teaching tailored to talent: Adverse weather restoration via prompt pool and depth-anything constraint.
\newblock In {\em European Conference on Computer Vision}, pages 95--115. Springer, 2024.

\bibitem{chen2023improving}
Tianqi Chen, Yongfei Liu, Zhendong Wang, Jianbo Yuan, Quanzeng You, Hongxia Yang, and Mingyuan Zhou.
\newblock Improving in-context learning in diffusion models with visual context-modulated prompts.
\newblock {\em arXiv preprint arXiv:2312.01408}, 2023.

\bibitem{janus-pro}
Xiaokang Chen, Zhiyu Wu, Xingchao Liu, Zizheng Pan, Wen Liu, Zhenda Xie, Xingkai Yu, and Chong Ruan.
\newblock Janus-pro: Unified multimodal understanding and generation with data and model scaling.
\newblock {\em arXiv preprint arXiv:2501.17811}, 2025.

\bibitem{instructir}
Marcos~V. Conde, Gregor Geigle, and Radu Timofte.
\newblock Instructir: High-quality image restoration following human instructions.
\newblock In {\em ECCV}, 2024.

\bibitem{cong2023sddnet}
Runmin Cong, Yuchen Guan, Jinpeng Chen, Wei Zhang, Yao Zhao, and Sam Kwong.
\newblock Sddnet: Style-guided dual-layer disentanglement network for shadow detection.
\newblock In {\em ACM MM}, 2023.

\bibitem{latentinpainting}
Ciprian Corneanu, Raghudeep Gadde, and Aleix~M Martinez.
\newblock Latentpaint: Image inpainting in latent space with diffusion models.
\newblock In {\em WACV}, 2024.

\bibitem{deng2022stytr2}
Yingying Deng, Fan Tang, Weiming Dong, Chongyang Ma, Xingjia Pan, Lei Wang, and Changsheng Xu.
\newblock Stytr2: Image style transfer with transformers.
\newblock In {\em CVPR}, 2022.

\bibitem{dong2023dreamllm}
Runpei Dong, Chunrui Han, Yuang Peng, Zekun Qi, Zheng Ge, Jinrong Yang, Liang Zhao, Jianjian Sun, Hongyu Zhou, Haoran Wei, et~al.
\newblock Dreamllm: Synergistic multimodal comprehension and creation.
\newblock {\em arXiv preprint arXiv:2309.11499}, 2023.

\bibitem{shadowrefiner}
Wei Dong, Han Zhou, Yuqiong Tian, Jingke Sun, Xiaohong Liu, Guangtao Zhai, and Jun Chen.
\newblock Shadowrefiner: Towards mask-free shadow removal via fast fourier transformer.
\newblock {\em arXiv preprint arXiv:2406.02559}.

\bibitem{esser2024scaling}
Patrick Esser, Sumith Kulal, Andreas Blattmann, Rahim Entezari, Jonas M{\"u}ller, Harry Saini, Yam Levi, Dominik Lorenz, Axel Sauer, Frederic Boesel, et~al.
\newblock Scaling rectified flow transformers for high-resolution image synthesis.
\newblock In {\em Forty-first International Conference on Machine Learning}, 2024.

\bibitem{sd3}
Patrick Esser, Sumith Kulal, Andreas Blattmann, Rahim Entezari, Jonas M{\"u}ller, Harry Saini, Yam Levi, Dominik Lorenz, Axel Sauer, Frederic Boesel, et~al.
\newblock Scaling rectified flow transformers for high-resolution image synthesis.
\newblock In {\em Forty-first international conference on machine learning}, 2024.

\bibitem{esser2021taming}
Patrick Esser, Robin Rombach, and Bjorn Ommer.
\newblock Taming transformers for high-resolution image synthesis.
\newblock In {\em Proceedings of the IEEE/CVF conference on computer vision and pattern recognition (CVPR)}, pages 12873--12883, 2021.

\bibitem{feng2024item}
Aosong Feng, Weikang Qiu, Jinbin Bai, Kaicheng Zhou, Zhen Dong, Xiao Zhang, Rex Ying, and Leandros Tassiulas.
\newblock An item is worth a prompt: Versatile image editing with disentangled control.
\newblock {\em arXiv preprint arXiv:2403.04880}, 2024.

\bibitem{fu2024mgie}
Tsu-Jui Fu, Wenze Hu, Xianzhi Du, William~Yang Wang, Yinfei Yang, and Zhe Gan.
\newblock Guiding instruction-based image editing via multimodal large language models.
\newblock {\em ICLR}, 2024.

\bibitem{gal2022image}
Rinon Gal, Yuval Alaluf, Yuval Atzmon, Or~Patashnik, Amit~H Bermano, Gal Chechik, and Daniel Cohen-Or.
\newblock An image is worth one word: Personalizing text-to-image generation using textual inversion.
\newblock {\em ICLR}, 2023.

\bibitem{gao2022get3d}
Jun Gao, Tianchang Shen, Zian Wang, Wenzheng Chen, Kangxue Yin, Daiqing Li, Or~Litany, Zan Gojcic, and Sanja Fidler.
\newblock Get3d: A generative model of high quality 3d textured shapes learned from images.
\newblock {\em NeurIPS}, 2022.

\bibitem{gatys2016image}
Leon~A Gatys, Alexander~S Ecker, and Matthias Bethge.
\newblock Image style transfer using convolutional neural networks.
\newblock {\em CVPR}, 2016.

\bibitem{seedx}
Yuying Ge, Sijie Zhao, Jinguo Zhu, Yixiao Ge, Kun Yi, Lin Song, Chen Li, Xiaohan Ding, and Ying Shan.
\newblock Seed-x: Multimodal models with unified multi-granularity comprehension and generation.
\newblock {\em arXiv preprint arXiv:2404.14396}, 2024.

\bibitem{goodfellow2020generative}
Ian Goodfellow, Jean Pouget-Abadie, Mehdi Mirza, Bing Xu, David Warde-Farley, Sherjil Ozair, Aaron Courville, and Yoshua Bengio.
\newblock Generative adversarial networks.
\newblock {\em Communications of the ACM}, 63(11):139--144, 2020.

\bibitem{gu2024mix}
Yuchao Gu, Xintao Wang, Jay~Zhangjie Wu, Yujun Shi, Yunpeng Chen, Zihan Fan, Wuyou Xiao, Rui Zhao, Shuning Chang, Weijia Wu, et~al.
\newblock Mix-of-show: Decentralized low-rank adaptation for multi-concept customization of diffusion models.
\newblock In {\em NeurIPS}, 2024.

\bibitem{heusel2017gans}
Martin Heusel, Hubert Ramsauer, Thomas Unterthiner, Bernhard Nessler, and Sepp Hochreiter.
\newblock Gans trained by a two time-scale update rule converge to a local nash equilibrium.
\newblock {\em NeurIPS}, 2017.

\bibitem{storydiffusion}
Qibin Hou, Yuying Ge, Jing Zhang, Yuchao Dai, and Ming-Ming Cheng.
\newblock Storydiffusion: Consistent self-attention for long-range image and video generation.
\newblock In {\em Advances in Neural Information Processing Systems (NeurIPS)}, 2024.

\bibitem{dsit}
Qiming Hu, Hainuo Wang, and Xiaojie Guo.
\newblock Single image reflection separation via dual-stream interactive transformers.
\newblock {\em Advances in Neural Information Processing Systems}, 37:55228--55248, 2024.

\bibitem{huang2024dual}
Jiancheng Huang, Yi~Huang, Jianzhuang Liu, Donghao Zhou, Yifan Liu, and Shifeng Chen.
\newblock Dual-schedule inversion: Training-and tuning-free inversion for real image editing.
\newblock {\em arXiv preprint arXiv:2412.11152}, 2024.

\bibitem{huang2025t2i}
Kaiyi Huang, Chengqi Duan, Kaiyue Sun, Enze Xie, Zhenguo Li, and Xihui Liu.
\newblock T2i-compbench++: An enhanced and comprehensive benchmark for compositional text-to-image generation.
\newblock {\em IEEE Transactions on Pattern Analysis and Machine Intelligence}, 2025.

\bibitem{huang2024context}
Lianghua Huang, Wei Wang, Zhi-Fan Wu, Yupeng Shi, Huanzhang Dou, Chen Liang, Yutong Feng, Yu~Liu, and Jingren Zhou.
\newblock In-context lora for diffusion transformers.
\newblock {\em arXiv preprint arXiv:2410.23775}, 2024.

\bibitem{huang2017arbitrary}
Xun Huang and Serge Belongie.
\newblock Arbitrary style transfer in real-time with adaptive instance normalization.
\newblock In {\em ICCV}, 2017.

\bibitem{huang2022layoutlmv3}
Yupan Huang, Tengchao Lv, Lei Cui, Yutong Lu, and Furu Wei.
\newblock Layoutlmv3: Pre-training for document ai with unified text and image masking.
\newblock In {\em ACM MM}, 2022.

\bibitem{huang2022planes}
Zixuan Huang, Stefan Stojanov, Anh Thai, Varun Jampani, and James~M Rehg.
\newblock Planes vs. chairs: Category-guided 3d shape learning without any 3d cues.
\newblock In {\em ECCV}, 2022.

\bibitem{jiang2024mc}
Jiaxiu Jiang, Yabo Zhang, Kailai Feng, Xiaohe Wu, Wenbo Li, Renjing Pei, Fan Li, and Wangmeng Zuo.
\newblock Mc2: Multi-concept guidance for customized multi-concept generation.
\newblock {\em arXiv preprint arXiv:2404.05268}, 2024.

\bibitem{johnson2016perceptual}
Justin Johnson, Alexandre Alahi, and Li~Fei-Fei.
\newblock Perceptual losses for real-time style transfer and super-resolution.
\newblock In {\em ECCV}, 2016.

\bibitem{ke2023repurposing}
Bingxin Ke, Anton Obukhov, Shengyu Huang, Nando Metzger, Rodrigo~Caye Daudt, and Konrad Schindler.
\newblock Repurposing diffusion-based image generators for monocular depth estimation.
\newblock In {\em Proceedings of the IEEE/CVF Conference on Computer Vision and Pattern Recognition (CVPR)}, 2024.

\bibitem{kirillov2023segment}
Alexander Kirillov, Eric Mintun, Nikhila Ravi, Hanzi Mao, Chloe Rolland, Laura Gustafson, Tete Xiao, Spencer Whitehead, Alexander~C Berg, Wan-Yen Lo, et~al.
\newblock Segment anything.
\newblock In {\em ICCV}, 2023.

\bibitem{kumari2023multi}
Nupur Kumari, Bingliang Zhang, Richard Zhang, Eli Shechtman, and Jun-Yan Zhu.
\newblock Multi-concept customization of text-to-image diffusion.
\newblock In {\em CVPR}, 2023.

\bibitem{flux}
Black~Forest Labs.
\newblock Flux.
\newblock \url{https://github.com/black-forest-labs/flux}, 2024.

\bibitem{lai2024unleashing}
Bolin Lai, Felix Juefei-Xu, Miao Liu, Xiaoliang Dai, Nikhil Mehta, Chenguang Zhu, Zeyi Huang, James~M Rehg, Sangmin Lee, Ning Zhang, et~al.
\newblock Unleashing in-context learning of autoregressive models for few-shot image manipulation.
\newblock {\em arXiv preprint arXiv:2412.01027}, 2024.

\bibitem{li2023matting}
Jiachen Li, Jitesh Jain, and Humphrey Shi.
\newblock Matting anything.
\newblock {\em arXiv: 2306.05399}, 2023.

\bibitem{mattinganything}
Jiachen Li, Jitesh Jain, and Humphrey Shi.
\newblock Matting anything.
\newblock In {\em Proceedings of the IEEE/CVF Conference on Computer Vision and Pattern Recognition}, pages 1775--1785, 2024.

\bibitem{bnnlan}
Junyi Li, Zhilu Zhang, Xiaoyu Liu, Chaoyu Feng, Xiaotao Wang, Lei Lei, and Wangmeng Zuo.
\newblock Spatially adaptive self-supervised learning for real-world image denoising.
\newblock In {\em Proceedings of the IEEE Conference on Computer Vision and Pattern Recognition}, 2023.

\bibitem{li2025edmb}
Yachuan Li, Xavier~Soria Poma, Yun Bai, Qian Xiao, Chaozhi Yang, Guanlin Li, and Zongmin Li.
\newblock Edmb: Edge detector with mamba.
\newblock {\em arXiv preprint arXiv:2501.04846}, 2025.

\bibitem{li2017universal}
Yijun Li, Chen Fang, Jimei Yang, Zhaowen Wang, Xin Lu, and Ming-Hsuan Yang.
\newblock Universal style transfer via feature transforms.
\newblock In {\em NIPS}, 2017.

\bibitem{li2024dual}
Zijie Li, Henry Li, Yichun Shi, Amir~Barati Farimani, Yuval Kluger, Linjie Yang, and Peng Wang.
\newblock Dual diffusion for unified image generation and understanding.
\newblock {\em arXiv preprint arXiv:2501.00289}, 2024.

\bibitem{ctrlcolor}
Zhexin Liang, Zhaochen Li, Shangchen Zhou, Chongyi Li, and Chen~Change Loy.
\newblock Control color: Multimodal diffusion-based interactive image colorization.
\newblock {\em arXiv preprint arXiv:2402.10855}, 2024.

\bibitem{dinoir}
Xin Lin, Chao Ren, Kelvin~CK Chan, Lu~Qi, Jinshan Pan, and Ming-Hsuan Yang.
\newblock Multi-task image restoration guided by robust dino features.
\newblock {\em arXiv preprint arXiv:2312.01677}, 2023.

\bibitem{scpgabnet}
Xin Lin, Chao Ren, and Xiao Liu.
\newblock Unsupervised image denoising in real-world scenarios via self-collaboration parallel generative adversarial branches.
\newblock In {\em ICCV}, 2023.

\bibitem{llrrnet}
Xin Lin, Jingtong Yue, Sixian Ding, Chao Ren, Lu~Qi, and Ming-Hsuan Yang.
\newblock Dual degradation representation for joint deraining and low-light enhancement in the dark.
\newblock {\em IEEE Transactions on Circuits and Systems for Video Technology}, 2024.

\bibitem{scpgan}
Xin Lin, Yuyan Zhou, Jingtong Yue, Chao Ren, Kelvin~CK Chan, Lu~Qi, and Ming-Hsuan Yang.
\newblock Re-boosting self-collaboration parallel prompt gan for unsupervised image restoration.
\newblock {\em arXiv preprint arXiv:2408.09241}, 2024.

\bibitem{playground-v3}
Bingchen Liu, Ehsan Akhgari, Alexander Visheratin, Aleks Kamko, Linmiao Xu, Shivam Shrirao, Chase Lambert, Joao Souza, Suhail Doshi, and Daiqing Li.
\newblock Playground v3: Improving text-to-image alignment with deep-fusion large language models.
\newblock {\em arXiv preprint arXiv:2409.10695}, 2024.

\bibitem{lumina-mgpt}
Dongyang Liu, Shitian Zhao, Le~Zhuo, Weifeng Lin, Yu~Qiao, Hongsheng Li, and Peng Gao.
\newblock Lumina-mgpt: Illuminate flexible photorealistic text-to-image generation with multimodal generative pretraining, 2024.

\bibitem{StrDiffusion}
Haipeng Liu, Yang Wang, Biao Qian, Meng Wang, and Yong Rui.
\newblock Structure matters: Tackling the semantic discrepancy in diffusion models for image inpainting.
\newblock In {\em Proceedings of the IEEE/CVF Conference on Computer Vision and Pattern Recognition}, 2024.

\bibitem{liu2024world}
Hao Liu, Wilson Yan, Matei Zaharia, and Pieter Abbeel.
\newblock World model on million-length video and language with ringattention.
\newblock {\em arXiv preprint arXiv:2402.08268}, 2024.

\bibitem{liu2023one}
Minghua Liu, Chao Xu, Haian Jin, Linghao Chen, Mukund Varma~T, Zexiang Xu, and Hao Su.
\newblock One-2-3-45: Any single image to 3d mesh in 45 seconds without per-shape optimization.
\newblock {\em Advances in Neural Information Processing Systems}, 2023.

\bibitem{liu2023zero}
Ruoshi Liu, Rundi Wu, Basile Van~Hoorick, Pavel Tokmakov, Sergey Zakharov, and Carl Vondrick.
\newblock Zero-1-to-3: Zero-shot one image to 3d object.
\newblock In {\em Proceedings of the IEEE/CVF international conference on computer vision}, 2023.

\bibitem{long2024wonder3d}
Xiaoxiao Long, Yuan-Chen Guo, Cheng Lin, Yuan Liu, Zhiyang Dou, Lingjie Liu, Yuexin Ma, Song-Hai Zhang, Marc Habermann, Christian Theobalt, et~al.
\newblock Wonder3d: Single image to 3d using cross-domain diffusion.
\newblock In {\em Proceedings of the IEEE/CVF conference on computer vision and pattern recognition}, 2024.

\bibitem{SEDD}
Aaron Lou, Chenlin Meng, and Stefano Ermon.
\newblock Discrete diffusion modeling by estimating the ratios of the data distribution.
\newblock {\em arXiv preprint arXiv:2310.16834}, 2023.

\bibitem{ma2024janusflow}
Yiyang Ma, Xingchao Liu, Xiaokang Chen, Wen Liu, Chengyue Wu, Zhiyu Wu, Zizheng Pan, Zhenda Xie, Haowei Zhang, Liang Zhao, et~al.
\newblock Janusflow: Harmonizing autoregression and rectified flow for unified multimodal understanding and generation.
\newblock {\em arXiv preprint arXiv:2411.07975}, 2024.

\bibitem{concrete}
Chenlin Meng, Kristy Choi, Jiaming Song, and Stefano Ermon.
\newblock Concrete score matching: Generalized score matching for discrete data.
\newblock {\em Advances in Neural Information Processing Systems}, 35:34532--34545, 2022.

\bibitem{mescheder2019occupancy}
Lars Mescheder, Michael Oechsle, Michael Niemeyer, Sebastian Nowozin, and Andreas Geiger.
\newblock Occupancy networks: Learning 3d reconstruction in function space.
\newblock In {\em Proceedings of the IEEE/CVF conference on computer vision and pattern recognition}, 2019.

\bibitem{Midjourney}
Midjourney.
\newblock Midjourney.
\newblock \url{https://www.midjourney.com}, 2024.

\bibitem{mildenhall2021nerf}
Ben Mildenhall, Pratul~P Srinivasan, Matthew Tancik, Jonathan~T Barron, Ravi Ramamoorthi, and Ren Ng.
\newblock Nerf: Representing scenes as neural radiance fields for view synthesis.
\newblock {\em Communications of the ACM}, 2021.

\bibitem{niemeyer2020differentiable}
Michael Niemeyer, Lars Mescheder, Michael Oechsle, and Andreas Geiger.
\newblock Differentiable volumetric rendering: Learning implicit 3d representations without 3d supervision.
\newblock In {\em Proceedings of the IEEE/CVF conference on computer vision and pattern recognition}, 2020.

\bibitem{openai2025gpt4o}
OpenAI.
\newblock Addendum to gpt-4o system card: 4o image generation, 2025.
\newblock Accessed: 2025-04-02.

\bibitem{pan2019deep}
Junyi Pan, Xiaoguang Han, Weikai Chen, Jiapeng Tang, and Kui Jia.
\newblock Deep mesh reconstruction from single rgb images via topology modification networks.
\newblock In {\em Proceedings of the IEEE/CVF International Conference on Computer Vision}, 2019.

\bibitem{pavlakos2019expressive}
Georgios Pavlakos, Vasileios Choutas, Nima Ghorbani, Timo Bolkart, Ahmed~AA Osman, Dimitrios Tzionas, and Michael~J Black.
\newblock Expressive body capture: 3d hands, face, and body from a single image.
\newblock In {\em Proceedings of the IEEE/CVF conference on computer vision and pattern recognition}, 2019.

\bibitem{podell2023sdxl}
Dustin Podell, Zion English, Kyle Lacey, Andreas Blattmann, Tim Dockhorn, Jonas M{\"u}ller, Joe Penna, and Robin Rombach.
\newblock Sdxl: Improving latent diffusion models for high-resolution image synthesis.
\newblock {\em arXiv preprint arXiv:2307.01952}, 2023.

\bibitem{sdxl}
Dustin Podell, Zion English, Kyle Lacey, Andreas Blattmann, Tim Dockhorn, Jonas M{\"u}ller, Joe Penna, and Robin Rombach.
\newblock {SDXL}: Improving latent diffusion models for high-resolution image synthesis.
\newblock In {\em The Twelfth International Conference on Learning Representations (ICLR)}, 2024.

\bibitem{qian2023magic123}
Guocheng Qian, Jinjie Mai, Abdullah Hamdi, Jian Ren, Aliaksandr Siarohin, Bing Li, Hsin-Ying Lee, Ivan Skorokhodov, Peter Wonka, Sergey Tulyakov, et~al.
\newblock Magic123: One image to high-quality 3d object generation using both 2d and 3d diffusion priors.
\newblock {\em arXiv preprint arXiv:2306.17843}, 2023.

\bibitem{ram}
Chu-Jie Qin, Rui-Qi Wu, Zikun Liu, Xin Lin, Chun-Le Guo, Hyun~Hee Park, and Chongyi Li.
\newblock Restore anything with masks: Leveraging mask image modeling for blind all-in-one image restoration.
\newblock In {\em ECCV}, 2024.

\bibitem{ramesh2022hierarchical}
Aditya Ramesh, Prafulla Dhariwal, Alex Nichol, Casey Chu, and Mark Chen.
\newblock Hierarchical text-conditional image generation with clip latents.
\newblock {\em arXiv preprint arXiv:2204.06125}, 2022.

\bibitem{ravi2024sam}
Nikhila Ravi, Valentin Gabeur, Yuan-Ting Hu, Ronghang Hu, Chaitanya Ryali, Tengyu Ma, Haitham Khedr, Roman R{\"a}dle, Chloe Rolland, Laura Gustafson, et~al.
\newblock {SAM 2}: Segment anything in images and videos.
\newblock {\em ICLR}, 2025.

\bibitem{reizenstein2021common}
Jeremy Reizenstein, Roman Shapovalov, Philipp Henzler, Luca Sbordone, Patrick Labatut, and David Novotny.
\newblock Common objects in 3d: Large-scale learning and evaluation of real-life 3d category reconstruction.
\newblock In {\em ICCV}, 2021.

\bibitem{ntire_sr}
Bin Ren, Yawei Li, Nancy Mehta, and Radu Timofte.
\newblock The ninth ntire 2024 efficient super-resolution challenge report.
\newblock In {\em Proceedings of the IEEE/CVF Conference on Computer Vision and Pattern Recognition (CVPR) Workshops}, 2024.

\bibitem{stablediffusion}
Robin Rombach, Andreas Blattmann, Dominik Lorenz, Patrick Esser, and Bj{\"o}rn Ommer.
\newblock High-resolution image synthesis with latent diffusion models.
\newblock In {\em Proceedings of the IEEE/CVF conference on computer vision and pattern recognition (CVPR)}, pages 10684--10695, 2022.

\bibitem{sd1.5}
Robin Rombach, Andreas Blattmann, Dominik Lorenz, Patrick Esser, and Bj\"orn Ommer.
\newblock High-resolution image synthesis with latent diffusion models.
\newblock In {\em Proceedings of the IEEE/CVF Conference on Computer Vision and Pattern Recognition (CVPR)}, pages 10684--10695, June 2022.

\bibitem{ruiz2023dreambooth}
Nataniel Ruiz, Yuanzhen Li, Varun Jampani, Yael Pritch, Michael Rubinstein, and Kfir Aberman.
\newblock Dreambooth: Fine tuning text-to-image diffusion models for subject-driven generation.
\newblock In {\em CVPR}, 2023.

\bibitem{saharia2022photorealistic}
Chitwan Saharia, William Chan, Saurabh Saxena, Lala Li, Jay Whang, Emily~L Denton, Kamyar Ghasemipour, Raphael Gontijo~Lopes, Burcu Karagol~Ayan, Tim Salimans, et~al.
\newblock Photorealistic text-to-image diffusion models with deep language understanding.
\newblock {\em Advances in neural information processing systems}, 2022.

\bibitem{mdlm}
Subham Sahoo, Marianne Arriola, Yair Schiff, Aaron Gokaslan, Edgar Marroquin, Justin Chiu, Alexander Rush, and Volodymyr Kuleshov.
\newblock Simple and effective masked diffusion language models.
\newblock {\em Advances in Neural Information Processing Systems}, 37:130136--130184, 2024.

\bibitem{shi2024relationbooth}
Qingyu Shi, Lu~Qi, Jianzong Wu, Jinbin Bai, Jingbo Wang, Yunhai Tong, Xiangtai Li, and Ming-Husan Yang.
\newblock Relationbooth: Towards relation-aware customized object generation.
\newblock {\em arXiv preprint arXiv:2410.23280}, 2024.

\bibitem{sun2024coser}
Haoze Sun, Wenbo Li, Jianzhuang Liu, Haoyu Chen, Renjing Pei, Xueyi Zou, Youliang Yan, and Yujiu Yang.
\newblock Coser: Bridging image and language for cognitive super-resolution.
\newblock In {\em Proceedings of the IEEE/CVF Conference on Computer Vision and Pattern Recognition}, pages 25868--25878, 2024.

\bibitem{sun2023generative}
Quan Sun, Qiying Yu, Yufeng Cui, Fan Zhang, Xiaosong Zhang, Yueze Wang, Hongcheng Gao, Jingjing Liu, Tiejun Huang, and Xinlong Wang.
\newblock Generative pretraining in multimodality.
\newblock {\em arXiv preprint arXiv:2307.05222}, 2023.

\bibitem{swerdlow2025unified}
Alexander Swerdlow, Mihir Prabhudesai, Siddharth Gandhi, Deepak Pathak, and Katerina Fragkiadaki.
\newblock Unified multimodal discrete diffusion.
\newblock {\em arXiv preprint arXiv:2503.20853}, 2025.

\bibitem{team2024chameleon}
Chameleon Team.
\newblock Chameleon: Mixed-modal early-fusion foundation models.
\newblock {\em arXiv preprint arXiv:2405.09818}, 2024.

\bibitem{gemini-2-0-flash}
Gemini Team, Rohan Anil, Sebastian Borgeaud, Jean-Baptiste Alayrac, Jiahui Yu, Radu Soricut, Johan Schalkwyk, Andrew~M Dai, Anja Hauth, Katie Millican, et~al.
\newblock Gemini: a family of highly capable multimodal models.
\newblock {\em arXiv preprint arXiv:2312.11805}, 2023.

\bibitem{tong2024metamorph}
Shengbang Tong, David Fan, Jiachen Zhu, Yunyang Xiong, Xinlei Chen, Koustuv Sinha, Michael Rabbat, Yann LeCun, Saining Xie, and Zhuang Liu.
\newblock Metamorph: Multimodal understanding and generation via instruction tuning.
\newblock {\em arXiv preprint arXiv:2412.14164}, 2024.

\bibitem{wang2024illume}
Chunwei Wang, Guansong Lu, Junwei Yang, Runhui Huang, Jianhua Han, Lu~Hou, Wei Zhang, and Hang Xu.
\newblock Illume: Illuminating your llms to see, draw, and self-enhance.
\newblock {\em arXiv preprint arXiv:2412.06673}, 2024.

\bibitem{wang2018pixel2mesh}
Nanyang Wang, Yinda Zhang, Zhuwen Li, Yanwei Fu, Wei Liu, and Yu-Gang Jiang.
\newblock Pixel2mesh: Generating 3d mesh models from single rgb images.
\newblock In {\em Proceedings of the European conference on computer vision (ECCV)}, 2018.

\bibitem{msdiffusion}
Xierui Wang, Siming Fu, Qihan Huang, Wanggui He, and Hao Jiang.
\newblock Ms-diffusion: Multi-subject zero-shot image personalization with layout guidance.
\newblock {\em arXiv preprint arXiv:2406.07209}, 2024.

\bibitem{wang2024emu3}
Xinlong Wang, Xiaosong Zhang, Zhengxiong Luo, Quan Sun, Yufeng Cui, Jinsheng Wang, Fan Zhang, Yueze Wang, Zhen Li, Qiying Yu, et~al.
\newblock Emu3: Next-token prediction is all you need.
\newblock {\em arXiv preprint arXiv:2409.18869}, 2024.

\bibitem{wang2023context}
Zhendong Wang, Yifan Jiang, Yadong Lu, Pengcheng He, Weizhu Chen, Zhangyang Wang, Mingyuan Zhou, et~al.
\newblock In-context learning unlocked for diffusion models.
\newblock {\em NeurIPS}, 2023.

\bibitem{wang2024genartist}
Zhenyu Wang, Aoxue Li, Zhenguo Li, and Xihui Liu.
\newblock Genartist: Multimodal llm as an agent for unified image generation and editing.
\newblock {\em NeurIPS}, 2024.

\bibitem{warren2024effective}
Alex Warren, Ke~Xu, Jiaying Lin, Gary~KL Tam, and Rynson~WH Lau.
\newblock Effective video mirror detection with inconsistent motion cues.
\newblock In {\em CVPR}, 2024.

\bibitem{DiffSensei}
Jianzong Wu, Chao Tang, Jingbo Wang, Yanhong Zeng, Xiangtai Li, and Yunhai Tong.
\newblock Diffsensei: Bridging multi-modal llms and diffusion models for customized manga generation.
\newblock {\em CVPR}, 2025.

\bibitem{wu2025harmonizing}
Size Wu, Wenwei Zhang, Lumin Xu, Sheng Jin, Zhonghua Wu, Qingyi Tao, Wentao Liu, Wei Li, and Chen~Change Loy.
\newblock Harmonizing visual representations for unified multimodal understanding and generation.
\newblock {\em arXiv preprint arXiv:2503.21979}, 2025.

\bibitem{wu2024vila}
Yecheng Wu, Zhuoyang Zhang, Junyu Chen, Haotian Tang, Dacheng Li, Yunhao Fang, Ligeng Zhu, Enze Xie, Hongxu Yin, Li~Yi, et~al.
\newblock Vila-u: a unified foundation model integrating visual understanding and generation.
\newblock {\em arXiv preprint arXiv:2409.04429}, 2024.

\bibitem{SEED-Story}
Yifan Xia, Yuying Ge, Jing Zhang, Yuchao Dai, and Ming-Ming Cheng.
\newblock Seed-story: Multimodal long story generation with large language model.
\newblock {\em arXiv preprint arXiv:2407.08683}, 2024.

\bibitem{xiao2024omnigen}
Shitao Xiao, Yueze Wang, Junjie Zhou, Huaying Yuan, Xingrun Xing, Ruiran Yan, Shuting Wang, Tiejun Huang, and Zheng Liu.
\newblock Omnigen: Unified image generation.
\newblock {\em arXiv preprint arXiv:2409.11340}, 2024.

\bibitem{xu2024instantmesh}
Jiale Xu, Weihao Cheng, Yiming Gao, Xintao Wang, Shenghua Gao, and Ying Shan.
\newblock Instantmesh: Efficient 3d mesh generation from a single image with sparse-view large reconstruction models.
\newblock {\em arXiv preprint arXiv:2404.07191}, 2024.

\bibitem{depthanything}
Lihe Yang, Bingyi Kang, Zilong Huang, Xiaogang Xu, Jiashi Feng, and Hengshuang Zhao.
\newblock Depth anything: Unleashing the power of large-scale unlabeled data.
\newblock In {\em Proceedings of the IEEE/CVF Conference on Computer Vision and Pattern Recognition}, pages 10371--10381, 2024.

\bibitem{yang2024mastering}
Ling Yang, Zhaochen Yu, Chenlin Meng, Minkai Xu, Stefano Ermon, and Bin Cui.
\newblock Mastering text-to-image diffusion: Recaptioning, planning, and generating with multimodal llms.
\newblock In {\em ICML}, 2024.

\bibitem{sgt}
Hang Yu, Ruilin Li, Shaorong Xie, and Jiayan Qiu.
\newblock Shadow-enlightened image outpainting.
\newblock In {\em CVPR}, 2024.

\bibitem{sa2va}
Haobo Yuan, Xiangtai Li, Tao Zhang, Zilong Huang, Shilin Xu, Shunping Ji, Yunhai Tong, Lu~Qi, Jiashi Feng, and Ming-Hsuan Yang.
\newblock Sa2va: Marrying sam2 with llava for dense grounded understanding of images and videos.
\newblock {\em arXiv}, 2025.

\bibitem{yuan2024generative}
Yu~Yuan, Xijun Wang, Yichen Sheng, Prateek Chennuri, Xingguang Zhang, and Stanley Chan.
\newblock Generative photography: Scene-consistent camera control for realistic text-to-image synthesis.
\newblock {\em arXiv preprint arXiv:2412.02168}, 2024.

\bibitem{pano-sd}
Cheng Zhang, Qianyi Wu, Camilo Cruz~Gambardella, Xiaoshui Huang, Dinh Phung, Wanli Ouyang, and Jianfei Cai.
\newblock Taming stable diffusion for text to 360◦ panorama image generation.
\newblock In {\em Proceedings of the IEEE/CVF Conference on Computer Vision and Pattern Recognition}, 2024.

\bibitem{Zhang2023MagicBrush}
Kai Zhang, Lingbo Mo, Wenhu Chen, Huan Sun, and Yu~Su.
\newblock Magicbrush: A manually annotated dataset for instruction-guided image editing.
\newblock In {\em NeurIPS}, 2023.

\bibitem{controlnet}
Lvmin Zhang, Anyi Rao, and Maneesh Agrawala.
\newblock Adding conditional control to text-to-image diffusion models, 2023.

\bibitem{iclight}
Lvmin Zhang, Anyi Rao, and Maneesh Agrawala.
\newblock Scaling in-the-wild training for diffusion-based illumination harmonization and editing by imposing consistent light transport.
\newblock In {\em ICLR}, 2025.

\bibitem{k-net}
Wenwei Zhang, Jiangmiao Pang, Kai Chen, and Chen~Change Loy.
\newblock K-net: Towards unified image segmentation.
\newblock {\em Advances in Neural Information Processing Systems}, 34:10326--10338, 2021.

\bibitem{zhang2024itercomp}
Xinchen Zhang, Ling Yang, Guohao Li, Yaqi Cai, Jiake Xie, Yong Tang, Yujiu Yang, Mengdi Wang, and Bin Cui.
\newblock Itercomp: Iterative composition-aware feedback learning from model gallery for text-to-image generation.
\newblock {\em arXiv preprint arXiv:2410.07171}, 2024.

\bibitem{zhang2024ssr}
Yuxuan Zhang, Yiren Song, Jiaming Liu, Rui Wang, Jinpeng Yu, Hao Tang, Huaxia Li, Xu~Tang, Yao Hu, Han Pan, et~al.
\newblock Ssr-encoder: Encoding selective subject representation for subject-driven generation.
\newblock In {\em CVPR}, 2024.

\bibitem{zhao2024monoformer}
Chuyang Zhao, Yuxing Song, Wenhao Wang, Haocheng Feng, Errui Ding, Yifan Sun, Xinyan Xiao, and Jingdong Wang.
\newblock Monoformer: One transformer for both diffusion and autoregression.
\newblock {\em arXiv preprint arXiv:2409.16280}, 2024.

\bibitem{zheng2024birefnet}
Peng Zheng, Dehong Gao, Deng-Ping Fan, Li~Liu, Jorma Laaksonen, Wanli Ouyang, and Nicu Sebe.
\newblock Bilateral reference for high-resolution dichotomous image segmentation.
\newblock {\em CCAI}, 2024.

\bibitem{zhou2024transfusion}
Chunting Zhou, Lili Yu, Arun Babu, Kushal Tirumala, Michihiro Yasunaga, Leonid Shamis, Jacob Kahn, Xuezhe Ma, Luke Zettlemoyer, and Omer Levy.
\newblock Transfusion: Predict the next token and diffuse images with one multi-modal model.
\newblock {\em arXiv preprint arXiv:2408.11039}, 2024.

\bibitem{zhou2024magictailor}
Donghao Zhou, Jiancheng Huang, Jinbin Bai, Jiaze Wang, Hao Chen, Guangyong Chen, Xiaowei Hu, and Pheng-Ann Heng.
\newblock {MagicTailor}: Component-controllable personalization in text-to-image diffusion models.
\newblock {\em arXiv preprint arXiv:2410.13370}, 2024.

\bibitem{disenvisioner}
Zhiyu Zhu, Yingcong Chen, Zhenyu Xie, and Jingyi Yu.
\newblock Disenvisioner: Disentangled and enriched visual prompt for customized image generation.
\newblock {\em arXiv preprint arXiv:2410.02067}, 2024.

\bibitem{zuffi2018lions}
Silvia Zuffi, Angjoo Kanazawa, and Michael~J Black.
\newblock Lions and tigers and bears: Capturing non-rigid, 3d, articulated shape from images.
\newblock In {\em CVPR}, 2018.

\end{thebibliography}
}

\end{document}